\pdfoutput=1  %
\documentclass{article}
\usepackage[final, nonatbib]{neurips_2022}
\usepackage[utf8]{inputenc} %
\usepackage[T1]{fontenc}    %
\usepackage{hyperref}       %
\usepackage{url}            %
\usepackage{booktabs}       %
\usepackage{amsfonts}       %
\usepackage{nicefrac}       %
\usepackage{microtype}      %
\usepackage{xcolor}         %
\usepackage{subfiles}
\usepackage{wrapfig}

\usepackage{tabularx}
\usepackage{tabulary}
\usepackage{booktabs}
\usepackage{graphicx}
\usepackage{tikz}
\usetikzlibrary{arrows.meta, fit, backgrounds, positioning, calc}
\usepackage{float}
\usepackage{enumitem}

\graphicspath{ {./figures/} }

\usepackage{subcaption}

\usepackage{algorithm,algorithmic}

\usepackage[sort, numbers]{natbib}
\definecolor{custom}{rgb}{0.0, 0.0, 0.0}
\hypersetup{colorlinks, citecolor=custom}

\usepackage[section,above,below]{placeins}
\usepackage[normalem]{ulem}

\newcommand\samethanks[1][\value{footnote}]{\footnotemark[#1]}

\usepackage{silence}
\usepackage{scrwfile}
\TOCclone[\appendixname]{toc}{atoc}
\newcommand\StartAppendixEntries{}
\AfterTOCHead[toc]{%
  \renewcommand\StartAppendixEntries{\value{tocdepth}=-10000\relax}%
}
\AfterTOCHead[atoc]{%
  \edef\maintocdepth{\the\value{tocdepth}}%
  \value{tocdepth}=-10000\relax
  \renewcommand\StartAppendixEntries{\value{tocdepth}=\maintocdepth\relax}%
}
\usepackage{amsmath, amssymb, mathtools, amsthm}
\usepackage{centernot}
\usepackage{float}
\usepackage{enumitem}
\usepackage{nccmath}
\usepackage{mdframed}
\usepackage{xcolor}
\usepackage{bbm}     %
\usepackage{bm}      %

\definecolor{mydarkblue}{rgb}{0,0.08,0.45}

\DeclareMathOperator{\tr}{tr}
\DeclareMathOperator{\diag}{diag}
\DeclareMathOperator{\rank}{rank}
\DeclareMathOperator{\ran}{ran}

\DeclareMathOperator*{\argmin}{arg\,min}

\newcommand{\Pprob}{\mathbb{P}} %

\DeclarePairedDelimiter{\norm}{\lVert}{\rVert}

\DeclarePairedDelimiterX{\klx}[2]{(}{)}{#1\,\delimsize\|\,#2}
\newcommand{\kl}[2]{\mathrm{KL}\klx{#1}{#2}}

\newcommand{\frob}[1]{\norm{#1}_{\mathrm{F}}}
\newcommand{\oper}[1]{\norm{#1}_{\mathrm{op}}}

\newcommand{\vect}[1]{\mathbf{#1}}
\newcommand{\x}{\vect{x}}

\newcommand{\mat}[1]{\mathsf{#1}}
\newcommand{\mE}{\mathbb{E}}

\newcommand{\set}[1]{\mathcal{#1}}

\newcommand{\Tu}{\mathrm{Tu}}
\newcommand{\varbudget}{\Gamma}
\newcommand{\specgap}{\delta}
\newcommand{\tempuncert}{\Tu(\specgap, \varbudget)}

\newcommand{\stiefel}[2]{\mathcal{V}_{#1}(\mathbb{R}^{#2})}
\newcommand{\grass}[2]{\mathcal{G}_{#1}(\mathbb{R}^{#2})}

\newcommand{\dist}{\mathrm{dist}}
\newcommand{\chord}[2]{d_{\mathrm{c}}(#1, #2)}
\newcommand{\proje}[2]{d_{2}(#1, #2)}

\newcommand{\defeq}{\coloneqq}
\newcommand{\xxT}[2]{{#1}_{#2}{#1}_{#2}^{\top}}
\DeclareMathSymbol{\shortminus}{\mathbin}{AMSa}{"39}
\def\totalbound{\Lambda}
\def\ioblock{\begin{pmatrix}
  \mat{I}_{k\times k} & 0 \\
  0 & 0
  \end{pmatrix}}
\newcommand{\QEDA}{\hfill\ensuremath{\blacksquare}}
\newcommand{\cX}{\mathcal{X}}

\theoremstyle{plain}
\newtheorem{theorem}{Theorem}
\newtheorem{lemma}[theorem]{Lemma}
\newtheorem{proposition}[theorem]{Proposition}
\newtheorem{corollary}[theorem]{Corollary}

\theoremstyle{definition}
\newtheorem{definition}{Definition}

\theoremstyle{definition}
\newtheorem{remark}{Remark}

\newmdtheoremenv[
  linewidth=0.5pt,
  skipabove=10pt,
  skipbelow=10pt,
  innerleftmargin=10pt,
  innerrightmargin=10pt,
  innertopmargin=10pt,
  innerbottommargin=10pt
]{goal}{Goal}

\definecolor{patchbg}{rgb}{1.0,0.96,0.85}
\definecolor{patchrule}{rgb}{0.78,0.49,0.0}

\setlist[itemize]{itemsep=2pt, topsep=2pt, leftmargin=1.5em}
\setlist[enumerate]{itemsep=2pt, topsep=2pt, leftmargin=1.5em}

\newcommand{\seqA}{\set{A}}  %
\newcommand{\seqX}{\set{X}}  %
\newcommand{\minimaxL}{\mathcal{R}^*} %

\title{Robust Streaming PCA}

\author{%
Daniel Bienstock\thanks{Authors are listed alphabetically.}\\
IEOR Department\\
Columbia University\\
\And
Minchan Jeong\samethanks\\
Graduate School of AI\\
KAIST\\
\And
Apurv Shukla\samethanks\\
IEOR Department\\
Columbia University\\
\And
Se-Young Yun\samethanks\\
Graduate School of AI\\
KAIST\\
\AND
\texttt{\{dano,as5197\}@columbia.edu, \{mcjeong,yunseyoung\}@kaist.ac.kr}\\
}

\begin{document}

\maketitle

\begin{abstract}
\label{abstract}
We consider streaming principal component analysis when the stochastic data generating model is subject to perturbations.
While existing models assume a fixed covariance, we adopt a robust perspective where the covariance matrix belongs to a temporal uncertainty set.
Under this setting, we provide fundamental limits on convergence of any algorithm recovering principal components.
We analyze the convergence of the noisy power method and Oja's algorithm, both studied for the stationary data generating model, and argue that the noisy power method is rate-optimal in our setting.
Finally, we demonstrate the validity of our analysis through numerical experiments on synthetic and real-world datasets. 
\end{abstract}

\section{Introduction}
\label{introduction}
Principal component analysis (PCA) is a standard method for obtaining a low-dimensional representation of observed data~\citep{dunteman1989principal}.
Its classical implementations store all observations and run in cubic time, with prohibitive memory and runtime cost.

Several works address this cost by designing streaming PCA algorithms with near-optimal memory complexity~\citep{edo2013, mitliagkas2013, ghhashami2016, npm}.
These algorithms assume that all observations lie in the same low-dimensional space, a fragile assumption when unobserved alterations corrupt the data.
Data attacks on power grids can change the covariance matrix estimated from sensors~\citep{stochdef, escobar2019learning, bienstock2019variance}; PCA applied to stock returns must accommodate macroeconomic drift~\citep{connor1986performance, bai2002determining, goldfarb2003robust}; product pricing requires tracking cross-product elasticity and demand~\citep{javanmard2019dynamic, mueller2019low}.
In all of these scenarios, the data generating model changes at every instant and downstream decisions depend on identifying the new model.

Existing work considers perturbations of data lying in a \textit{fixed} low-dimensional space~\citep{candes2011robust, pimentel2017adversarial}.
It determines the worst-case position of the adversarial data point that incurs the maximum subspace-estimation error, with the distance between subspaces measured by their principal angle.
PCA has also been cast as stochastic optimization~\citep{shamir1, shamir2, mianjy2018stochastic, arora_1}.
Our work assumes that the data generating model changes at every time instant and proposes near-optimal algorithms for recovering principal components under that framework.

We assume a system relies on the time-series of $p$-dimensional vectors sampled from a time-varying model.
The available observations are the vectors $(\vect{x}_{t})_{t=1}^{T}$.
Following the spiked covariance model~\citep{Iain2001, johnstone2018pca}, we consider the time-dependent environment:
\begin{equation}
\label{eqn:spike}
\vect{x}_t = \bigl(\xxT{\mat{A}}{t} + \sigma^2 \mat{I}_{p \times p}
\bigr)^{1/2} \vect{z}_t,
\end{equation}
where $\mat{A}_t \in \mathbb{R}^{p \times k}$ can vary with time, and $\vect{z}_t \in \mathbb{R}^{p}$ is a random vector with independent, mean-zero, unit-variance, sub-Gaussian entries~\citep{Vershynin2018}.
Precisely, each coordinate $z_{t,i}$ satisfies $\mathbb{E}[z_{t,i}] = 0$, $\mathbb{E}[z_{t,i}^2] = 1$, and $\| z_{t,i} \|_{\psi_2} \leq \kappa$ for an absolute constant $\kappa > 0$, representing the sub-Gaussian norm of the entries, which we absorb into the universal constants throughout the analysis.
This ensures $\mathbb{E}[\vect{z}_t \vect{z}_t^\top] = \mat{I}_{p \times p}$ and hence $\mathrm{Cov}(\vect{x}_t) = \Sigma_t = \xxT{\mat{A}}{t} + \sigma^2 \mat{I}_{p \times p}$, recovering the covariance structure of the standard spiked model~\citep{Iain2001}.
The Gaussian model ($\vect{z}_t \sim \mathcal{N}(\vect{0}, \mat{I})$) is a special case.
The i.i.d. structure of $\vect{z}_t$ is a distributional assumption for concentration, not a restriction on the covariance: the square root $\Sigma_t^{1/2}$ realizes any spiked $\Sigma_t$.

The parameter $k$, with $p > 2k$, is the desired number of principal components.
The signal $\mat{A}_t$ varies slowly within a temporal uncertainty set $\tempuncert$, with signal strength at least $\delta$ and per-step drift bounded by a budget $\varbudget$ (defined in equation~\eqref{eq:robustmodel}).
Our goal is to recover the top-$k$ principal components of the terminal subspace $\mat{A}_T$.
In financial applications, our model captures market evolution through the changing $\mat{A}_T$, with the goal of explaining the market conditions at the chosen horizon $T$.

Previous work on the stationary setting assumes $\mat{A}_t\!=\!
\mat{A}$ for all $t$ and computes a basis for the column space of $\mat{A}$ via streaming algorithms~\citep{mitliagkas2013, npm, li2018near, HuangNW21, johnstone2018pca}.
The recovery rate in this setting depends on the noise level $\sigma$, the ambient dimension $p$, the rank $k$ of $\mat{A}$, and the spectral gap $\delta$ between the $k$-th and $(k\!+\!1)$-th singular values of $\mat{A}\mat{A}^\top$.
We study the complementary case where the column space of $\mat{A}$ varies across time, propose a tractable analysis framework for streaming PCA under perturbations to the data generating model, and make the following contributions.

Throughout the paper, $O$, $\Omega$, $\Theta$ denote standard asymptotic notation, and $\tilde{O}$, $\tilde{\Theta}$ are the variants that additionally suppress multiplicative factors polynomial in $\log(pT)$ and~$\kappa$.
We write $a \lesssim b$ ($a \gtrsim b$) to mean $a \le C b$ ($a \ge c b$) for absolute constants $C, c > 0$.

\begin{enumerate}[leftmargin=*, topsep=0pt, partopsep=0pt, parsep=0pt, itemsep=0.4em]%
\item (Lower bound; Section~\ref{lowerbound}) We prove a minimax lower bound over $\tempuncert$ for recovering the top singular vectors (Theorem~\ref{thm:lower_bound}), reducing estimation to a hypothesis test over a packing of the Grassmannian and applying Fano's method~\citep{tsybakov2008introduction}.
The error decreases as $O(p^{1/2}T^{\shortminus 1/2})$ until $T$ reaches the critical scale $\Theta(p^{1/3}\varbudget^{\shortminus 2/3})$, then plateaus at $O(p^{1/3}\varbudget^{1/3})$: under drift the far past carries no information about $\mat{A}_T$, so more observations stop helping ($\specgap,\sigma$ fixed; Theorem~\ref{thm:lower_bound} and~\eqref{eq:critical-time} give the full form).
The exponent is a double penalty on the adversary's separation: each drifting step contributes $s^2$ to the KL budget, and a larger separation also requires proportionally more such steps.
At $\varbudget=0$ the bound reduces to the classical $\Theta(T^{\shortminus 1/2})$ rate for the stationary spiked model~\citep{Cai2015, Vu2013}.
\item (Algorithm analysis; Section~\ref{algorithms}) We analyze two standard streaming-PCA algorithms with contrasting designs: the noisy power method~\citep{npm}, which processes data in blocks, and Oja's algorithm~\citep{oja_1}, which updates on single observations.
We determine the optimal block size (Lemma~\ref{lem:power-error}) and learning rate (Lemma~\ref{lem:oja-error}), analogous roles whose optimal values both scale as $\tilde{\Theta}(\varbudget^{\shortminus 2/3})$.
Theorems~\ref{thm:robust-power-method} and~\ref{thm:ojas-algorithm} then bound the convergence error of each; for the noisy power method the upper bound matches the minimax error in $p$, $\delta$, and $\varbudget$, making it rate-optimal under a mild condition.
Both guarantees follow from a single convergence theorem for the noisy power method under bounded perturbation and subspace drift (Theorem~\ref{thm:drifting-power-method}), which specializes to the stationary noisy power method when the drift vanishes.
\end{enumerate}

\textbf{Notation.}
We fix notation.
Bold uppercase letters denote matrices (e.g., $\mat{M}$) and bold lowercase letters denote vectors (e.g., $\vect{x}$).
For $\mat{M} \in \mathbb{R}^{p\times k}$, $\mat{M}_{i}$ denotes the $i$-th column, $\mat{M}_{i:j}$ the submatrix with columns $i$ through $j$, and $\mat{M}_{i,j}$ the $(i,j)$-entry.
$\lVert\,\cdot\,\rVert$ denotes the matrix 2-norm (operator norm) for matrices and the Euclidean norm for vectors.
We write $\stiefel{k}{p}$ for the set of $p\times k$ matrices with orthonormal columns, and $\mathrm{ran}(\mat{M})$ for the column space of $\mat{M}$.

The singular value decomposition of $\mat{M}$ is $\mathrm{SVD}(\mat{M})\!=\!\mat{U}\mat{D}\mat{V}^{\top}$, where $\mat{U}\!\in\!\stiefel{p}{p}$, $\mat{V}\!\in\!\stiefel{k}{k}$, and $\mat{D}\!\in\!\mathbb{R}^{p \times k}$ is diagonal with $i$-th entry $s_{i}(\mat{M})$.
Without loss of generality, we order the singular values and their vectors from largest to smallest, so $s_{i}(\mat{M})$ is the $i$-th singular value of $\mat{M}$.
When $\mathrm{rk}(\mat{M})\!=\!\mathrm{rk}(\tilde{\mat{M}})$ and $\mathrm{SVD}(\tilde{\mat{M}})\!=\!\tilde{\mat{U}}\tilde{\mat{D}}\tilde{\mat{V}}^{\top}$, the distance between $\mathrm{ran}(\mat{M})$ and $\mathrm{ran}(\tilde{\mat{M}})$ is $\lVert\xxT{\mat{U}}{1:k} - \xxT{\tilde{\mat{U}}}{1:k}\rVert$, abbreviated $d(\mat{M},\tilde{\mat{M}})$ when context is clear.

We write $\Sigma_t := \xxT{\mat{A}}{t} + \sigma^2\mat{I}$ for the population covariance at time~$t$, $\kappa := \max_{t,i} \lVert z_{t,i}\rVert_{\psi_2}$ for the sub-Gaussian parameter (with $\vect{z}_t = \Sigma_t^{-1/2}\vect{x}_t$), and $\mu \geq 1$ for the condition-number parameter of $\tempuncert$ (Definition~\ref{defn:tempuncert}), bounding $s_1(\xxT{\mat{A}}{t}) \leq \mu\delta$ uniformly in $t$.

$\mathcal{A}$ abbreviates the sequence of matrices $(\mat{A}_{t})_{t=1}^T$.
We write $\mathcal{X}\!\sim\!\mathcal{A}$ when each element $\vect{x}_t$ of $\mathcal{X} = (\vect{x}_{t})_{t=1}^T$ is drawn from our model~\eqref{eqn:spike}, and write $\mE_{\mathcal{X} \sim \seqA}[f]$ for the expectation of $f(\mathcal{X})$ under $\vect{x}_t = \Sigma_t^{1/2}\vect{z}_t$.

\section{Related work}
\label{sec:related-work}
Principal component analysis (PCA) has been studied across operations research, computer science, and statistics.
We highlight how our work differs from existing literature.

\textbf{Robust PCA.} 
Robust PCA retrieves principal components in the presence of outliers.
The cornerstone is principal component pursuit~\citep{candes2011robust}, which decomposes the observation matrix into a low-rank matrix and a sparse matrix with entries of arbitrarily large magnitude.
Follow-up algorithms span offline, batch, and online settings~\citep{feng2013online, goes2014robust, cherapanamjeri2017thresholding, narayanamurthy2019provable}.
Our work differs on two axes.
First, the data generating model is unrelated: robust PCA tolerates sparse outliers in a static observation matrix, whereas we model a time-varying covariance with bounded drift.
Second, the convex-optimization techniques (low-rank-plus-sparse decomposition, nuclear-norm relaxation) target a static observation matrix and do not extend to the time-varying covariance setting.

\textbf{Streaming PCA.} Streaming algorithms for PCA include~\citep{mitliagkas2013, jain2016streaming, npm, PAST, OPAST}; the noisy power method~\citep{npm} and Oja's algorithm~\citep{oja_1} we analyze are iterative methods that instantiate stochastic approximation-based solutions to the PCA optimization formulation~\citep{arora_1}.
Gradient-based variants include GRASTA~\citep{GRASTA}, an incremental online gradient method for learning over different subspaces; GROUSE~\citep{balzano2010online}, using gradient updates over the Grassmannian manifold; and Oja variants for the standard streaming PCA problem~\citep{song2016image, chen2018dimensionality, yang2018history, henriksen2019adaoja, OOja, Power-Oja}.
All these works assume a stationary (drift-free) data generating model and provide no theoretical guarantees for our time-varying setting.
Recent finite-sample work refines Oja in alternative non-IID regimes (Markovian streams~\citep{kumar2023markovian}, streaming sparse PCA~\citep{kumar2024sparse}, entrywise uncertainty quantification~\citep{kumar2025entrywise}, adversarial-order streams~\citep{price2024adversarial}), orthogonal to the bounded covariance drift here.

\textbf{Non-stationary online learning.} Our cube-root drift rate $\Gamma^{1/3}$ matches the variation-budget rate from non-stationary bandits and online learning: under a budget $V_T=\sum_t\lVert f_t-f_{t-1}\rVert$ on the reward-function change, the minimax dynamic regret is of order $V_T^{1/3} T^{2/3}$~\citep{besbes2014mab, garivier2011sliding}.
Closest to our setting is~\citet{krishnamurthy2025slowly}, who impose a pointwise per-step variation bound on each arm and obtain a matching cube-root regret.
Subsequent work extends the lower bound to smooth reward sequences~\citep{jia2023smooth} and the same rate to linear bandits~\citep{cheung2022hedging} and linear-MDP reinforcement learning~\citep{zhou2022nonstationary}.
The shared algorithmic template, sliding-window estimation with window length $\tau_{\mathrm{opt}}\sim V_T^{-2/3}$, parallels our optimal block size $B_{\mathrm{opt}}\sim\Gamma^{-2/3}$.
The subspace-tracking setting, however, calls for machinery absent from scalar-reward bandits: matrix Bernstein and matrix-product concentration for operator-norm control, random-initialization analysis for the power iteration, and the Davis--Kahan theorem to convert eigenvalue stability into eigenvector stability.

\section{Mathematical framework}
\label{framework}
Under the standard spiked covariance model, the top-$k$ principal components of $\Sigma$ span a fixed $k$-dimensional column space of $\mat{A}\in\mathbb{R}^{p\times k}$.
Previous work has focused on reconstructing this space from the observed time series.
Under our framework, we assume that the sequence of observations $(\vect{x}_{t})_{t=1}^{T}$ follows the time-dependent spiked covariance model~\eqref{eqn:spike}, and we consider the problem of computing top-$k$ singular vectors of $\mat{A}_T$.

Our setting models a slowly drifting signal subspace.
When the adversary is allowed to select an arbitrary sequence $(\mat{A}_{t})_{t=1}^T$, accurate recovery of the column space of $\mat{A}_T$ is impossible; we therefore restrict the adversary to a temporal uncertainty set that bounds the per-step change.

\begin{definition}
\label{defn:tempuncert}
Let $\delta, \Gamma \geq 0$ and $\mu \geq 1$.
The temporal uncertainty set $\tempuncert$ is:
\begin{equation}
\tempuncert := \bigl\{\,(\mat{A}_{t})_{t=1}^T :
  s_{k}(\xxT{\mat{A}}{t}) \geq \delta\,,\;
  s_{1}(\xxT{\mat{A}}{t}) \leq \mu\delta\,,\;
  \Vert \xxT{\mat{A}}{t} - \xxT{\mat{A}}{t-1} \Vert
  \leq \varbudget
\,\bigr\}.
\label{eq:robustmodel}
\end{equation}
\end{definition}

The floor $s_{k}(\xxT{\mat{A}}{t}) \geq \delta$ ensures the signal subspace $\ran(\mat{A}_t)$ is identifiable at every step.
The drift constraint $\lVert \xxT{\mat{A}}{t} - \xxT{\mat{A}}{t-1} \rVert \leq \Gamma$ bounds the rate at which the covariance structure evolves.
We assume $\delta > \varbudget$ throughout, so the per-step rotation cap $\varbudget/\delta$ stays strictly below one (Appendix~\ref{appendix:lower-bound} gives the Grassmannian translation of the drift bound).
Geometrically, $\varbudget/\delta < 1$ keeps the largest principal angle (Definition~\ref{def:principal_angles}) between consecutive subspaces below $90^{\circ}$, so every direction in the new subspace retains some alignment with the previous one and earlier observations still carry information about the current target.
At $\varbudget \ge \delta$ this angle can reach $90^{\circ}$, opening a direction in the new subspace orthogonal to the previous one and carrying no information about the current target.
The stationary case $\varbudget = 0$ is the lower endpoint.

The condition number $\mu$ bounds the ratio $s_1(\xxT{\mat{A}}{t})/s_k(\xxT{\mat{A}}{t})$ uniformly over time: $\mu=1$ at $k=1$, and $\mu=\Theta(1)$ for general $k$ requires comparable signal singular values.
Stationary streaming PCA has no analog of $\mu$ in its convergence rates~\citep{npm, Allen2017, jain2016streaming}: averaging over the whole history makes the sampling error vanish, leaving only the gap-determined limit.
Drift instead forces averaging over a finite block, making the per-block sampling error the binding term.
Bounding this per-block error uses the operator-norm and trace bounds $\lVert\Sigma_t\rVert \leq \mu\delta + \sigma^2$ and $\mathrm{tr}(\Sigma_t) \leq k\mu\delta + p\sigma^2$, which is where $\mu$ enters our NPM and Oja upper bounds (Theorems~\ref{thm:robust-power-method} and~\ref{thm:ojas-algorithm}).
At $\mu = \Theta(1)$ the $\mu$-slack collapses; the NPM bound then matches the minimax (Theorem~\ref{thm:lower_bound}) up to logarithmic factors in the noise-energy-dominated regime ($p\sigma^2 \gtrsim k\delta$), while Oja's bound retains a $p^{1/3}$ slack independent of $\mu$.
Recovering the $\mu$-dependence from the lower bound provably fails; the slack lies on the upper-bound side (Appendix~\ref{appendix:mu-gap}).

We consider streaming algorithms: a single chronological pass with limited (sublinear) memory, producing $k$ orthonormal vectors in $\mathbb{R}^{p}$.
Write $\Phi$ for this family; for observations $\mathcal{X}=(\mathbf{x}_{t})_{t=1}^T$ from model~\eqref{eqn:spike}, the output $\phi_\mathcal{X}$ of $\phi\in\Phi$ is viewed interchangeably as a matrix in $\mathbb{R}^{p \times k}$ or the subspace it spans.

Definition~\ref{defn:algo_class} fixes how we treat the lower bound and when we call an algorithm \textit{rate-optimal}, following~\citep{tsybakov2008introduction, Cai2015, Vu2013}.

\begin{definition}
\label{defn:algo_class}
Let $\seqA=(\mat{A}_t)_{t=1}^T \in \tempuncert$ and $\phi \in \Phi$.
\begin{enumerate}[nosep, leftmargin=*]
\item The estimation error of $\phi$ given $\mathcal{X}\sim \mathcal{A}$ is the distance $d(\text{ran}(\mat{A}_T), \phi_\mathcal{X})$ between the column space of $\mat{A}_T$ and the space spanned by $\phi_\mathcal{X}$.
See Appendix~\ref{appendix:cumulative-metric} for the cumulative extension and Appendix~\ref{appendix:distance-metric} for the chordal-distance alternative.
\item The worst-case expected estimation error of $\phi$ over $\tempuncert$ is
\[
  \mathcal{R}^\phi := \sup_{\seqA \in \tempuncert}\mE_{\mathcal{X} \sim \seqA}\bigl[d(\mathrm{ran}(\mat{A}_T), \phi_\mathcal{X})\bigr].
\]
\item The minimax estimation error is $\minimaxL := \inf_{\phi \in \Phi}\mathcal{R}^\phi$.
\item $\phi$ is \textbf{rate-optimal} if $\mathcal{R}^\phi \le C \cdot \minimaxL$ for a constant $C > 0$ independent of the problem parameters $T$, $p$, $k$, $\sigma$, $\delta$, $\varbudget$.
\end{enumerate}
\end{definition}
We establish the minimax estimation error $\minimaxL$ and propose rate-optimal sublinear-time, single-pass algorithms for robust streaming PCA.

\section{Minimax lower bound}
\label{lowerbound}
For $\mathcal{A}=(\mat{A}_t)_{t=1}^T\in\tempuncert$ (Definition~\ref{defn:tempuncert}), no algorithm recovers $\mathrm{ran}(\mat{A}_T)$ with zero estimation error.
We derive a minimax lower bound on this error in $T,p,\sigma,\delta,\varbudget$.

\begin{theorem}[Lower Bound]
\label{thm:lower_bound}
Assume $\delta> \varbudget \geq 0$ and $p>2k$.
For any algorithm $\phi \in \Phi$, there exists a sequence $\mathcal{A}\in\tempuncert$ such that
\begin{equation}
\mE_{\mathcal{X}  \sim \seqA} \big[d\big(\mathrm{ran}(\mat{A}_T),\phi_{\cX} \big)\big] = \Omega \left( \min \left\{1\,, \left(\frac{\varbudget}{\delta} \right)^{{1}/{3}} \left(\frac{p\sigma^2(\sigma^2 + \delta)}{\delta^2} \right)^{{1}/{3}}\!\! +\,\, \frac{1}{\sqrt{T}}\left(\frac{p\sigma^2(\sigma^2 + \delta)}{\delta^2} \right)^{{1}/{2}}\,\right\} \right).
\end{equation}
Taking $\inf_{\phi\in\Phi}$ yields the same bound for $\mathcal{R}^*$.
\end{theorem}

The proof reduces estimation to an $M$-way hypothesis test via Fano's method~\citep{tsybakov2008introduction}.
The reduction places two competing demands on the candidate sequences $\mathcal{A}_1,\dots,\mathcal{A}_M\in\tempuncert$.
They must be pairwise $2s$-separated in terminal subspace distance, so any estimator with sub-$s$ error identifies the true index uniquely; they must also stay statistically close in KL, $\tfrac{1}{M}\sum_j\kl{\Pprob_{\mathcal{A}_j}}{\Pprob_{\mathcal{A}_0}}\le\alpha\log M$, so Fano keeps the test failing with constant probability.
Here $s$ is the testing tolerance: half the pairwise separation.
Markov's inequality then turns the test's $\Omega(1)$ failure probability into a minimax bound of $\Omega(s)$ for the largest $s$ compatible with the KL budget.
Confining all candidates to a Grassmannian ball of radius $3s$ around a common reference caps each per-step covariance KL at $O(s^2)$; independence across time makes the pathwise KL additive, $\kl{\Pprob_{\mathcal{A}_j}}{\Pprob_{\mathcal{A}_0}}=\sum_{t=1}^T\mathrm{KL}_t$.
This leaves the KL budget the binding constraint while the volume-driven $\log M=\Theta(kp)$ stays slack.

The adversary picks whichever strategy spends the KL budget more efficiently, i.e., incurs less total KL per unit $s$.
Two regimes saturate the KL budget.
When $T\le 3s\delta/\varbudget$, the candidates rotate at the per-step drift cap $\varbudget/\delta$ throughout, accumulating $T\cdot s^2$ in total KL and giving $s\sim 1/\sqrt T$, the classical parametric rate that recovers~\citep{Cai2015, Vu2013}'s $\Theta(1/\sqrt T)$ form at $\varbudget=0$.
When $T>3s\delta/\varbudget$, the candidates use a \emph{lazy} trajectory: stay at the reference subspace for $T-\tau$ steps and rotate only over the last $\tau\defeq\lceil 3s\delta/\varbudget\rceil$ steps, contributing zero KL in the lazy stretch and $\tau\cdot s^2\sim(\delta/\varbudget)s^3$ in the active one.
The cube root has a one-line explanation: a larger separation is doubly expensive, costing $s^2$ per drifting step \emph{and} needing proportionally more drifting steps.
The error therefore plateaus at $O((\varbudget/\delta)^{1/3}(p\sigma^2(\sigma^2+\delta)/\delta^2)^{1/3})$ for every $T$ above the critical time
\begin{equation}
\label{eq:critical-time}
T_c := \left( \frac{\varbudget}{\delta} \right)^{-{2}/{3}} \left(\frac{p\sigma^2(\sigma^2 + \delta)}{\delta^2} \right)^{{1}/{3}}\,,
\end{equation}
where information becomes stale as fast as it is collected.
The two regimes combine into the bound of Theorem~\ref{thm:lower_bound}; the full proof is in Appendix~\ref{appendix:lower-bound}.

Theorems~\ref{thm:robust-power-method} and~\ref{thm:ojas-algorithm} prove that the noisy power method and Oja's algorithm attain near-optimal convergence guarantees.
When $s_1(\xxT{\mat{A}}{t})=\Theta(\delta)$, the spectral gap $s_k \ge \delta$ forces $\mu=\Theta(1)$, and Theorem~\ref{thm:robust-power-method} simplifies to
\begin{equation}
\mathcal{R}^{\mathrm{NPM}}=\tilde{O}\left(\Big(\frac{\Gamma}{\delta}\Big)^{1/3} \Big( \frac{(p\sigma^2+k\delta)(\sigma^2+\delta)}{\delta^2}\Big)^{1/3}\right),
\end{equation}
matching the lower bound up to logarithmic factors when $p\sigma^2\gtrsim k\delta$, which compares total noise energy $p\sigma^2$ with signal energy $k\delta$ (mild in high-dimensional settings).

\section{Convergence analysis}
\label{algorithms}
A generic template for algorithms $\phi \in \Phi$ of interest to us is as follows:
\textbf{(i)}~$\phi$ is initialized with a random matrix with
orthonormal columns $\hat{\mat{U}} \in \mathbb{R}^{p \times k}$;
\textbf{(ii)}~a running estimate of the principal components is
maintained as the columns of $\hat{\mat{U}}$;
\textbf{(iii)}~observations are projected onto the column space
of $\hat{\mat{U}}$ to update this estimate.
We consider two algorithms: a robust version of the noisy power method (Algorithm~\ref{algo:npm}) and Oja's algorithm (Algorithm~\ref{algo:oja}).
The noisy power method updates its estimate after a batch of $B$ observations, whereas Oja's algorithm processes each observation individually with learning rate~$\zeta$.

\subsection{Noisy power method}
\label{sec:robust_spike}
\begin{algorithm}[h]
\caption{Noisy power method with block size $B$~\citep{npm}.}
\label{algo:npm}
\begin{algorithmic}[1]
\STATE \textbf{Input:} Stream of vectors: $(\vect{x}_{t})_{t=1}^T$, block size: $B$, dimensions: $p,k$
\STATE Sample each element of $\mat{\hat{U}}(0)$ in $\mathcal{N}(0,1)$
\FOR{$\ell=1:L= \lfloor T/B \rfloor $}
\STATE $\mat{Y}(\ell)\leftarrow \mat{0}\in\mathbb{R}^{p\times k}$
\FOR{$t=(\ell-1)B+1:\ell B$}
\STATE $\mat{Y}(\ell)\leftarrow \mat{Y}(\ell)+\frac{1}{B} \xxT{\vect{x}}{t}\mat{\hat{U}}(\ell-1)$
\ENDFOR
\STATE $\mat{\hat{U}}(\ell)\leftarrow \text{Gram-Schmidt}(\mat{Y}(\ell))$
\ENDFOR
\STATE \textbf{Output:} $\mat{\hat{U}}(L)$
\end{algorithmic}
\end{algorithm}

The noisy power method is an iterative algorithm for computing the top-$k$ principal components of a matrix.
Starting from the random matrix $\hat{\mat{U}}(0)$ in $\mathbb{R}^{p\times k}$, the algorithm runs for $L$ iterations, each processing $B$ samples.
For observations drawn from a fixed distribution, the iterates converge to the top-$k$ principal subspace of the covariance.

When observations are drawn from model~\eqref{eqn:spike}, the covariance matrix shifts between iterations due to the perturbations allowed by~$\tempuncert$.
The columns of $\hat{\mat{U}}(\ell)$ do not converge towards a fixed set of vectors but track the time-varying principal components.
Since our objective is to recover the principal components associated with the terminal observation, we decompose the block sample covariance in terms of the last observation in each block.
For the $\ell$-th block, let $\mat{M}(\ell) := \Sigma_{\ell B}$ denote the population covariance at the block boundary:

\begin{equation}
\label{eqn:npmblock}
\frac{1}{B}\sum_{\mathclap{t=(\ell-1)B+1}}^{\ell B}
  \xxT{\vect{x}}{t}
= \underbrace{\mat{A}_{\ell B}\mat{A}_{\ell B}^\top
  + \sigma^2 \mat{I}}_{\mat{M}(\ell)}
+ \mathcal{E}(\ell)\,.
\end{equation}

The error matrix $\mathcal{E}(\ell)$ has a nonzero mean due to the within-block covariance drift.
In Lemma~\ref{lem:power-error}, we decompose $\lVert\mathcal{E}(\ell)\rVert$ into a stochastic term (decreasing in~$B$) and a drift term (increasing in~$B$).
The proof is given in Appendix~\ref{appendix:power-error-bound}.

\begin{lemma}[Block covariance concentration]
\label{lem:power-error}
Under model~\eqref{eqn:spike} with $(\mat{A}_t) \in \tempuncert$, the following holds with probability at least $1 - 1/T$:
\begin{equation}
\label{eq:npm-error-bound}
\max_{1 \leq \ell \leq L}
  \lVert\mathcal{E}(\ell)\rVert
  \leq C_{\mathrm{B}}
    \sqrt{\frac{\mathcal{V}\log(2pT^2)}{B}}
  + \frac{B\Gamma}{2}\,,
\end{equation}
provided $B \geq \mathcal{M}^2 \log(2pT^2)/\mathcal{V}$, where $\mathcal{V}
  \leq C_{\mathcal{V}}\,\kappa^4(\mu\delta+\sigma^2)
    (k\mu\delta+p\sigma^2)$ and $\mathcal{M}
  \leq \bigl(C_{\mathcal{M}}\,\kappa^2(k\mu\delta+p\sigma^2)
    + (\mu\delta+\sigma^2)\bigr)\log T$, where $C_{\mathcal{M}},C_{\mathcal{V}},C_{\mathrm{B}}>0$ are absolute matrix-concentration constants of comparable scale; $C_{\mathrm{B}} \leq 3.2$ (Remark~\ref{rmk:absolute-constants}).
\end{lemma}

The first term exhibits the classical inverse-square-root decay in the block size, while the second term grows linearly in~$B$ due to covariance drift.
When $\Gamma = 0$, a larger block is always better; when $\Gamma > 0$, an optimal block size exists that balances the two terms.

We establish the convergence guarantee of the noisy power method under drift in Theorem~\ref{thm:robust-power-method}.
In the proof (Appendix~\ref{appendix:robust-power-method}), we identify the optimal block size and bound the subspace distance between the output of Algorithm~\ref{algo:npm} and the column space of~$\mat{A}_T$.

\begin{theorem}[Noisy power method under drift]
\label{thm:robust-power-method}
Assume that $\delta \ge \tfrac{12}{17}\sigma^2$ and $\Gamma=\tilde{O}(\delta^3/((\mu\delta+\sigma^2)(p\sigma^2+k\mu\delta)))$.
For $B=$ $\tilde{\Theta}(\Gamma^{-2/3}\cdot(\mu\delta+\sigma^2)^{1/3}(p\sigma^2+k\mu\delta)^{1/3})$, we have:
\begin{equation}
d(\mathrm{ran}(\mat{A}_T), \mathrm{NPM}_\mathcal{X})=\tilde{O}\left( \Gamma^{1/3} \cdot \frac{(\mu\delta + \sigma^2)^{1/3} (p\sigma^2 + k\mu\delta)^{1/3}}{\delta} \!+\! 0.7^{{T}/{B}}\right)\,,
\end{equation}
with probability $1-2/T-e^{-\Omega(p)}$.
\end{theorem}
The $2/T$ collects one $1/T$ from the sub-exponential matrix Bernstein of Lemma~\ref{lem:power-error} and one from the random-initialization analysis of~\citep{npm} (Appendix~\ref{appendix:robust-power-method}).
The condition on $\Gamma$ ensures that the estimation error remains non-trivial: when $\Gamma$ exceeds the stated regime, the upper bound becomes $\Theta(1)$, which is vacuous since the subspace distance is at most~$1$.

Within this regime, the noisy power method achieves the minimax lower bound (Theorem~\ref{thm:lower_bound}) up to logarithmic factors under the conditions $p\sigma^2 \gtrsim k\delta$ and $\mu = \Theta(1)$:
\begin{equation}
\mathcal{R}^{\mathrm{NPM}}
  = \tilde{O}\!\left(
    \left(\frac{\Gamma}{\delta}\right)^{\!1/3}
    \left(\frac{p\sigma^2(\sigma^2+\delta)}
      {\delta^2}\right)^{\!1/3}\right),
\end{equation}
provided $T = \Omega(T_c)$ and $T^3\Gamma = \Omega(\delta^3/(p\sigma^2(\sigma^2+\delta)))$.
The first condition requires the time horizon to exceed the critical time~$T_c$ defined in~\eqref{eq:critical-time}, so that both the geometric decay term $0.7^{T/B}$ and the $2/T$-scale failure budget are dominated by the statistical error.
The second condition ensures the $1/T$-scale failure budget itself does not dominate the estimation error.
This conversion from high-probability bound to expectation is formalized as $\mathbb{E}[d] \leq \mathcal{R}^{\mathrm{NPM}}+\tilde{O}(1/T)$.
In the operating regime $T\ge T_c$, this condition is automatic (Appendix~\ref{appendix:npm-tail-to-expectation}).

\paragraph{Proof sketch.}
Write $\mat{U}_{1:k}(\ell)$ and $\mat{U}_{k+1:p}(\ell)$ for the leading $k$ and trailing $(p\!-\!k)$ singular vectors of $\mat{M}(\ell)$.
The standard convergence argument requires either $\|\mathcal{E}(\ell)\mat{U}(\ell)\|=O(\sqrt{k/p})$ on a \emph{fixed} subspace $\mat{U}$~\citep{npm} or a time-invariant covariance~\citep{HuangNW21}, both violated here: the per-block drift $\|\mat{M}(\ell)-\mat{M}(\ell\!-\!1)\|$ alone exceeds $\sqrt{k/p}$, so no per-iteration improvement toward a single subspace can be guaranteed.

We instead build per block two reference frames: a $k$-dimensional signal frame $\mat{W}^{(\ell)}$ aligned with $\mat{U}_{1:k}(\ell)$ (grown forward from a seed $\mat{W}^{(1)}$, reaching $\mat{W}^{(L+1)}$ after $L$ blocks) and a $(p\!-\!k)$-dimensional complement frame $\mat{N}^{(\ell)}$ aligned with $\mat{U}_{k+1:p}(\ell)$ (pulled backward).
The two frames anchor differently: the signal seed may be chosen freely (forward dynamics carry it onto $\mat{U}_{1:k}(L)$), but the complement must anchor at the terminal block, since $\mat{U}_{k+1:p}(L)$ is the only fixed handle on the trailing direction to avoid.
The intermediate $\mat{U}_{k+1:p}(\ell)$ themselves drift between blocks and are unknown a priori.
One multiplication by $\mat{M}(\ell)+\mathcal{E}(\ell)$ applies a \emph{floor} on the signal (its $k$-th singular value stays at least $s_k(\mat{M}(\ell))-O(\|\mathcal{E}(\ell)\|)$) and a \emph{ceiling} on the complement (capped at the noise level $\sigma^2+O(\|\mathcal{E}(\ell)\|)$).
Their ratio grows by a per-step factor $\beta\ge 1.4464>1$ (Appendix~\ref{appendix:amplification}, Lemma~\ref{lem:thm2ratio}), so the iterate, which at random start overlaps both frames, contracts onto $\mat{W}^{(L+1)}$ at rate $\beta^{-L}$.
The standing bound $\beta^{-1}<0.7$ converts this into $0.7^L$, the source of the $0.7^{T/B}$ tail in Theorem~\ref{thm:robust-power-method}; $\mat{W}^{(L+1)}$ stays $O(\|\mathcal{E}(L)\|)$ from the true subspace.

Between blocks the leading subspace $\mat{U}_{1:k}$ itself rotates by a subspace distance $\eta\approx B\Gamma/\delta$ from block $\ell{-}1$ to $\ell$ (Davis--Kahan applied to the per-block drift $B\Gamma$ against the gap $\delta$), reopening each frame's subspace-distance tolerance from $\epsilon$, the per-block target distance $d(\mat U_{1:k}(\ell), \mat W^{(\ell+1)})\le\epsilon$, back to $\epsilon{+}\eta$ at block $\ell{+}1$.
The next multiply re-tightens it.
Appendix~\ref{appendix:amplification} (Figure~\ref{fig:amplification} sketches the one-block flow) makes this precise: the per-vector floor and ceiling are Lemma~\ref{lem:floor_ceiling}, the $\beta\ge 1.4464$ inequalities are Lemma~\ref{lem:thm2ratio}, and the two frame constructions are Propositions~\ref{lem:proj_nonstat} and~\ref{lem:orth_nonstat}.
Chaining over blocks gives Theorem~\ref{thm:drifting-power-method}, which Theorems~\ref{thm:robust-power-method} and~\ref{thm:ojas-algorithm} instantiate.

\subsection{Oja's algorithm}
\label{sec:oja}
\begin{algorithm}[H]
\caption{Oja's algorithm with learning rate $\zeta$~\citep{oja_1}.}
\label{algo:oja}
\begin{algorithmic}[1]
\STATE \textbf{Input:} Stream of vectors: $(\vect{x}_{t})_{t=1}^T$, learning rate: $\zeta$, and dimensions: $p,k$
\STATE Sample each element of $\mat{\hat{U}}(0)$ in $\mathcal{N}(0,1)$
\label{algo2:rand_basis}\\
\FOR{$t=1:T$}
\label{algo2:outer_iter}
\STATE $\mat{\hat{U}}(t) \leftarrow \text{Gram-Schmidt}((\mat{I}_{p\times p}+\zeta\vect{x}_{t}\vect{x}^{\top}_{t})\mat{\hat{U}}(t-1))$
\label{algo2:output}
\ENDFOR
\STATE \textbf{Output:} $\mat{\hat{U}}(T)$
\end{algorithmic}
\end{algorithm}

We now establish the convergence guarantee for Oja's algorithm (Algorithm~\ref{algo:oja}).
Unlike the noisy power method, Oja's algorithm updates its estimate multiplicatively at every time step.
We analyze it by introducing a virtual block of size $B_\zeta = \zeta^{-1}$, following the approach of~\citep{jain2016streaming, HuangNW21, Allen2017}.
Using the approximation $(1 + \zeta x)^{1/\zeta} \approx e^{x}$, a product of $B_\zeta$ multiplicative updates exponentiates the spectrum of the within-block covariance.
This block product plays the role that the block average plays in the noisy power method.

The per-block concentration uses a matrix-product inequality~\citep[Cor.~6.1]{huang2020matrix}, which requires an almost-sure bound on $\lVert\vect{x}_t\vect{x}_t^\top\rVert$.
The sub-exponential matrix Bernstein behind Lemma~\ref{lem:power-error} did not need such a bound, so it is new to the Oja analysis.
We supply it by conditioning on the high-probability event
\[
\mathfrak{E}
  := \bigl\{\, |z_{t,i}| \leq \rho
    \;\;\forall\, t \in [T],\; i \in [p]
  \,\bigr\},
\qquad
\rho = \kappa\sqrt{2\log(2pT^2)},
\]
which satisfies $\mathbb{P}[\mathfrak{E}] \geq 1 - 1/T$ (Lemma~\ref{lem:trunc-prob}).
On~$\mathfrak{E}$, $\lVert\vect{x}_t\rVert^2 \leq p\rho^2\lVert\Sigma_t\rVert$ almost surely, which gives the uniform operator-norm bound $\mathcal{M}_\infty$ that the matrix-product inequality takes as input (Appendix~\ref{appendix:tail-bounds}).

For the $\ell$-th virtual block, we decompose
\begin{equation}
\label{eqn:ojablock}
\prod_{\mathclap{t=(\ell-1)B+1}}^{\ell B}
  (\mat{I} + \zeta\,\xxT{\vect{x}}{t})
= \underbrace{\prod_{\mathclap{t=(\ell-1)B+1}}^{\ell B}
  (\mat{I} + \zeta\,\Sigma_{\ell B})}_{
  \mat{M}^{\mathrm{Oja}}(\ell)}
+ \mathcal{E}(\ell)
= \mat{M}^{\mathrm{Oja}}(\ell)
  + e^{\mu\delta+\sigma^2}\,\mathcal{E}'(\ell),
\end{equation}
where $\mathcal{E}'(\ell)$ is the scaled error matrix.
In Lemma~\ref{lem:oja-error}, we bound the scaled error in terms of the learning rate~$\zeta$ (or equivalently the virtual block size $B_\zeta = \zeta^{-1}$).
The proof is given in Appendix~\ref{appendix:oja-error-bound}.

\begin{lemma}[Error bound for Oja's algorithm]
\label{lem:oja-error}
Under model~\eqref{eqn:spike} with $(\mat{A}_t) \in \tempuncert$, the following holds with probability at least $1 - 2/T$:
\begin{equation}
\label{eq:oja-error-main}
\max_{1 \leq \ell \leq L}
  \lVert\mathcal{E}'(\ell)\rVert
  \leq \sqrt{2}e\mathcal{M}_\infty
    \sqrt{\frac{\log(pT^2)}{B_\zeta}}
  + \frac{B_\zeta \Gamma}{2}
  + \epsilon_{B_\zeta,\Gamma}\,,
\end{equation}
when $B_\zeta \geq 2\mathcal{M}_\infty^2\log(pT^2)$, where $\mathcal{M}_\infty$ is the a.s.\ operator-norm bound on the truncation event from Appendix~\ref{appendix:tail-bounds} and $\epsilon_{B_\zeta,\Gamma} = (B_\zeta\Gamma)^2/(4 - 2B_\zeta\Gamma)$ is negligible when $B_\zeta\Gamma \ll 1$.
\end{lemma}

The appearance of $\mathcal{M}_\infty$ rather than $\mathcal{V}$ in the stochastic term reflects the use of a different concentration tool.
The noisy power method averages outer products, which is amenable to the sub-exponential matrix Bernstein inequality (exploiting the matrix variance~$\mathcal{V}$ and the $\psi_1$ boundedness~$\mathcal{M}$).
Oja's algorithm instead takes a product of random matrices, which requires the multiplicative concentration inequality of~\citep{huang2020matrix}.
That inequality controls the product through the a.s.\ bound~$\mathcal{M}_\infty$ rather than through distributional quantities.

Combining Lemma~\ref{lem:oja-error} with the iterative framework developed for the noisy power method yields the following convergence guarantee.
The complete proof is given in Appendix~\ref{appendix:oja-algorithm-convergence}.

\begin{theorem}[Oja's algorithm]
\label{thm:ojas-algorithm}
Assume that $\mu = \Theta(1)$, $\delta,\sigma^2 = O(1)$, $\delta > \log(29/17)$, and $\Gamma = \tilde{O}(\delta^3 / (p(\mu\delta+\sigma^2))^2)$.
For $\zeta^{-1} = \tilde{\Theta}(\Gamma^{-2/3}\, (p(\mu\delta+\sigma^2))^{2/3})$, we have:
\begin{equation}
d(\mathrm{ran}(\mat{A}_T),\, \mathrm{Oja}_\mathcal{X})
  = \tilde{O}\!\left(
    \frac{(p(\mu\delta+\sigma^2))^{2/3}\,
      \Gamma^{1/3}}{\delta}
    + 0.7^{\zeta T}\right),
\end{equation}
with probability $1 - 3/T - e^{-\Omega(p)}$.
\end{theorem}
The $3/T$ collects one $1/T$ each from the truncation event $\mathfrak{E}$ (Appendix~\ref{appendix:tail-bounds}), the matrix-product concentration of Lemma~\ref{lem:oja-error}, and the random-initialization analysis of~\citep{npm}.
The accompanying $e^{-\Omega(p)}$ is the random-initialization failure probability of that same analysis.

Unlike the noisy power method, the upper bound in Theorem~\ref{thm:ojas-algorithm} scales as $\tilde{O}(p^{2/3})$ rather than the optimal $\tilde{O}(p^{1/3})$ from Theorem~\ref{thm:lower_bound}.
This gap arises because the multiplicative concentration inequality~\citep{huang2020matrix} depends on the a.s.\ bound~$\mathcal{M}_\infty$ rather than the matrix variance~$\mathcal{V}$.
The two scalings differ by a factor of~$p$: $\mathcal{M}_\infty \sim p(\mu\delta+\sigma^2)$ has the ambient dimension as a prefactor, whereas $\mathcal{V} \sim (\mu\delta+\sigma^2)(k\mu\delta+p\sigma^2)$ does not.
In the stationary setting this factor of $p$ is avoidable: the sharp finite-sample analysis of Oja's algorithm attains the matrix Bernstein rate through a variance proxy~\citep{jain2016streaming, Allen2017, HuangNW21}, but it relies on convergence to a fixed target subspace, which drift removes.
Whether the variance rate survives under drift is left open (Appendix~\ref{appendix:oja-error-bound}).

\section{Numerical results}
\label{numerical}
Theorems~\ref{thm:robust-power-method} and~\ref{thm:ojas-algorithm} make two predictions.
\textbf{(i)}~When $\Gamma > 0$, there exists an optimal block size $B$ (resp.\ learning rate $\zeta$) that minimizes the recovery error by balancing the stochastic and drift terms.
\textbf{(ii)}~These optimal parameters scale as $B_{\mathrm{opt}}, \zeta^{-1}_{\mathrm{opt}} \propto \Gamma^{-2/3}$.
The $\Gamma = 0$ case is a boundary: no drift term is present, so the error decreases monotonically as $B$ increases (equivalently, as $\zeta$ decreases).
This section validates these predictions on both synthetic and real-world data.

\subsection{Synthetic experiments}

\begin{figure}[H]
\begin{subfigure}{0.495\columnwidth}
\centering
\includegraphics[width=0.99\columnwidth]{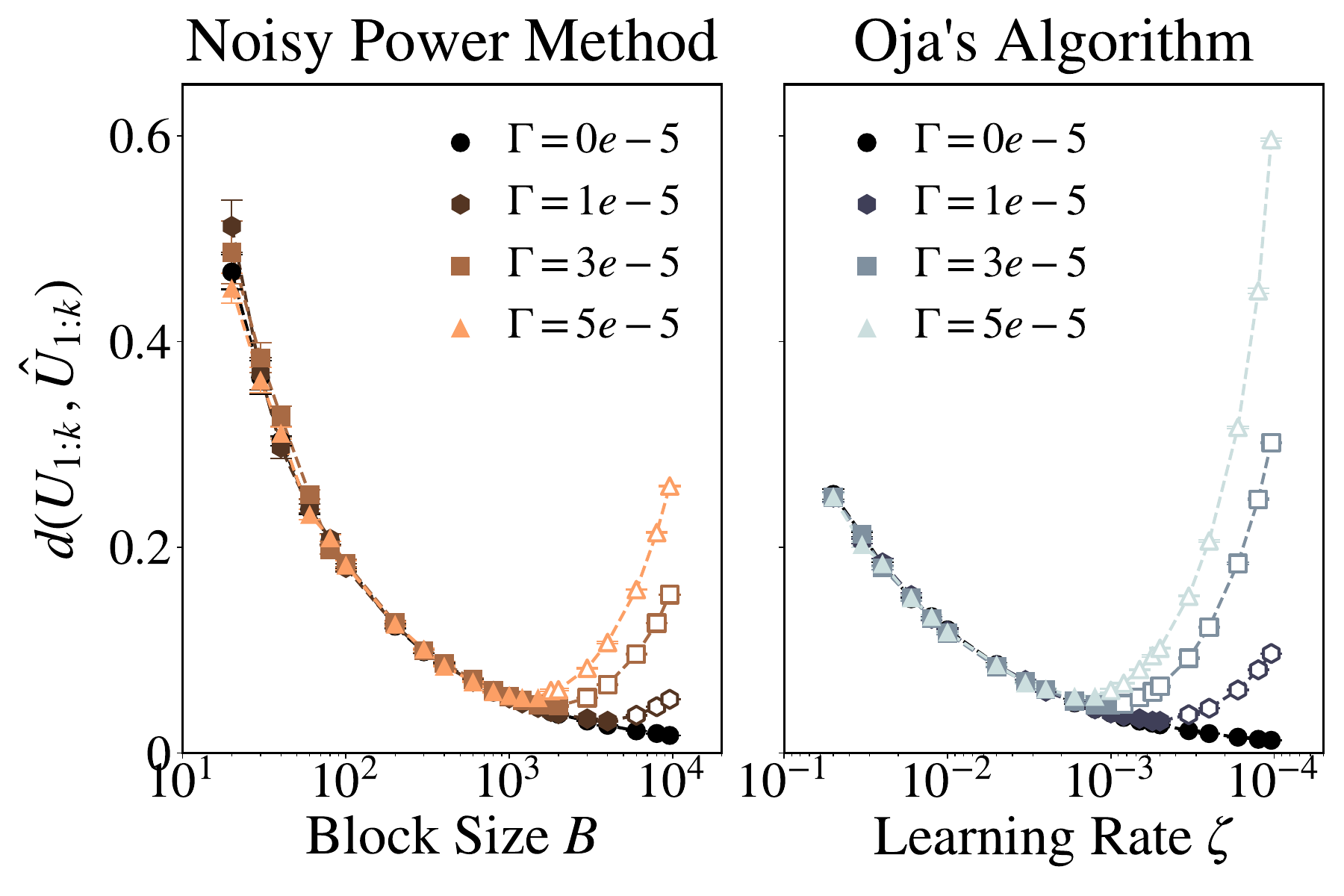}
\caption{Recovery error $d(\mat{U}_{1:k}, \hat{\mat{U}}_{1:k})$ at $t = T$ for varying $\Gamma$ and learning parameters.}
\label{fig:adversarial-gamma}
\end{subfigure}
\begin{subfigure}{0.495\columnwidth}
\centering
\includegraphics[width=0.99\columnwidth]{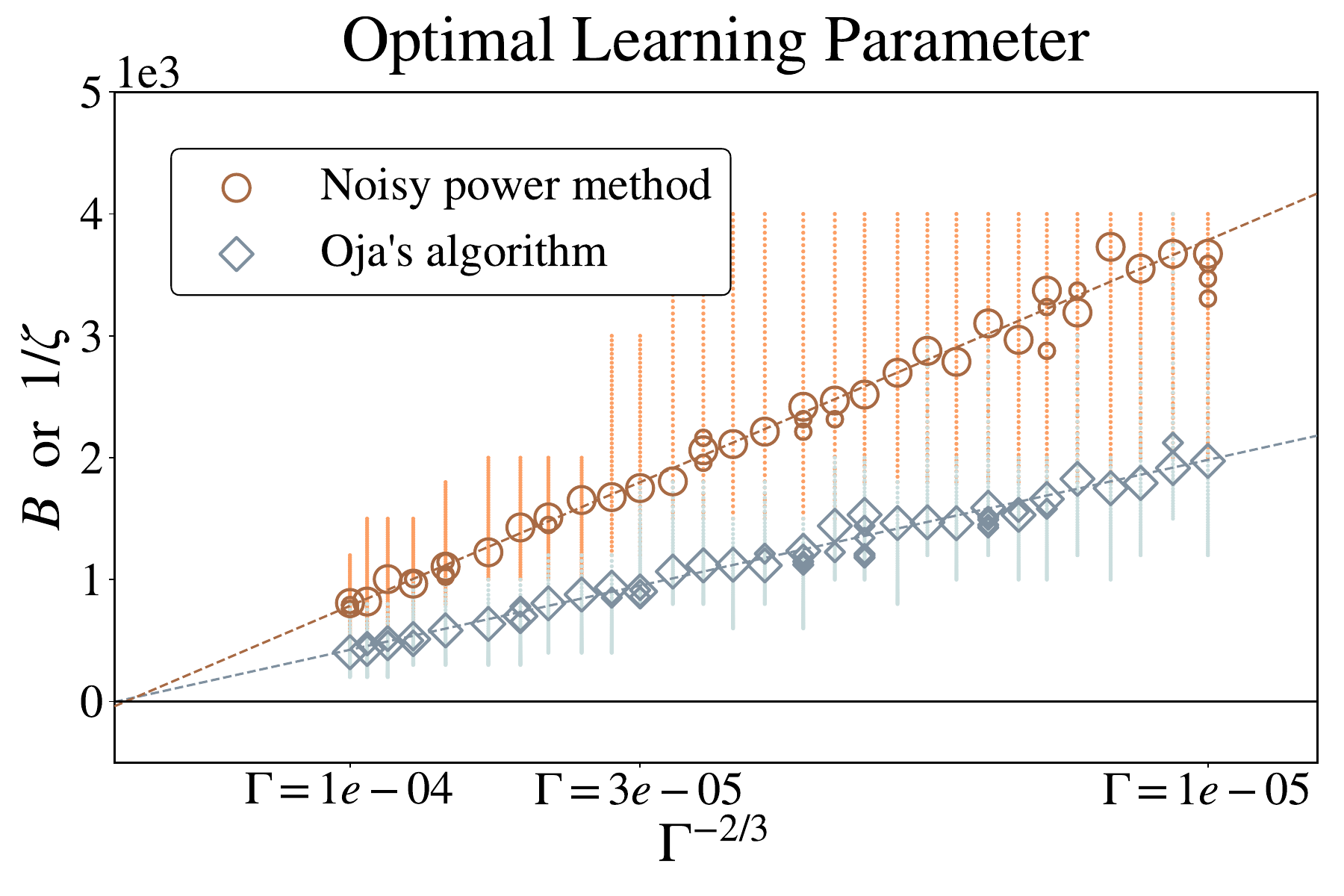}
\caption{Empirical optimal $B$ and $\zeta^{-1}$ versus $\Gamma$, plotted against the theoretical $\Gamma^{-2/3}$ scaling.}
\label{fig:optimal_lp}
\end{subfigure}
\caption{Synthetic experiments with $(\sigma, \delta, p, k) = (0.15, 1.0, 100, 5)$.}
\label{fig:synthetic_main}
\end{figure}

We generate $(\mat{A}_t)_{t=1}^{T} \in \mathbb{R}^{p \times k}$ as the product $\mat{U}_t \mat{D}_t \mat{V}_t^\top$, where $\mat{U}_t \in \stiefel{p}{p}$, $\mat{D}_t \in \mathbb{R}^{p \times k}$ is diagonal, and $\mat{V}_t \in \stiefel{k}{k}$.
To simulate non-stationarity, we rotate $\mat{U}_t = \mat{U}_{t-1}\mat{R}_t$ with $\mat{R}_t \in \mathrm{SO}(p)$, controlling $\Gamma$ through the rotation magnitude.
The observations $\vect{x}_t$ are sampled from model~\eqref{eqn:spike} with Gaussian noise, i.e., $\vect{z}_t \sim \mathcal{N}(\vect{0}, \mat{I}_p)$ ($\kappa = 1$).
Each curve in Figure~\ref{fig:adversarial-gamma} shows the mean recovery error across $10$ independent runs.
Each run draws a fresh seed for the signal sequence $(\mat{A}_t)$, the noise sequence $(\vect{z}_t)_{t=1}^T$, and the random initialization $\hat{\mat{U}}(0)$.
Error bars show one standard error of the mean, i.e., the across-run sample standard deviation divided by $\sqrt{10}$.
Further details and an additional perturbation regime are in Appendix~\ref{appendix:experimental-details}.

\vspace{-8pt}

Figure~\ref{fig:adversarial-gamma} confirms the trade-off between the stochastic and drift terms predicted by Lemma~\ref{lem:power-error} and~\ref{lem:oja-error}.
When $\Gamma = 0$, the recovery error decreases monotonically with the block size, since more observations reduce the sampling noise without introducing any drift penalty.
When $\Gamma > 0$, U-shaped curves emerge: small block sizes yield high stochastic error, while large block sizes incur excessive drift.
The location of the minimum shifts leftward as $\Gamma$ increases: stronger perturbations render past observations stale more quickly.
Figure~\ref{fig:optimal_lp} tests the $\Gamma^{-2/3}$ scaling prediction directly, plotting the empirically optimal block size $B$ and inverse learning rate $\zeta^{-1}$ against the $\Gamma^{-2/3}$ reference curve; both track it over the tested range of $\Gamma$, and differ from each other only by a multiplicative constant.

\vspace{-4pt}

\subsection{Experiments on stock price dataset}

We evaluate both algorithms on non-stationary S\&P500 stock market daily returns~\citep{spdataset}, which exhibit heavier-than-Gaussian tails~\citep{Cont01022001}, taking us outside the sub-Gaussian model.

\vspace{-6pt}

\begin{figure}[H]
\centering
\includegraphics[width=.90\textwidth]{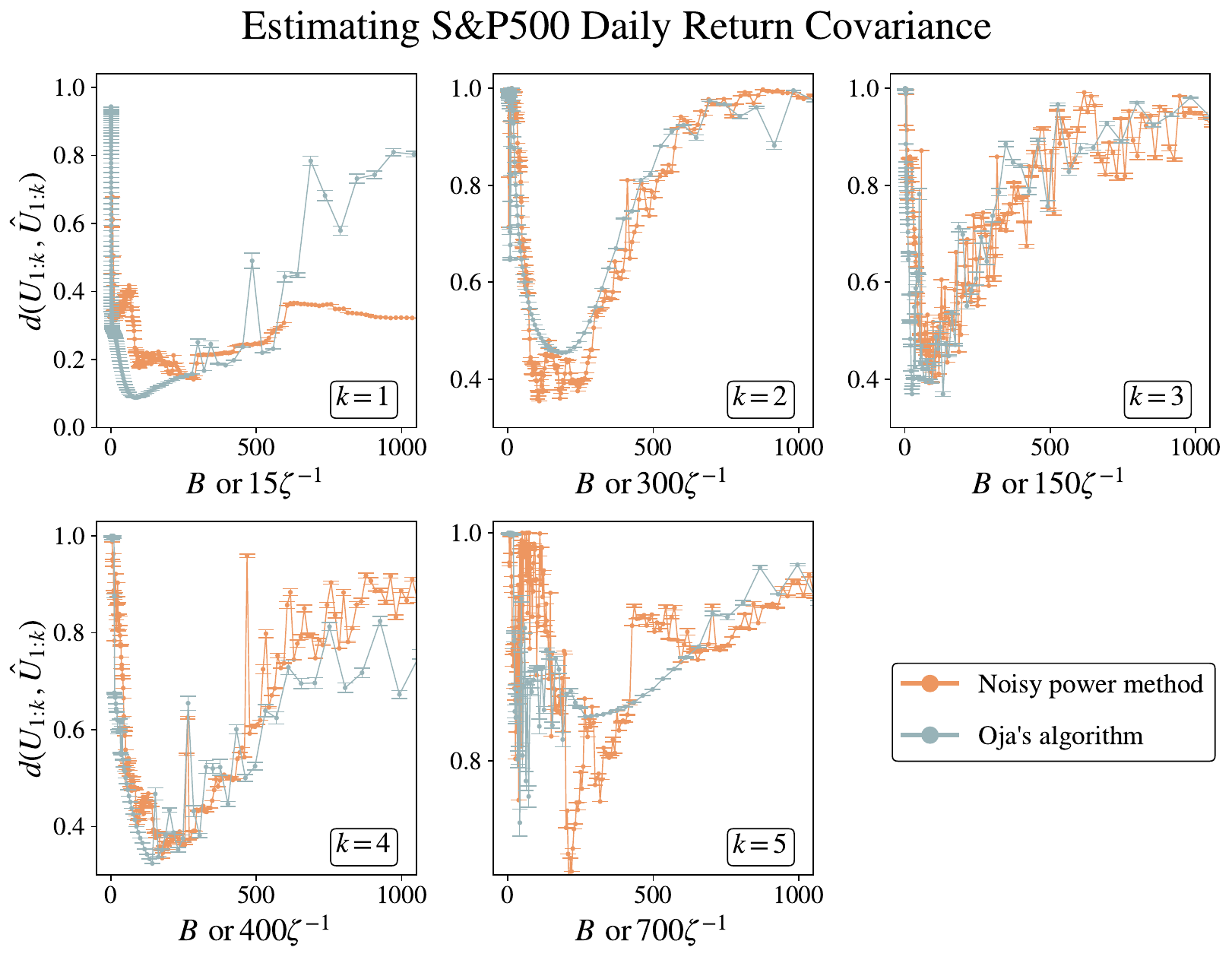}
\caption{Recovery error against the top-$k$ singular vectors of the sample covariance on S\&P500 daily returns. The horizontal axis shows the block size $B$ for the noisy power method and a $k$-rescaled inverse learning rate $\zeta^{-1}$ for Oja's algorithm; the rescaling collapses the two curves per $k$.}
\label{fig:sp500_result}
\end{figure}

\vspace{-5pt}

The data consist of normalized daily returns of $137$ ($=p$) companies from March~18, 1980 to July~22, 2022 ($T = 10{,}677$).
The resulting stream exhibits time-varying covariance, with a drift budget of order $10^{-4}$ read at the shortest window the estimator resolves and a spectral gap that satisfies $\specgap > \varbudget$ for every $k$ we run (Appendix~\ref{appendix:real-budget}).

The task is to estimate the top-$k$ principal components of the terminal covariance matrix of daily returns, following the portfolio analysis framework of~\citep{Yang2015AnAO, Fatima2019, Souma2021}.
As a reference target for the recovery error, we use the top-$k$ singular vectors of the sample covariance computed from the final $500$ observations ($k = 1, \ldots, 5$); Appendix~\ref{appendix:real-detail} gives the run count, the parameter ranges and the error bars.

Figure~\ref{fig:sp500_result} confirms that the theoretical phenomena persist in a real-world, heavy-tailed environment.
First, U-shaped curves appear for every~$k$, confirming the existence of an optimal learning parameter even when the data distribution departs from the sub-Gaussian model.
Second, the noisy power method and Oja's algorithm trace nearly identical error curves when $\zeta^{-1}$ is rescaled by a $k$-dependent constant, indicating that the two algorithms share the same trade-off structure.
Third, the rescaling constant varies across~$k$ because the spectral gap $\delta$ between the $k$-th and $(k+1)$-th eigenvalues differs for each~$k$.
A smaller spectral gap increases the sensitivity to perturbations.
This both shifts the optimal parameter and raises the minimum achievable error.

\section{Conclusion}
\label{conclusion}
For streaming PCA under a time-varying covariance, we established the minimax recovery error over the temporal uncertainty set $\tempuncert$ (Theorem~\ref{thm:lower_bound}). The bound separates two regimes. Below the critical time $T_c$ of~\eqref{eq:critical-time} the error falls at the parametric rate $p^{1/2}T^{\shortminus 1/2}$; above it the error plateaus at $O(p^{1/3}\varbudget^{1/3})$, since observations go stale as fast as they arrive. The noisy power method attains this rate up to logarithmic factors when the spike is well conditioned, $\mu=\Theta(1)$, and the noise energy dominates, $p\sigma^2\gtrsim k\delta$. This makes it rate-optimal in the sense of Definition~\ref{defn:algo_class}. Oja's algorithm shares the $\varbudget^{\shortminus 2/3}$ scaling of the optimal learning parameter, but its error has an extra factor $p^{1/3}$ inherited from the matrix-product concentration its analysis requires (Section~\ref{sec:oja}). Both predictions hold empirically (Section~\ref{numerical}), on synthetic streams and on S\&P500 daily returns, whose tails are heavier than Gaussian~\citep{Cont01022001}.

The condition number $\mu$ enters our upper bound but not the lower bound, and the lower-bound side provably cannot supply it; whether the residual slack is an artifact of our operator-norm analysis or is genuine is open (Appendix~\ref{appendix:mu-gap}).
The extra $p^{1/3}$ in Oja's rate would disappear if the variance-based analysis of the stationary setting survived drift, which we cannot establish because that analysis converges to a fixed target (Appendix~\ref{appendix:oja-error-bound}).
The model admits two further relaxations: the spectral-gap assumption $s_k(\xxT{\mat{A}}{t})\ge\delta$ could give way to an effective-rank condition~\citep{EFFRANK,EFFRANK2}, and the block size and learning rate, set from $\varbudget$ here, would have to be estimated online.

\begin{ack}
DB and AS were supported by a DARPA Lagrange award. MJ and SY were supported by Institute of Information \& Communications Technology Planning \& Evaluation (IITP) grant funded by the Korea government (MSIT) (No.2022-0-00311, Development of Goal-Oriented Reinforcement Learning Techniques for Contact-Rich Robotic Manipulation of Everyday Objects; No.2019-0-00075, Artificial Intelligence Graduate School Program (KAIST)).
\end{ack}

\clearpage
\bibliography{ref}
\bibliographystyle{plainnat}

\clearpage
\appendix

\newpage
\addtocontents{toc}{\protect\StartAppendixEntries}
\setcounter{tocdepth}{2}
\listofatoc

\clearpage

\section{Notation table}
\label{appendix:notation-table}
\begin{table}[h]
\caption{Table of notations throughout the appendix. Some notations defined in Section~\ref{introduction} are omitted. }
{\small
\renewcommand{\arraystretch}{0.95}
\setlength{\tabcolsep}{4pt}
\begin{tabularx}{\textwidth}{p{0.22\textwidth}X}
      \toprule
      {Model and problem parameters:} \\
      $T$ & time horizon length\\
      $k$ & number of principal components\\
      $p$ & dimension of observation vectors\\
      $\sigma$ & magnitude of observation noise \\
      $\Sigma_t = \xxT{\mat{A}}{t}+\sigma^2\mat{I}_{p\times p}$ & covariance matrix at the time $t$\\
      $\delta$ & lower bound on $s_k(\xxT{\mat{A}}{t})$\\
      $\mu$ & condition-number parameter of $\tempuncert$ (Definition~\ref{defn:tempuncert}); $\mu\ge 1$, with $s_1(\xxT{\mat{A}}{t})\le\mu\delta$ uniformly in $t$\\
      $\Gamma$ & per-step drift budget, $\Vert \xxT{\mat{A}}{t} - \xxT{\mat{A}}{t-1} \Vert\le\Gamma$\\
      \midrule
      {Algorithm parameters:} \\
      $B$ & block size for the noisy power method\\
      $\zeta$ & learning rate for Oja's algorithm\\
      $B_\zeta\,(=\zeta^{-1})$ & virtual block size for Oja's algorithm\\
      $\zeta_{\mathrm{opt}}\,,B_{\mathrm{opt}}$ & optimal learning parameter under covariance shifts ($\tempuncert$)\\
      $L$ & iteration count; $L\simeq T/B$ for the noisy power method, $L\simeq T/B_\zeta=\zeta T$ for Oja\\
      \midrule
      {High-probability event and concentration constants:} \\
      $\mathfrak{E}$ & truncation event $\{\max_{t,i}|z_{t,i}|\le\rho\}$, $\mathbb{P}[\mathfrak{E}]\geq 1-1/T$ (Oja-error analysis only)\\
      $\rho$ & truncation threshold $\kappa\sqrt{2\log(2pT^2)}$ (Oja-error analysis only)\\
      $\nu$ & truncation bias, $1-\mathbb{E}[z_{t,i}^2\mid|z_{t,i}|\le\rho]=\tilde{O}(1/(pT^2))$ (Oja-error analysis only)\\
      $\Delta_t$ & off-diagonal residual in $\mathbb{E}[\xxT{\vect{x}}{t}\mid\mathfrak{E}]=(1-\nu)\Sigma_t+\Delta_t$, $\lVert\Delta_t\rVert=\tilde{O}(1/(pT^2))$ (Oja-error analysis only)\\
      $\mathcal{M}\,,\mathcal{V}$ & $\psi_1$-boundedness and variance proxies for the sub-exponential matrix Bernstein (power-error analysis)\\
      $\mathcal{M}_\infty$ & a.s.\ spectral-norm bound on $\mathfrak{E}$, for the matrix-product concentration (Oja-error analysis)\\
      $C_{\mathrm{q}},C_4,C_4',C_{\mathcal{M}},C_{\mathcal{V}},C_{\mathrm{B}}$ & absolute constants in $\mathcal{M},\mathcal{V}$ and the matrix-Bernstein quadratic-solve (Remark~\ref{rmk:absolute-constants})\\
      $\tau_0$ & random-init scale ($\tau_0\ge 1$) in $R\le\tau_0\sqrt{pk}$~\citep{npm}; set to $\tau_0=T$ in Theorems~\ref{thm:robust-power-method},~\ref{thm:ojas-algorithm}\\
      \midrule
      {Subspace geometry:} \\
      $\stiefel{k}{p}$   & Stiefel manifold, matrices $\mat{M}\in\mathbb{R}^{p\times k}$ with $\mat{M}^\top \mat{M}=\mat{I}_{k\times k}$\\
      $\grass{k}{p}$   & Grassmann manifold of $k$-dimensional subspaces of $\mathbb{R}^p$\\
      $[\mat{M}]\in\grass{k}{p}$ & subspace spanned by the columns of $\mat{M}\in\stiefel{k}{p}$\\
      $\mathcal{G}_{[\mat{M}]\rightarrow [\mat{N}]}(\Psi')$ & principal rotation from $[\mat{M}]$ to $[\mat{N}]$\\
      $s$ & separation radius of the lower-bound packing\\
      \bottomrule
\end{tabularx}
}
\end{table}

\section{Technical preliminaries}
\subsection{Auxiliary lemmas}
\begin{lemma}[Theorem~2.5.1, \citep{golub2013matrix}]
\label{lem:proj_nonstat.6.1_golub}
Let $\mathcal{S}_{1}$ and $\mathcal{S}_{2}$ be two subspaces of $\mathbb{R}^{p}$, such that $\mathrm{dim}(\mathcal{S}_{1})$=$\mathrm{dim}(\mathcal{S}_{2})$.
We define the distance between these two subspaces ($\mathcal{S}_{1}$,$\mathcal{S}_{2}$) by $\Vert \mat{P}_{1}-\mat{P}_{2}\Vert$, where $\mat{P}_{i}, i=1,2$ is the orthogonal projection onto $\mathcal{S}_{i}\:(i=1,2)$.
Suppose further that $\mat{M}=[\underset{k}{\mat{M}_{1}}\ \underset{p-k}{\mat{M}_{2}}]$ and $\mat{N}=[\underset{k}{\mat{N}_{1}}\ \underset{p-k}{\mat{N}_{2}}]$ are $p \times p$ orthogonal matrices.
If $\mathcal{S}_{1}=\mathrm{ran}(\mat{M}_{1})$ and $\mathcal{S}_{2}=\mathrm{ran}(\mat{N}_{1})$, then:
\[
\mathrm{dist}(\mathcal{S}_{1},\mathcal{S}_{2})=\Vert\mat{M}^{\top}_{1}\mat{N}_{2}\Vert=\Vert\mat{M}^{\top}_{2}\mat{N}_{1}\Vert\,.
\]
\end{lemma}
\begin{lemma}[Davis-Kahan $\sin(\theta)$ theorem, \citep{DavisKahan1970}; statement from Theorem VII.3.1, \citep{bhatia}]
\label{lem:dk_sin}
For given symmetric matrices $\mat{M}, \mat{N}$ with singular value decomposition $\mathrm{SVD}(\mat{M})=\mat{U}\mat{D}\mat{U}^\top$ and $\mathrm{SVD}(\mat{M}+\mat{N})=\hat{\mat{U}}\hat{\mat{D}}\hat{\mat{U}}^\top$, we have:
\[
\Vert \mat{U}_{1:k}\mat{U}^{\top}_{1:k}-\mat{\hat{U}}_{1:k}\mat{\hat{U}}^{\top}_{1:k} \Vert \le \frac{\Vert \mat{N} \Vert}{s_{k}(\mat{M})-s_{k+1}(\mat{M})-\Vert \mat{N} \Vert}\,.
\]
\end{lemma}
\begin{lemma}[Weyl's theorem]
\label{weyl}
For any $\mat{M},\mat{N} \in \mathbb{R}^{m \times n}$ and $1\leq i \leq \min(m,n)$,
\[
s_{i}(\mat{M}+\mat{N}) \le s_{i}(\mat{M})+s_{1}(\mat{N})\,.
\]
\end{lemma}
\begin{lemma}[Sub-additivity of rank]
\label{lem:subaddrank}
For any $\mat{M},\mat{N} \in \mathbb{R}^{m \times n}$,
\[
\mathrm{rk}(\mat{M}+\mat{N})\leq \mathrm{rk}(\mat{M}) + \mathrm{rk}(\mat{N})\,.
\]
\end{lemma}

\begin{lemma}[KL divergence between spiked Gaussians]
\label{lem:spiked-kl}
Let $\Sigma^{(i)} = \sigma^2\mat{I}_p + \mat{U}^{(i)} D (\mat{U}^{(i)})^\top$ and $\Sigma^{(j)} = \sigma^2\mat{I}_p + \mat{U}^{(j)} D (\mat{U}^{(j)})^\top$ with $\mat{U}^{(i)}, \mat{U}^{(j)} \in \stiefel{k}{p}$ and a shared diagonal spectrum $D = \mathrm{diag}(d_1, \dots, d_k) \succ 0$.
Writing $c(d) := d/(\sigma^2+d)$ and $V := (\mat{U}^{(j)})^\top \mat{U}^{(i)}$,
\[
\kl{\mathcal{N}(\bm{0}, \Sigma^{(i)})}{\mathcal{N}(\bm{0}, \Sigma^{(j)})}
  = \frac{1}{2\sigma^2}\sum_{a=1}^{k} c(d_a)\bigl[d_a - (V D V^\top)_{aa}\bigr].
\]
In the isotropic case $D = \delta\mat{I}_k$ this reduces to
\[
\kl{\mathcal{N}(\bm{0}, \Sigma^{(i)})}{\mathcal{N}(\bm{0}, \Sigma^{(j)})}
  = \frac{\delta^2}{4\sigma^2(\sigma^2+\delta)}\,\frob{\mat{U}^{(i)} (\mat{U}^{(i)})^\top - \mat{U}^{(j)} (\mat{U}^{(j)})^\top}^2.
\]
\end{lemma}
\begin{proof}
The two covariances share the eigenvalues $\{\sigma^2 + d_a\}_{a=1}^k \cup \{\sigma^2\}$, so $\log(|\Sigma^{(j)}| / |\Sigma^{(i)}|) = 0$ and $\kl{\cdot}{\cdot} = \tfrac{1}{2}\bigl(\tr((\Sigma^{(j)})^{-1}\Sigma^{(i)}) - p\bigr)$.
The Woodbury identity gives $(\Sigma^{(j)})^{-1} = \sigma^{-2}\bigl(\mat{I} - \mat{U}^{(j)} \mathrm{diag}(c(d_a)) (\mat{U}^{(j)})^\top\bigr)$; expanding $\tr((\Sigma^{(j)})^{-1}\Sigma^{(i)}) - p$ and using $d_a - \sigma^2 c(d_a) = c(d_a)\, d_a$ to combine terms leaves the stated sum.
For $D = \delta\mat{I}_k$, $c(d_a) = c(\delta)$ and $(V D V^\top)_{aa} = \delta (V V^\top)_{aa}$, so the divergence equals $\tfrac{\delta\, c(\delta)}{2\sigma^2}\bigl(k - \frob{V}^2\bigr)$; substituting $c(\delta) = \delta/(\sigma^2+\delta)$ and $\frob{\mat{U}^{(i)} (\mat{U}^{(i)})^\top - \mat{U}^{(j)} (\mat{U}^{(j)})^\top}^2 = 2k - 2\frob{V}^2$ gives the isotropic form.
\end{proof}

\subsection{Grassmann manifold}
To rigorously handle the $k$-dimensional subspaces, we adopt the geometric framework of the Grassmannian manifold~\citep{dai2005,conway1996,neretin2001,mandolesi2021grassmann,bendokat2020grassmann}.

\begin{definition}[Grassmannian Manifold]
The Grassmannian manifold $\grass{k}{p}$ is the set of all $k$-dimensional linear subspaces of $\mathbb{R}^p$.
It forms a compact Riemannian manifold of dimension $k(p-k)$.
Elements of $\grass{k}{p}$ are typically represented as equivalence classes $[\mat{M}]$, where $\mat{M} \in \stiefel{k}{p}$ is an orthonormal matrix (i.e., $\mat{M}^\top \mat{M} = \mat{I}_k$).
The equivalence relation $\sim$ for equivalence $\grass{k}{p} \approx \stiefel{k}{p} / \sim $ is defined by the equality of projection matrices:
\begin{equation*}
    \mat{M}_1 \sim \mat{M}_2 \iff \mat{M}_1\mat{M}_1^\top = \mat{M}_2\mat{M}_2^\top, \quad \text{for } \mat{M}_1, \mat{M}_2 \in \stiefel{k}{p}.
\end{equation*}
\end{definition}

\begin{definition}[Principal Angles]
\label{def:principal_angles}
The geometric relationship between two subspaces $[\mat{M}], [\mat{N}] \in \grass{k}{p}$ is fully characterized by their $k$ principal angles $\Psi = (\psi_1, \dots, \psi_k) \in [0, \pi/2]^k$.
These angles are defined via the singular value decomposition (SVD) of the inner product matrix:
\begin{equation*}
    \mat{M}^\top \mat{N} = \mat{U} \mat{D} \mat{V}^\top, \quad \text{where } \mat{D} = \diag(\cos \psi_1, \dots, \cos \psi_k).
\end{equation*}
The set of principal angles is invariant under the choice of realization $\mat{M}$ and $\mat{N}$ for $[\mat{M}]$ and $[\mat{N}]$.
\end{definition}

To quantify the divergence between subspaces, we use the \textit{projection 2-distance}, which corresponds to the spectral norm of the difference between projection matrices.

\begin{definition}[Projection 2-distance]
\label{def:projection_dist}
For any $[\mat{U}], [\mat{V}] \in \grass{k}{p}$ with principal angles $\Psi$, the projection 2-distance is defined as:
\begin{equation*}
    \proje{[\mat{U}]}{[\mat{V}]} 
    \defeq \oper{\mat{U}\mat{U}^\top - \mat{V}\mat{V}^\top} 
    = \norm{\sin \Psi}_{\infty} 
    = \max_{1\leq i \leq k} \sin \psi_i.
\end{equation*}
This metric relates to the chordal distance $\chord{[\mat{U}]}{[\mat{V}]} \defeq \frob{\sin \Psi}$ (which sums all principal angles), but we use projection 2-distance throughout to focus on the worst-case angular deviation.
\end{definition}

Finally, we establish a volume bound for metric balls on the Grassmannian.
This result is adapted from the chordal metric bounds in \citet[Theorem~1]{dai2005}.
In finite dimensions, all norms are equivalent, so the projection 2-distance ball $\mathcal{B}_{d_2}(r)$ is contained within and contains chordal metric balls with comparable radii.
Consequently, the volume scaling laws transfer: for small radii $r$, the volume scales with the manifold dimension $k(p-k)$.

\begin{proposition}[Volume of $d_2$-balls]
\label{prop:grassvolume}
Let $\mathcal{B}([\mat{M}], r) \subset \grass{k}{p}$ be a ball of radius $r \in (0, 1)$ centered at $[\mat{M}]$ with respect to the projection 2-distance.
There exists a constant $c_{p,k}$ such that the volume $\mu(\mathcal{B}([\mat{M}], r))$ satisfies:
\begin{equation*}
    c_{p,k} r^{k(p-k)} \leq \mu(\mathcal{B}([\mat{M}], r)) \leq \frac{c_{p,k} r^{k(p-k)}}{(1-r^2)^{k/2}}.
\end{equation*}
\end{proposition}

Rotation between two vectors is elementary, but characterizing the rotation between $k$-dimensional subspaces requires considering multiple angles simultaneously.
To formally describe this, we use the principal angles (Definition~\ref{def:principal_angles}).
Let $\Psi=\diag(\psi_i)_{i=1}^k$ denote the principal angles between $[\mat{M}]$ and $[\mat{N}]$ (where $\mat{M}, \mat{N} \in \stiefel{k}{p}$).
The $\mathcal{G}$-mapping defined below generalizes standard rotation to the Grassmannian manifold.

\begin{definition}[Principal Rotation]
\label{def:principal_rotation}
Let $[\mat{M}], [\mat{N}] \in \grass{k}{p}$ be two subspaces with principal angles $\Psi = \diag(\psi_i) \in [0, \pi/2]^k$.
We define the \textbf{principal rotation mapping}
\begin{equation*}
    \mathcal{G}_{[\mat{M}]\to[\mat{N}]} : \prod_{i=1}^k [0, \psi_i] \to \grass{k}{p}
\end{equation*}
as a function of an angle vector $\Theta \in \prod_{i=1}^k [0, \psi_i]$ satisfying:
\begin{enumerate}
    \item \textbf{Boundary Conditions:} $\mathcal{G}(\mathbf{0}) = [\mat{M}]$ and $\mathcal{G}(\Psi) = [\mat{N}]$.
    \item \textbf{Geodesic Property:} For any two intermediate angle configurations $\mathbf{0} \preceq \Theta_1, \Theta_2 \preceq \Psi$, the principal angles between $\mathcal{G}(\Theta_1)$ and $\mathcal{G}(\Theta_2)$ are exactly $|\Theta_1 - \Theta_2|$.
\end{enumerate}
Consequently, the projection 2-distance satisfies:
\begin{equation*}
    \proje{\mathcal{G}(\Theta_1)}{\mathcal{G}(\Theta_2)} = \norm{\sin(\Theta_1 - \Theta_2)}_\infty.
\end{equation*}
\end{definition}

\begin{proof}
Consider the singular value decomposition $\mat{M}^\top \mat{N} = \mat{U} \mat{\Sigma} \mat{V}^\top$, which implies $(\mat{M}\mat{U})^\top (\mat{N}\mat{V}) = \mat{\Sigma} = \cos\Psi$, where $0 \preceq \Psi \preceq \pi/2$.
Here $[\mat{M}\mat{U}] = [\mat{M}]$ and $[\mat{N}\mat{V}] = [\mat{N}]$ by the unitary invariance of the subspace.

We explicitly construct the mapping as:
\begin{equation*}
    \mathcal{G}(\Theta) = \left[ (\mat{M}\mat{U}) \cos\Theta + \mat{Q} \sin\Theta \right], \quad \text{where } \mat{Q} \defeq -\mat{M}\mat{U}\cot\Psi + \mat{N}\mat{V}\csc\Psi.
\end{equation*}
For indices where $\psi_i = 0$, the corresponding terms are defined via the limit (effectively treated as identity components).

The boundary conditions are trivially satisfied.
To verify the geodesic property, we first establish the orthonormality of the basis vectors.
Observing that $(\mat{M}\mat{U})^\top \mat{Q} = -\cot\Psi + \cos\Psi \csc\Psi = 0$, we proceed to compute the inner product between two intermediate subspaces:
\begin{align*}
    \mathcal{G}(\Theta_1)^\top \mathcal{G}(\Theta_2)
    &= (\cos\Theta_1)(\cos\Theta_2) + (\sin\Theta_1) \mat{Q}^\top \mat{Q} (\sin\Theta_2) \\
    &= \cos\Theta_1 \cos\Theta_2 + \sin\Theta_1 \sin\Theta_2 \\
    &= \cos(\Theta_1 - \Theta_2).
\end{align*}
Thus, the principal angles are determined by $|\Theta_1 - \Theta_2|$.
\end{proof}

\section{Proof of Theorem~\ref{thm:lower_bound}: minimax lower bound}
\label{appendix:lower-bound}
\subsection{General reduction scheme via Fano's method}
\label{reduction_to_finite}

We consider the class of estimators $\Phi$ consisting of mappings $\phi: (\mathbb{R}^p)^T \to \grass{k}{p}$ that output a $k$-dimensional subspace $\phi_{\seqX}$ given observations $\seqX$.
Our goal is to estimate the terminal column space $\ran(\mat{A}_T) \in \grass{k}{p}$.
We quantify the estimation error using the projection 2-distance:
\begin{equation*}
    \dist(\ran(\mat{A}_T), \phi_{\seqX}) \defeq \proje{\ran(\mat{A}_T)}{\phi_{\seqX}} = \norm{\mat{P}_{\mat{A}_T} - \mat{P}_{\phi_{\seqX}}}_2,
\end{equation*}
where $\mat{P}_{\mathcal{S}}$ denotes the orthogonal projector onto the subspace $\mathcal{S}$.

\paragraph{What we want to prove.}
The minimax lower bound $\mathcal{R}^* = \Omega(s)$ asserts that every estimator $\phi \in \Phi$ incurs expected error of order~$s$ on some sequence $\seqA \in \tempuncert$.
We prove it via the Generalized Fano Method~\citep{tsybakov2008introduction}, which reduces the estimation problem to a finite $M$-way hypothesis test.
The reduction constrains the pairwise distance between candidates from both sides.
They must be far apart so that estimator failure forces test failure, a \emph{lower} bound, yet close enough to keep each pair's KL small, an \emph{upper} bound (since KL grows with the squared subspace distance).

\paragraph{Step 1: a test from any estimator.}
Pick $M$ data generating sequences $\seqA_1, \dots, \seqA_M \in \tempuncert$, along with a reference $\seqA_0$; we will design these candidates below.
Each candidate is parametrized by a Grassmannian trajectory $[\mat{U}^{(i)}_t]_{t=1}^T$ of $k$-dimensional subspaces of $\mathbb{R}^p$, with signal matrix $\mat{A}^{(i)}_t = \sqrt{\delta}\,\mat{U}^{(i)}_t$ at every step.
Every candidate therefore shares the same isotropic spike spectrum, eigenvalues $\sigma^2 + \delta$ on the rank-$k$ signal subspace $\ran(\mat{U}^{(i)}_t)$ and $\sigma^2$ on its orthogonal complement, and the candidates differ only in how that signal subspace rotates with $t$.
The drift constraint of $\tempuncert$, $\norm{\xxT{\mat{A}}{t} - \xxT{\mat{A}}{t-1}} \le \Gamma$, becomes a per-step Grassmannian rotation cap $\proje{[\mat{U}^{(i)}_{t-1}]}{[\mat{U}^{(i)}_t]} \le \Gamma/\delta$ after factoring out the $\delta$ in the spectrum.
The hypothesis $\delta > \Gamma$ makes this cap non-trivial ($\Gamma/\delta < 1$, so consecutive subspaces are bounded away from an orthogonal flip).
Each trajectory starts at a common subspace shared with $\seqA_0$ and ends at a unique terminal $[\mat{U}^{(i)}_T]$.
Given any estimator $\phi \in \Phi$, associate to it the \emph{minimum-distance test}
\begin{equation*}
\xi^*(\seqX) \defeq \arg\min_{i \in \{1,\dots,M\}}\, \dist(\phi_{\seqX}, \ran(\mat{A}^{(i)}_T)),
\end{equation*}
which returns the candidate index whose terminal subspace is closest to the estimator's output.
Heuristically, if the estimator identifies the true subspace well, the test identifies the true sequence well.

\paragraph{Step 2: metric separation makes estimator failure imply test failure.}
For the test in Step~1 to inherit the estimator's quality, we require the candidates to be pairwise far apart in subspace distance:
\begin{equation}
\label{eq:separation_condition}
\proje{\ran(\mat{A}^{(i)}_T)}{\ran(\mat{A}^{(j)}_T)} \ge 2s, \quad \forall i \neq j \in \{1, \dots, M\}.
\end{equation}
Under~\eqref{eq:separation_condition}, if the true index is $j^\ast$ and $\dist(\phi_{\seqX}, \ran(\mat{A}^{(j^\ast)}_T)) < s$, then the triangle inequality gives $\dist(\phi_{\seqX}, \ran(\mat{A}^{(i)}_T)) > s$ for every $i \neq j^\ast$, so $\xi^*(\seqX) = j^\ast$.
The contrapositive (test failure implies estimator error $\geq s$) yields
\begin{equation}
\label{eq:test-to-est}
\Pprob_{\seqA_{j}}\!\Big[\dist\big(\phi_{\seqX}, \ran(\mat{A}^{(j)}_T)\big) \ge s\Big] \;\geq\; \Pprob_{\seqA_{j}}\!\big(\xi^* \neq j\big).
\end{equation}

\paragraph{Step 3: information constraint makes the test fail with constant probability.}
For the test's failure probability to be bounded below by a constant, we require the candidates to be statistically close in KL on average:
\begin{equation}
\label{eq:kl_constraint}
\frac{1}{M} \sum_{j=1}^M \kl{\Pprob_{\seqA_j}}{\Pprob_{\seqA_0}} \le \alpha \log M, \quad 0 < \alpha < 1/8.
\end{equation}
The role of $\log M$ is information-theoretic: identifying which of $M$ candidates generated the data requires the data to carry $\log M$ nats of information, and the KL divergence is the natural per-candidate measure of how much information the observations provide.
When~\eqref{eq:kl_constraint} holds, Fano's inequality (Lemma~\ref{lem:fano}) lower-bounds the test's minimax error:
\begin{equation}
\label{eq:fano-floor}
p_{e,M} \;\defeq\; \inf_\xi \max_{j} \Pprob_{\seqA_j}\!\big(\xi \neq j\big) \;\ge\; \frac{\sqrt{M}}{1+\sqrt{M}}\left(1 - 2\alpha - \sqrt{\frac{2\alpha}{\log M}}\right),
\end{equation}
which stays a positive constant for $\alpha < 1/8$ and large $M$, so $p_{e,M} = \Omega(1)$.

\paragraph{Step 4: combine.}
Chaining~\eqref{eq:test-to-est} (with the worst-case $j$) and~\eqref{eq:fano-floor}, every estimator $\phi$ admits some $j$ with
\begin{equation*}
\Pprob_{\seqA_{j}}\!\big[\dist(\phi_{\seqX}, \ran(\mat{A}^{(j)}_T)) \ge s\big] \;\geq\; p_{e,M} \;=\; \Omega(1).
\end{equation*}
Markov's inequality lifts this to expected error: $\mathbb{E}_{\seqA_{j}}[\dist(\phi_{\seqX}, \ran(\mat{A}^{(j)}_T))] \ge s \cdot p_{e,M} = \Omega(s)$.
Since this holds for some $\seqA_{j} \in \tempuncert$,
\begin{equation*}
\mathcal{R}^* \;=\; \inf_\phi \sup_{\seqA \in \tempuncert} \mathbb{E}_{\mathcal{X}\sim\seqA}[\dist(\phi_{\mathcal{X}}, \ran(\mat{A}_T))] \;\geq\; \Omega(s).
\end{equation*}
\paragraph{The value of $s$.}
The bound $\mathcal{R}^* \ge \Omega(s)$ holds for any $s$ meeting both constraints, so the lower bound is the largest such $s$.
The separation side is cheap: the Gilbert--Varshamov volume argument keeps $\log M$ at scale $\Theta(kp)$, essentially independent of $s$, so~\eqref{eq:kl_constraint} is the binding constraint.
Geometrically close candidates are statistically hard to tell apart: the per-pair KL scales as the square of the subspace distance~\eqref{eq:kl_final_bound}.
Confining all candidates to a radius-$3s$ ball about the reference $[\mat{U}^{(0)}_T]$ caps each pairwise distance at $6s$~\eqref{2s6s} and each KL at $O(s^2)$.

The trajectory budget then sets the prefactor through two regimes.
If $T \le 3s\delta/\Gamma$, the $T$ steps at the per-step cap $\Gamma/\delta$ span only $T\Gamma/\delta \le 3s$ of distance, so each candidate rotates at the cap throughout, the total KL is $Ts^2$, and~\eqref{eq:kl_constraint} gives $s \sim 1/\sqrt{T}$.
If $T > 3s\delta/\Gamma$, the drift finishes in $\tau \defeq \lceil 3s\delta/\Gamma\rceil$ steps (the ``lazy'' allocation), the total KL is $\tau s^2 = (3\delta/\Gamma)s^3$, and~\eqref{eq:kl_constraint} gives $s \sim \Gamma^{1/3}$.
The cube root appears because a larger separation is doubly expensive: each drifting step costs about $s^2$ in KL, and a larger $s$ needs proportionally more drifting steps.
The two regimes give the two arguments of the maximum in~\eqref{eq:s_value}.

Conversely, any packing that raises the per-pair KL at the same separation pushes the saturation point of~\eqref{eq:kl_constraint} downward and shrinks the bound, the diagnostic we use in Appendix~\ref{appendix:mu-gap} to rule out a natural attempt at sharpening the $\mu$ dependence.

\paragraph{What remains.}
The construction proceeds in three steps: (i) building the candidates on the Grassmannian to satisfy~\eqref{eq:separation_condition}, (ii) bounding the KL under the spiked covariance model, and (iii) balancing $M$ against the KL constraint to extract the explicit $s$ in~\eqref{eq:s_value}.

\subsection{Construction of the packing set}
\label{constructAt}

We construct a packing set of $M+1$ sequences $\{\seqA_{i}\}_{i=0}^{M} \subset \tempuncert$.
To satisfy the spectral condition $s_k(\mat{A}^{(i)}_{t}\mat{A}^{(i)\top}_{t}) = \specgap$, we parameterize each matrix as:
\begin{equation*}
    \mat{A}^{(i)}_{t} = \sqrt{\specgap} \cdot \mat{U}^{(i)}_{t}, \quad \text{where } \mat{U}^{(i)}_{t} \in \stiefel{k}{p}.
\end{equation*}
This restricts the candidate class to isotropic spikes ($\mu = 1$, with $s_1 = s_k = \specgap$), a sub-class of $\tempuncert$: a minimax lower bound on this sub-class is a lower bound on the full $\tempuncert$ by inclusion.
Since the projection distance $\proje{\cdot}{\cdot}$ depends solely on the subspace (Grassmannian), the scaling factor $\sqrt{\specgap}$ does not affect the geometric construction.
Thus, we describe the construction in terms of the orthonormal bases $\mat{U}^{(i)}_{t}$.

The base hypothesis corresponds to a stationary sequence:
\begin{equation*}
    \seqA_{0} = \{\mat{A}^{(0)}_{1}, \dots, \mat{A}^{(0)}_{T}\}, \quad \text{with } \mat{U}^{(0)}_{t} \equiv [\vect{e}_1, \dots, \vect{e}_k].
\end{equation*}

The construction must satisfy two conditions, each with a distinct role downstream.
(i) \emph{Terminal separation}: the pairwise distances of the candidates' terminal subspaces lie in a band $[2s, 6s]$.
The lower bound $2s$ is what lets Step~2's metric-separation argument convert estimator error into test failure (Fano's identifiability requirement); the upper bound $6s$ caps each per-pair KL via~\eqref{eq:kl_final_bound}, so the Fano constraint~\eqref{eq:kl_constraint} pins down $s$ rather than letting the per-pair KL scale with the covering ball's radius.
(ii) \emph{Trajectory consistency}: every per-step Grassmannian rotation must respect $\proje{[\mat{U}^{(i)}_{t-1}]}{[\mat{U}^{(i)}_{t}]} \le \varbudget/\specgap$, the Grassmannian translation of $\tempuncert$'s drift cap on the candidate sequence.
We address the two conditions in turn.

\paragraph{Terminal separation.}
We construct $M \ge (3/2)^{k(p-k-1/2)}$ terminal subspaces $[\mat{U}^{(i)}_{T}] \in \grass{k}{p}$ such that for all distinct $i, j \in [M]$:
\begin{equation}
\label{2s6s}
    2s \le \proje{[\mat{U}^{(i)}_{T}]}{[\mat{U}^{(j)}_{T}]} \le 6s.
\end{equation}
Let $\set{B}([\mat{X}], r)$ denote the projection 2-distance ball of radius $r$ centered at $[\mat{X}]$.
Define the covering set $\set{S}_c = \set{B}([\mat{U}^{(0)}_{T}], 3s)$ and per-candidate exclusion balls $\set{S}_i = \set{B}([\mat{U}^{(i)}_{T}], 2s)$.
The plan is to pack as many candidates as possible into the radius-$3s$ covering set $\set{S}_c$ such that their radius-$2s$ exclusion balls are pairwise disjoint; the count of such candidates is the packing number $M$.

\emph{Volume ratio.}
Provided $s \le \frac{1}{2\sqrt{3}}$, applying Proposition~\ref{prop:grassvolume} to the numerator (lower bound at $r = 3s$) and the denominator (upper bound at $r = 2s$) yields
\begin{equation*}
    \frac{\mu(\set{S}_c)}{\mu(\set{S}_i)} \ge \left(\frac{3}{2}\right)^{k(p-k)} (1 - 4s^2)^{k/2} \ge \left(\frac{3}{2}\right)^{k(p-k-1/2)},
\end{equation*}
where the last inequality absorbs the $(1 - 4s^2)^{k/2}$ factor by noting that $(1 - 4s^2)^{k/2} \ge (2/3)^{k/2}$ precisely when $s \le \frac{1}{2\sqrt{3}}$.

\emph{From volume ratio to packing count.}
The Gilbert--Varshamov volume argument turns this ratio into a lower bound on $M$.
Take a maximal $2s$-separated subset $\{[\mat{U}^{(i)}_T]\}_{i=1}^M$ of $\set{S}_c$: pairwise distances $\ge 2s$, with $M$ as large as possible.
Such a maximum exists with $M < \infty$ because $\set{S}_c$ is compact (a closed ball in the compact Grassmannian); an infinite $2s$-separated set would have a limit point and contain arbitrarily close pairs, contradicting the $2s$-separation.
Maximality then forces the radius-$2s$ balls about the chosen points to cover all of $\set{S}_c$: any uncovered point would sit at distance $\ge 2s$ from every chosen point and could be added without breaking the separation.
Summing volumes gives $M\,\mu(\set{S}_i) \ge \mu(\set{S}_c)$.
The Grassmannian is homogeneous, so all radius-$2s$ balls share one volume; dividing,
\[
M \;\ge\; \frac{\mu(\set{S}_c)}{\mu(\set{S}_i)} \;\ge\; (3/2)^{k(p-k-1/2)},
\]
which is the lower-bound half of~\eqref{2s6s}.

\emph{Upper bound by construction.}
The upper-bound half $\proje{[\mat{U}^{(i)}_{T}]}{[\mat{U}^{(j)}_{T}]} \le 6s$ falls out of confining the packing to the radius-$3s$ ball $\set{S}_c$: each candidate lies within distance $3s$ of the center $[\mat{U}^{(0)}_T]$, so the triangle inequality gives
\begin{equation*}
\proje{[\mat{U}^{(i)}_T]}{[\mat{U}^{(j)}_T]}
  \;\leq\; \proje{[\mat{U}^{(i)}_T]}{[\mat{U}^{(0)}_T]}
         + \proje{[\mat{U}^{(0)}_T]}{[\mat{U}^{(j)}_T]}
  \;\leq\; 3s + 3s \;=\; 6s
\end{equation*}
for every $i \neq j$.
The upper bound is actively used downstream: \eqref{eq:kl_final_bound} bounds $\kl{\Pprob_{\seqA_i}}{\Pprob_{\seqA_j}}$ by a constant times $\proje{[\mat{A}^{(i)}_T]}{[\mat{A}^{(j)}_T]}^2$, and the substitution $\proje{[\mat{A}^{(i)}_T]}{[\mat{A}^{(j)}_T]} \le 6s$ at~\eqref{eq:kl_bound_total} is what turns the abstract Fano constraint~\eqref{eq:kl_constraint} into the concrete arithmetic that yields~\eqref{eq:s_value}.
Without the upper-bound half of~\eqref{2s6s}, the per-pair KL could in principle grow with the radius of the covering ball, and the Fano constraint would not pin down $s$ to a clean expression.

\paragraph{Trajectory consistency.}
The trajectories $\mat{U}^{(i)}_{t}$ for $t < T$ must satisfy the per-step drift constraint
\begin{equation}
\label{eq:traj-consistency}
    \proje{[\mat{U}^{(i)}_{t-1}]}{[\mat{U}^{(i)}_{t}]} \le \frac{\varbudget}{\specgap}.
\end{equation}
Each candidate $i$ must travel from $[\mat{U}^{(0)}_{T}]$ to $[\mat{U}^{(i)}_{T}]$ along the Grassmannian within $T$ steps, moving at most $\Gamma/\delta$ per step (the per-step constraint).
Because $[\mat{U}^{(i)}_{T}] \in \set{S}_c$, the radius-$3s$ ball around $[\mat{U}^{(0)}_{T}]$, the journey covers at most distance $3s$, so it takes at least
\begin{equation*}
\tau \;\defeq\; \lceil 3s\specgap/\varbudget \rceil
\end{equation*}
steps.
We allocate these $\tau$ steps to the tail of the trajectory: candidate $i$ stays put at $[\mat{U}^{(0)}_{T}]$ for $t \le T - \tau$, and only over the last $\tau$ steps does it move along the geodesic to $[\mat{U}^{(i)}_{T}]$ at a constant per-step distance $\le \Gamma/\delta$.
We implement the geodesic motion via the principal rotation mapping $\mathcal{G}$ (Definition~\ref{def:principal_rotation}).
The point of this ``lazy'' allocation is that $\mat{A}^{(i)}_t = \mat{A}^{(0)}_t$ exactly for $t \le T - \tau$, so when we bound the KL between $\Pprob_{\seqA_{i}}$ and $\Pprob_{\seqA_{0}}$ in the next subsection, those $T - \tau$ identical steps contribute zero, and the entire KL accumulates over only the last $\tau$ steps.

\subsection{Bounding the KL divergence}

It suffices to derive the lower bound for the Gaussian sub-class $z_{t,i}\sim\mathcal{N}(0,1)$ of model~\eqref{eqn:spike}: a minimax lower bound on this sub-class is a lower bound on the full sub-Gaussian class by inclusion.
We henceforth compute the KL under Gaussian observations and invoke Lemma~\ref{lem:spiked-kl}.

We bound the KL divergence between hypotheses.
Since the observations are independent across time, the KL decomposes additively.
The lazy allocation of Appendix~\ref{constructAt} ensures $\mat{A}^{(i)}_t = \mat{A}^{(0)}_t$ for every $t \le T - \tau$, so those steps contribute zero per-step KL and the sum runs only over the last $\tau = \lceil 3s\delta/\varbudget \rceil$ steps.
Bounding each per-step KL by its maximum,
\begin{equation}
    \kl{\Pprob_{\seqA_{i}}}{\Pprob_{\seqA_{j}}} = \sum_{t=1}^T \kl{\mat{A}^{(i)}_{t}}{\mat{A}^{(j)}_{t}} \le \min\left\{T, \frac{3s\specgap}{\varbudget}\right\} \max_t \kl{\mat{A}^{(i)}_{t}}{\mat{A}^{(j)}_{t}},
    \label{eq:kl_decomposition}
\end{equation}
where the prefactor is $\tau = 3s\delta/\Gamma$ in the regime where the trajectory budget admits a partial-motion construction and degenerates to $T$ when $\tau \ge T$ (the small-$s$ regime where the journey takes the entire stream).
This $\min$ is the source of the phase transition in the lower bound between the $1/\sqrt{T}$ and $\Gamma^{1/3}$ scalings.

For the spiked covariance model $\Sigma^{(i)} = \specgap \mat{U}^{(i)} (\mat{U}^{(i)})^\top + \sigma^2 \mat{I}_p$, where $\mat{U}^{(i)} \in \stiefel{k}{p}$ is the orthonormal basis of $\mat{A}^{(i)}$, Lemma~\ref{lem:spiked-kl} in the isotropic case $D = \specgap\mat{I}_k$ gives
\begin{equation*}
    \kl{\mat{A}^{(i)}}{\mat{A}^{(j)}} = \frac{\specgap^2}{4\sigma^2(\sigma^2+\specgap)} \frob{ \mat{U}^{(i)} (\mat{U}^{(i)})^\top - \mat{U}^{(j)} (\mat{U}^{(j)})^\top }^2.
\end{equation*}
Using the relationship between the Frobenius norm and the operator norm, $\frob{\mat{M}}^2 \le \rank(\mat{M}) \oper{\mat{M}}^2$, and noting that the rank of the difference of projectors is at most $2k$ (Lemma~\ref{lem:subaddrank}):
\begin{equation}
    \kl{\mat{A}^{(i)}}{\mat{A}^{(j)}} \le \frac{k\specgap^2}{2\sigma^2(\sigma^2+\specgap)} \proje{[\mat{A}^{(i)}]}{[\mat{A}^{(j)}]}^2.
    \label{eq:kl_final_bound}
\end{equation}
Substituting the separation bound $\proje{[\mat{A}^{(i)}_T]}{[\mat{A}^{(j)}_T]} \le 6s$ into \eqref{eq:kl_final_bound}, we obtain:
\begin{equation}
    \kl{\Pprob_{\seqA_{i}}}{\Pprob_{\seqA_{j}}} \le \frac{18k\specgap^2}{\sigma^2(\sigma^2+\specgap)} \min\left\{T, \frac{3s\specgap}{\varbudget}\right\} s^2.
    \label{eq:kl_bound_total}
\end{equation}

The same inequality bounds the candidate-versus-reference divergences $\kl{\Pprob_{\seqA_{j}}}{\Pprob_{\seqA_{0}}}$ that enter the Fano constraint~\eqref{eq:kl_constraint}: the reference $\seqA_0$ lies at the center of the radius-$3s$ ball, so $\proje{[\mat{A}^{(j)}_T]}{[\mat{A}^{(0)}_T]} \le 3s \le 6s$ and the bound~\eqref{eq:kl_bound_total} applies unchanged.

\subsection{Completing the proof}

By the terminal-separation construction of Appendix~\ref{constructAt} (the lower-bound half of~\eqref{2s6s}), the terminal subspaces satisfy:
\begin{equation*}
    \proje{[\mat{U}^{(i)}_{T}]}{[\mat{U}^{(j)}_{T}]} \ge 2s, \quad \forall i \neq j \in \{1, \dots, M\}.
\end{equation*}
Let $\phi_{\seqX}$ be any estimator of the terminal principal subspace.
By the triangle inequality, for any $j \in \{1, \dots, M\}$, we have:
\begin{equation*}
    \Pprob_{\seqA_{j}}\Big[\dist\big(\ran(\mat{A}^{(j)}_{T}), \phi_{\seqX}\big) \ge s\Big] \ge \Pprob_{\seqA_{j}}(\xi^{\ast} \neq j),
\end{equation*}
where $\xi^{\ast}: \seqX \mapsto \{1, \dots, M\}$ denotes the minimum distance test defined by $\xi^{\ast} \defeq \argmin_{i} \dist(\ran(\mat{A}^{(i)}_{T}), \phi_{\seqX})$.
The minimax risk is thus lower-bounded by the minimax probability of error $p_{e,M}$:
\begin{equation*}
    \inf_{\phi} \sup_{\seqA \in \tempuncert} \Pprob_{\seqA}\Big[\dist\big(\ran(\mat{A}_{T}), \phi_{\seqX}\big) \ge s \Big] \ge \inf_{\xi} \max_{j} \Pprob_{\seqA_{j}}(\xi \neq j) \defeq p_{e,M}.
\end{equation*}
To lower bound $p_{e,M}$, we invoke the generalized Fano's inequality.

\begin{lemma}[Theorem~2.5 in \citet{tsybakov2008introduction}]
\label{lem:fano}
Let $\Pprob_0, \Pprob_1, \dots, \Pprob_M$ be probability measures satisfying
\begin{equation*}
    \frac{1}{M}\sum_{j=1}^{M} \kl{\Pprob_{\seqA_{j}}}{\Pprob_{\seqA_{0}}} \le \alpha \log M,
\end{equation*}
for some $0 < \alpha < 1/8$.
Then, the minimax probability of error is bounded by:
\begin{equation*}
    p_{e,M} \ge \frac{\sqrt{M}}{1+\sqrt{M}} \left(1 - 2\alpha - \sqrt{\frac{2\alpha}{\log M}}\right).
\end{equation*}
\end{lemma}

We choose $s$ such that the Fano condition holds with $\alpha = 1/10$:
\begin{equation*}
    \frac{1}{M} \sum_{j=1}^M \kl{\Pprob_{\seqA_{j}}}{\Pprob_{\seqA_{0}}} \le \frac{1}{10} \log M.
\end{equation*}
Using the bound in \eqref{eq:kl_bound_total} and the fact that $\log M \ge k(p-k-1/2) \log(3/2) \ge \frac{kp}{2}\log(3/2)$ (since $p > 2k$), it suffices to satisfy:
\begin{equation*}
    \frac{180 \specgap^2}{\sigma^2(\sigma^2+\specgap)} \min\left(T, \frac{3\specgap s}{\varbudget} \right) s^2 \le \frac{1}{2} p \log \frac{3}{2}.
\end{equation*}
The left-hand side increases in $s$, and the $\min$ takes only its drift-limited branch $3\delta s/\varbudget$ when $s$ is small and its sample-limited branch $T$ when $s$ is large.
Since the cost is the \emph{smaller} of the two branches, it suffices for $s$ to fall under \emph{either} branch threshold, so the inequality holds for every $s$ up to the larger of them:
\begin{equation}
    \label{eq:s_value}
    s = \max\left\{ \sqrt[3]{\frac{\log(3/2)}{1080}}\left(\frac{\varbudget}{\specgap}\right)^{1/3}\!\left( \frac{p\sigma^2(\sigma^2 +\specgap)}{\specgap^2} \right)^{1/3}\!,\ \ \sqrt{\frac{\log(3/2)}{360}}\,\frac{1}{\sqrt{T}}\left(\frac{p\sigma^2(\sigma^2 +\specgap)}{\specgap^2}\right)^{1/2} \right\}.
\end{equation}
The first argument solves the drift-limited cubic branch and the second the sample-limited quadratic branch.
Since $\max\{a,b\}$ and $a+b$ agree up to a factor of two, this $s$ realizes the rate stated in Theorem~\ref{thm:lower_bound}.
The packing construction in Appendix~\ref{constructAt} requires $s \le \frac{1}{2\sqrt{3}}$.
Since $s = O\big((\varbudget/\specgap)^{1/3}(p\sigma^2(\sigma^2+\specgap)/\specgap^2)^{1/3} + T^{-1/2}(p\sigma^2(\sigma^2+\specgap)/\specgap^2)^{1/2}\big)$, this condition is satisfied whenever $p$, $\sigma^2$, $\varbudget$, and $T$ are in the regime where the lower bound is non-trivial.

With this choice, Lemma~\ref{lem:fano} with $\alpha = 1/10$ gives
\begin{equation*}
    p_{e,M} \ge \frac{\sqrt{M}}{1+\sqrt{M}} \left(\frac{4}{5} - \frac{1}{\sqrt{5\log M}}\right).
\end{equation*}
Under our assumption $p > 2k$ with $k \ge 1$, we have $k(p-k-1/2) \ge \frac{3}{2}$ (tight at $k=1$, $p = 2k+1$), so $\log M \ge \frac{3}{2}\log\frac{3}{2}$ and $M \ge (3/2)^{3/2} > 1$.
Consequently, $\frac{\sqrt{M}}{1+\sqrt{M}} > \frac{1}{2}$, and using $\log(3/2) > \frac{2}{5}$, $\frac{1}{\sqrt{5\log M}} \le \frac{1}{\sqrt{3}} < \frac{3}{5}$, which yields $p_{e,M} > \frac{1}{2}\bigl(\frac{4}{5} - \frac{3}{5}\bigr) = \frac{1}{10}$.
Since the minimax risk is lower-bounded by $s \cdot p_{e,M}$, we conclude
\begin{equation*}
    \minimaxL = \Omega(s).
\end{equation*}

\section{Proof of Lemma~\ref{lem:power-error}: power-method error bound}
\label{appendix:power-error-bound}

\subsection{Concentration parameters}
\label{appendix:npm-params}

The sub-exponential matrix Bernstein of~\citep[Proposition~4.1]{KlochkovZhiv2020} (see also~\citealt{kroshnin2024bernstein}) demands a $\psi_1$ control on the per-summand operator norm and a spectral bound on the matrix variance.
We derive both directly from the sub-Gaussian structure of $\vect{z}_t$; no truncation is required at this stage.
(Oja's Huang matrix-product concentration does need an a.s.\ bound; see Appendix~\ref{appendix:tail-bounds}.)

Recall under model~\eqref{eqn:spike} that $\vect{x}_t = \Sigma_t^{1/2}\vect{z}_t$ where $\vect{z}_t \in \mathbb{R}^p$ has independent, mean-zero, unit-variance entries satisfying $\lVert z_{t,i}\rVert_{\psi_2}\le\kappa$.
The centered observations
\[
\mat{X}_t := \xxT{\vect{x}}{t} - \Sigma_t
\]
are independent, mean-zero, and symmetric.

\paragraph{Sub-exponential spectral norm $\mathcal{M}$.}
The summand norm $\lVert\vect{x}_t\rVert^2 = \vect{z}_t^\top\Sigma_t\vect{z}_t$ is a sub-Gaussian quadratic form.
By Hanson--Wright~\citep[Theorem~6.2.1]{Vershynin2018},
\[
\bigl\lVert\vect{z}_t^\top\Sigma_t\vect{z}_t - \mathrm{tr}(\Sigma_t)\bigr\rVert_{\psi_1}
  \;\le\; c\kappa^2\bigl(\lVert\Sigma_t\rVert_F + \lVert\Sigma_t\rVert\bigr)
  \;\le\; 2c\kappa^2\mathrm{tr}(\Sigma_t),
\]
the second step using $\lVert\Sigma_t\rVert_F, \lVert\Sigma_t\rVert \le \mathrm{tr}(\Sigma_t)$ for PSD $\Sigma_t$.
The deterministic shift $\mathrm{tr}(\Sigma_t)$ contributes $\mathrm{tr}(\Sigma_t)/\log 2$ to the $\psi_1$ norm, so absorbing all absolute constants into a single $C_{\mathrm{q}}>0$ gives
\[
\bigl\lVert\vect{z}_t^\top\Sigma_t\vect{z}_t\bigr\rVert_{\psi_1}
  \;\le\; C_{\mathrm{q}}\kappa^2\mathrm{tr}(\Sigma_t).
\]

The operator-norm triangle inequality then yields $\lVert\mat{X}_t\rVert \le \lVert\vect{x}_t\rVert^2 + \lVert\Sigma_t\rVert$.
Taking $\psi_1$ norms, substituting $\lVert\Sigma_t\rVert \le \mu\delta+\sigma^2$ and $\mathrm{tr}(\Sigma_t) \le k\mu\delta+p\sigma^2$, and absorbing the deterministic $\psi_1$ contribution $\lVert\Sigma_t\rVert/\log 2$ into the absolute constants below,
\begin{equation}
\label{eq:M-bound}
\bigl\lVert\,\lVert\mat{X}_t\rVert\,\bigr\rVert_{\psi_1}
  \;\le\; C_{\mathrm{q}}\,\kappa^2\,(k\mu\delta+p\sigma^2) + (\mu\delta+\sigma^2).
\end{equation}
Matrix Bernstein (Appendix~\ref{appendix:robust-power-method}) requires a uniform handle on $\max_{t\le B}\lVert\mat{X}_t\rVert$ across the $B$-summand block, not just per-summand $\psi_1$ control.
The standard sub-exponential max-norm inflation $\bigl\lVert\max_{t\le B}\lVert\mat{X}_t\rVert\bigr\rVert_{\psi_1}\le C\log B \cdot \max_{t\le B}\bigl\lVert\,\lVert\mat{X}_t\rVert\,\bigr\rVert_{\psi_1}$~\citep[Remark~4.1]{KlochkovZhiv2020} thus multiplies the bound in~\eqref{eq:M-bound} by a factor of $\log B\le\log T$.
Absorbing this logarithmic inflation into an absolute constant $C_{\mathcal{M}}>0$, we work with the parameter
\begin{equation}
\label{eq:M-simplified}
\mathcal{M} := \bigl(C_{\mathcal{M}}\,\kappa^2\,(k\mu\delta+p\sigma^2) + (\mu\delta+\sigma^2)\bigr)\log T.
\end{equation}

\paragraph{Matrix variance $\mathcal{V}$.}
For $\vect{x}=\Sigma_t^{1/2}\vect{z}$ with $\vect{z}$ having independent mean-zero sub-Gaussian entries, the fourth-moment identity $\mathbb{E}\bigl[(\vect{z}^\top A\vect{z})\,\xxT{\vect{z}}{}\bigr]=\mathrm{tr}(A)\mat{I}+2A+\mathrm{diag}\bigl(A_{ii}(\mathbb{E}[z_i^4]-3)\bigr)$, together with the sub-Gaussian fourth-moment bound $\mathbb{E}[z_i^4]\le C_4\kappa^4$~\citep[Proposition~2.5.2]{Vershynin2018}, yields $\mathbb{E}[\mat{X}_t^2]\preceq\mathbb{E}\bigl[\lVert\vect{x}_t\rVert^2\xxT{\vect{x}}{t}\bigr]\preceq C_4'\,\kappa^4\,\mathrm{tr}(\Sigma_t)\,\Sigma_t$ for an absolute constant $C_4'>0$.
The identity is component-wise in $\mathbb{E}[z_k z_l z_i z_j]$ under independence: the four indices must pair up, which gives $\mathrm{tr}(A)\mat{I}+2A$ from the three pairings and the diagonal kurtosis correction from $k=l=i=j$.
Hence
\begin{equation}
\label{eq:V-bound}
\bigl\lVert\mathbb{E}[\mat{X}_t^2]\bigr\rVert
  \;\le\; C_4'\,\kappa^4\,(k\mu\delta+p\sigma^2)(\mu\delta+\sigma^2),
\end{equation}
and $\bigl\lVert\sum_{t\in\text{block}}\mathbb{E}[\mat{X}_t^2]\bigr\rVert\le B\,\mathcal{V}$ for
\begin{equation}
\label{eq:V-simplified}
\mathcal{V} := C_{\mathcal{V}}\,\kappa^4\,(k\mu\delta+p\sigma^2)(\mu\delta+\sigma^2),
\end{equation}
where $C_{\mathcal{V}}>0$ absorbs the per-summand constants $C_4'$/Bernstein.

\begin{remark}[Magnitude of the absolute constants]
\label{rmk:absolute-constants}
The constants $C_{\mathrm{q}},C_4,C_4',C_{\mathcal{M}},C_{\mathcal{V}},C_{\mathrm{B}}>0$ are all absolute (free of $p,k,T,\delta,\sigma^2,\kappa$) and split by origin: $C_{\mathrm{q}},C_4,C_4'$ are per-summand Hanson--Wright / fourth-moment constants (each a small absolute number), and $C_{\mathcal{M}},C_{\mathcal{V}},C_{\mathrm{B}}$ are matrix-concentration constants of comparable scale.
The bounded-summand matrix Bernstein tail~\citep[Theorem~5.4.1]{Vershynin2018} admits the explicit anchor $(2+3\sqrt{2})/3 \approx 2.08$; the sub-exponential matrix Bernstein actually used here yields a somewhat larger constant of the same order, and Appendix~\ref{appendix:robust-power-method} takes $C_{\mathrm{B}}\le 3.2$.

In the Gaussian case ($\kappa=1$) the fourth-moment identity is exact, giving $\bigl\lVert\mathbb{E}[\mat{X}_t^2]\bigr\rVert\le\mathrm{tr}(\Sigma_t)\lVert\Sigma_t\rVert+2\lVert\Sigma_t\rVert^2$; consolidating via $\lVert\Sigma_t\rVert\le\mathrm{tr}(\Sigma_t)$ gives $\le 3\,\mathrm{tr}(\Sigma_t)\lVert\Sigma_t\rVert$, so $C_4'\le 3$.
\end{remark}

\subsection{Bounding the stochastic error}

The centered matrices $\mat{X}_t = \xxT{\vect{x}}{t} - \Sigma_t$ satisfy
\[
\bigl\lVert\max_t\lVert\mat{X}_t\rVert\bigr\rVert_{\psi_1} \leq \mathcal{M},
\qquad
\Bigl\lVert\sum_{t \in \text{block}}
  \mathbb{E}[\mat{X}_t^2]\Bigr\rVert
\leq B\mathcal{V},
\]
with $\mathcal{M}$ and $\mathcal{V}$ as in~\eqref{eq:M-simplified}--\eqref{eq:V-simplified}.

We apply the following matrix concentration inequality.

\begin{theorem}[Sub-exponential matrix Bernstein~{\citep[Proposition~4.1]{KlochkovZhiv2020}}]
\label{thm:matrix_bernstein}
Let $\mat{X}_1, \ldots, \mat{X}_B \in \mathbb{R}^{p\times p}$ be independent, mean-zero, symmetric random matrices with $\bigl\lVert\,\lVert\mat{X}_i\rVert\,\bigr\rVert_{\psi_1}<\infty$, and set $\mathcal{M}=\bigl\lVert\max_i\lVert\mat{X}_i\rVert\bigr\rVert_{\psi_1}$.
Let
\[
\Bigl\lVert\sum_{i=1}^{B}
     \mathbb{E}[\mat{X}_i^2]\Bigr\rVert \leq B \mathcal{V},
\]
and write $r=\mathrm{tr}\bigl(\sum_i\mathbb{E}[\mat{X}_i^2]\bigr)/\bigl\lVert\sum_i\mathbb{E}[\mat{X}_i^2]\bigr\rVert\le p$ for the effective rank.
Then for every $t\ge 0$ and absolute constants $C,c>0$,
\[
\mathbb{P}\!\Bigl(\Bigl\lVert\frac{1}{B}
  \sum_{i=1}^{B}\mat{X}_i\Bigr\rVert \geq t\Bigr)
\;\leq\; C\,r\,\exp\!\left(
  \frac{-Bt^2}{c\,(\mathcal{V}
  + \mathcal{M}t)}\right).
\]
The $(\mathcal{V}+\mathcal{M}t)$-denominator form is the standard reparameterization of the $\min\{u^2/\sigma^2,u/M\}$-exponent form in~\citep[Proposition~4.1]{KlochkovZhiv2020}, equivalent up to absolute constants since $\mathcal{V}+\mathcal{M}t\in[\max\{\mathcal{V},\mathcal{M}t\},\,2\max\{\mathcal{V},\mathcal{M}t\}]$.
\end{theorem}

\paragraph{Applying matrix Bernstein.}
We set the right-hand side of Theorem~\ref{thm:matrix_bernstein} to at most $1/T^2$, using $r\le p$ to absorb the dimension prefactor $Cr$ into $\log(2pT^2)$, and solve the resulting quadratic in $t$.
Under $B \ge \mathcal{M}^2\log(2pT^2)/\mathcal{V}$ the boundedness term is dominated by the variance term, giving
\[
t \;\le\; C_{\mathrm{B}}\,\sqrt{\frac{\mathcal{V}\log(2pT^2)}{B}}
\]
for an absolute constant $C_{\mathrm{B}}>0$ from the quadratic-solve, of the same order as the explicit $(2+3\sqrt{2})/3 \approx 2.08$ arising in the bounded-summand matrix Bernstein analog~\citep[Theorem~5.4.1]{Vershynin2018} (see Remark~\ref{rmk:absolute-constants}).

\paragraph{Union bound over blocks.}
Since $\mathbb{E}[\xxT{\vect{x}}{t}]=\Sigma_t$, the stochastic error of~\eqref{eqn:npmblock} equals $\mathcal{E}_1(\ell) = \frac{1}{B}\sum_{t}\bigl(\xxT{\vect{x}}{t} - \Sigma_t\bigr) = \frac{1}{B}\sum_t \mat{X}_t$, so a union bound over the $L \leq T$ blocks (each with failure probability at most $1/T^2$) of the matrix Bernstein tail above gives, with probability at least $1 - 1/T$,
\begin{equation}
\label{eq:E1bound}
\max_{1\leq\ell\leq L}\lVert\mathcal{E}_1(\ell)\rVert
  \leq C_{\mathrm{B}}\,
    \sqrt{\frac{\mathcal{V}\log(2pT^2)}{B}}\,.
\end{equation}

\subsection{Bounding the drift error}
Since our model limits the per-step perturbation of the covariance matrix, we may bound $\lVert \mathcal{E}_2(\ell) \rVert$ as follows:
\begin{align*}
\lVert \mathcal{E}_2(\ell) \rVert &\leq \Big\lVert \frac{1}{B} \sum_{t=(\ell-1)B+1}^{\ell B} \Big( \mathbb{E}[\xxT{\vect{x}}{t}]-\mathbb{E}[\xxT{\vect{x}}{\ell B}]\Big) \Big\rVert \leq  \frac{1}{B} \sum_{t=(\ell-1)B+1}^{\ell B} \lVert \mathbb{E}[\xxT{\vect{x}}{t}]-\mathbb{E}[\xxT{\vect{x}}{\ell B}] \rVert\\
& = \frac{1}{B} \sum_{t=(\ell-1 )B+1}^{\ell B} \lVert \xxT{\mat{A}}{t} - \xxT{\mat{A}}{\ell B} \rVert \quad (\text{noise term cancels})\leq \frac{1}{B} \sum_{t=(\ell-1)B+1}^{\ell B} (\ell B - t)\Gamma \\
& \leq \frac{1}{B} \frac{B(B-1)}{2} \Gamma \leq \frac{B\Gamma}{2}\,.
\end{align*}

\subsection{Combining the error bounds}

Combining the stochastic bound on $\max_{\ell}\lVert\mathcal{E}_1(\ell)\rVert$ from~\eqref{eq:E1bound} with the drift bound on $\lVert\mathcal{E}_2(\ell)\rVert$, we obtain: with probability at least $1-1/T$,
\begin{equation}
\label{eq:npmlemmabound}
\max_{\ell}{\lVert \mathcal{E}(\ell) \rVert}
  \leq \max_{\ell}{\lVert \mathcal{E}_1(\ell) \rVert}
     + \max_{\ell}{\lVert \mathcal{E}_2(\ell) \rVert}
  \leq  C_{\mathrm{B}}\,\sqrt{\frac{\mathcal{V}\log(2pT^2)}{B}}
    + \frac{B\Gamma}{2}\,,
\end{equation}
provided $B \geq \mathcal{M}^2 \log(2pT^2)/\mathcal{V}$.
\QEDA

\section{Amplification under perturbation and subspace drift}
\label{appendix:amplification}
This appendix proves the deterministic statement that both Theorem~\ref{thm:robust-power-method} and Theorem~\ref{thm:ojas-algorithm} invoke: Theorem~\ref{thm:drifting-power-method}, which bounds the power iteration's error given only that each per-block error matrix $\mathcal{E}(\ell)$ has small operator norm.
It generalizes the noisy power method's stationary convergence proof~\citep{npm}: we recall that proof in the notation used below (the stationary baseline), isolate the single step that drift breaks (what drift removes), and give the frame construction that restores it.
The stochastic bounds on $\|\mathcal{E}(\ell)\|$ are supplied separately, by Appendix~\ref{appendix:power-error-bound} for the power method and Appendix~\ref{appendix:oja-error-bound} for Oja.

Both algorithms reach this statement because one of their steps is a multiplication by a covariance.
The power method forms a block sample covariance directly, and Oja's algorithm~\citep{oja_1} exponentiates the same covariance by grouping its $\zeta^{-1}$ consecutive multiplicative updates into a virtual block.
A single deterministic argument therefore covers both.

\subsection{Notation}
Fix a target rank $1\le k<p$.
Let $\mat{M}\in\mathbb{R}^{p\times p}$ be positive definite with $\mathrm{SVD}(\mat{M})=\mat{U}\mat{D}\mat{U}^\top$, and let $\delta,\sigma^2>0$ be a spectral gap and a noise floor with
\[
s_k(\mat{M})\ge \delta+\sigma^2,\qquad s_{k+1}(\mat{M})=\sigma^2 .
\]
We write $\mat{U}_{1:k}$ and $\mat{U}_{k+1:p}$ for the leading $k$ and trailing $p-k$ singular vectors of $\mat{M}$; they span the signal subspace and its orthogonal complement.
Here $d(\cdot,\cdot)$ is the subspace distance and $\stiefel{m}{p}$ the set of $p\times m$ matrices with orthonormal columns.
We write $b(\mat{M})\in\stiefel{\mathrm{rk}(\mat{M})}{p}$ for \emph{any} matrix whose columns form an orthonormal basis of $\mathrm{ran}(\mat{M})$, $s_{i}(\mat{M})$ for the $i$-th largest singular value, and $\mat{M}_{\perp}=\mat{I}-\xxT{\mat{U}}{1:r}$ for the projection onto the complement of $\mathrm{ran}(\mat{M})$ when $\mathrm{rk}(\mat{M})=r$.
Such a basis is fixed only up to a right orthogonal factor, $b(\mat{M})\mapsto b(\mat{M})\mat{Q}$ with $\mat{Q}$ orthogonal.
This changes neither $\mathrm{ran}(b(\mat{M}))=\mathrm{ran}(\mat{M})$ nor the singular values of any product $\mat{A}\,b(\mat{M})$, and we use $b(\mat{M})$ only through these, so the particular choice of basis is immaterial.

Let $\mathcal{E}\in\mathbb{R}^{p\times p}$ be a perturbation with $\lVert\mathcal{E}\rVert\le\totalbound$, and let $\epsilon,\eta>0$ denote a projection tolerance and a subspace-drift parameter; their algebraic relations are fixed by the standing assumptions~\eqref{eq:amp-standing} below.

The amplification analysis tracks two frames against $\mat{U}_{1:k}$ and $\mat{U}_{k+1:p}$: a signal frame $\mat{W}\in\stiefel{k}{p}$ close to $\mat{U}_{1:k}$ and a complement frame $\mat{N}\in\stiefel{p-k}{p}$ close to $\mat{U}_{k+1:p}$.
Chained across $L$ blocks, $\mat{M}(\ell)$ and $\mathcal{E}(\ell)$ are the per-block signal matrix and perturbation, $\mat{U}_{1:k}(\ell), \mat{U}_{k+1:p}(\ell)$ their leading and trailing singular vectors, $\{\mat{W}^{(\ell)}\}_{\ell=1}^{L+1}$ and $\{\mat{N}^{(\ell)}\}_{\ell=1}^{L+1}$ the per-block frame sequences, $\mathcal{M}^{(L)}=\prod_{\ell=1}^{L}(\mat{M}(\ell)+\mathcal{E}(\ell))$ the $L$-block product, and $\widehat{\mat{U}}_{1:k}(\ell)$ the running estimate of $\mat{U}_{1:k}(\ell)$.

\subsection{Relation to the existing stationary proof}

\paragraph{The stationary baseline.} With the signal matrix fixed at $\mat{M}$, the argument follows the convergence proof of the noisy power method of Hardt and Price~\citep{npm}.
It tracks the tangent of the largest principal angle between the iterate $\mat{\widehat{U}}(\ell)=b\big((\mat{M}+\mathcal{E})\mat{\widehat{U}}(\ell-1)\big)$ and $\mat{U}_{1:k}$.
By Lemma~\ref{lem:subspace_geom}\,\ref{sg:dist}, the complement projection $\|\mat{U}_{k+1:p}^\top\mat{\widehat{U}}(\ell)\|$ is the sine $\sin\theta$ of that angle and the least singular value $s_k(\mat{U}_{1:k}^\top\mat{\widehat{U}}(\ell))$ of the signal projection is its cosine $\cos\theta$, so the ratio $\rho(\ell)=\|\mat{U}_{k+1:p}^\top\mat{\widehat{U}}(\ell)\|/s_k(\mat{U}_{1:k}^\top\mat{\widehat{U}}(\ell))$ is the tangent $\tan\theta$.
One multiplication scales the signal part of a direction by at least the $k$-th singular value $s_k(\mat{M})$ and the complement part by at most the $(k{+}1)$-th, $s_{k+1}(\mat{M})$, each up to the stochastic error $\totalbound$ (Lemma~\ref{lem:floor_ceiling}).
Dividing the complement ceiling by the signal floor bounds the next ratio.
Under the two-part bound below, this ratio obeys the noise-free recursion of~\citep[Lemma~2.2]{npm}, clamped at the projection tolerance $\epsilon=4\totalbound/\delta$ of~\eqref{eq:amp-standing}:
\begin{equation*}
\rho(\ell)\;\le\;\max\!\Big(\epsilon,\ \max\!\big(\epsilon,\ (s_{k+1}(\mat{M})/s_k(\mat{M}))^{1/4}\big)\,\rho(\ell-1)\Big).
\end{equation*}
The two-part bound controls the signal-direction component $\|\mat{U}_{1:k}^\top\mathcal{E}\|$, the part of the perturbation along the fixed signal subspace, separately from the full norm $\|\mathcal{E}\|$.
The first can be far smaller when $\mathcal{E}$ is nearly orthogonal to $\mat{U}_{1:k}$, and that projection-aware control is the refinement the moving target removes below.
The fourth root comes from the budget.
The two-part bound admits a signal-direction perturbation up to the quarter-gap $\Delta=\delta/4$ and a full-norm perturbation $\totalbound=\Delta\epsilon$, so $\epsilon=4\totalbound/\delta$ is the noise level fixing the clamp.
Together these leave, above the floor, the contraction factor $s_{k+1}(\mat{M})/(s_{k+1}(\mat{M})+\Delta)$.
Four quarter-gaps fill the gap, $s_{k+1}(\mat{M})+4\Delta=s_k(\mat{M})$.
Since $\big(a/(a+\Delta)\big)^4\le a/(a+4\Delta)$ for all $a,\Delta>0$, equivalently $(a+\Delta)^4-a^3(a+4\Delta)=6a^2\Delta^2+4a\Delta^3+\Delta^4\ge0$, the fourth power of the factor is at most $s_{k+1}(\mat{M})/s_k(\mat{M})$, giving $(s_{k+1}(\mat{M})/s_k(\mat{M}))^{1/4}$.
The recursion composes to the floor~\citep[Theorem~2.3]{npm} from a random start $\rho(0)=O(\tau_0\sqrt{pk})$~\citep[Lemma~2.4]{npm}, because the projectors are the same at every step: $\mat{U}_{k+1:p}^\top\mat{M}=s_{k+1}(\mat{M})\,\mat{U}_{k+1:p}^\top$, so the complement is invariant.

\paragraph{What drift removes.} Drift moves the target: $\mat{U}_{1:k}(\ell)$ rotates between blocks, so the two projectors change from block to block, and the stationary proof gives way at two points.
\begin{itemize}
\item \emph{The recursion stops composing.} The per-vector floor and ceiling (Lemma~\ref{lem:floor_ceiling}) still hold inside a block, against the target of that block.
Across blocks the target rotates by $\eta$ (the drift bound~\eqref{eq:amp-drift}), so a slice of size $\eta$ of the previous signal is reclassified as complement against the new target.
The per-step bound drives the old complement toward the floor, but this slice is reinjected fresh each block and is not contracted with it, so the bound no longer composes to a small floor.
What fails is not the fourth-root absorption but the chaining: against a moving target the iterate recursion does not compose across blocks, so there is no fixed-target recursion to absorb into.
The frames below take over that role, with the per-block factor $\beta$ of Lemma~\ref{lem:thm2ratio}, whose $(1-(\epsilon+\eta)^2)$ factor is the drift tilt, as the absorbed factor.
\item \emph{The sharp perturbation bound fails.} The signal-direction component $\|\mat{U}_{1:k}^\top\mathcal{E}\|$ is measured against the rotating $\mat{U}_{1:k}$, so it is not preserved as the per-block perturbations compose.
Only the full operator norm $\|\mathcal{E}(\ell)\|\le\totalbound$, whose stochastic and drift parts are bounded in the preceding subsections, survives the product.
\end{itemize}
The construction below works against a fixed reference instead, adding the perturbation at a coarser constant than the stationary bound but one that survives the moving target.

\subsection{The two-frame construction}

\begin{figure}[t]
\centering
\resizebox{\columnwidth}{!}{%
\begin{tikzpicture}[
  >=Stealth, font=\footnotesize,
  io/.style ={draw, rounded corners, align=center, inner sep=4pt},
  amp/.style={draw, rounded corners, align=center, inner sep=4pt, fill=gray!10},
  reg/.style={draw, rounded corners, align=center, inner sep=4pt, fill=gray!4},
  blk/.style={draw, rounded corners, align=center, inner sep=4pt, fill=gray!25, font=\small\bfseries},
  uh/.style ={draw, rounded corners, align=center, inner sep=5pt, fill=gray!4, text width=13cm},
  rt/.style ={draw, rounded corners, align=center, inner sep=3pt, fill=white, font=\scriptsize},
  nb/.style ={draw, dashed, rounded corners, align=center, inner sep=3pt, font=\scriptsize, fill=white},
  ar/.style ={->, thick},
  cpl/.style={densely dotted},
]
  \node[reg] (reg) at (5.7,3.9)
    {Regime~$(\star)$ governing the selection of $\mat{N}^{(\ell)},\mat{W}^{(\ell)}$:\\[1pt]
     $\epsilon=4\totalbound/\delta,\quad \totalbound\le\delta/16,\quad \eta\le\epsilon/5,\quad \delta\ge\tfrac{12}{17}\sigma^2$};
  \node[blk] (blk) at (5.7,2.55) {block $\ell$:\ \ $\times(\mat{M}(\ell){+}\mathcal{E}(\ell))$};
  \draw[ar, dashed] (reg) -- (blk);
  \node[io]  (Win)  at (0,1.4)    {$\mat{W}^{(\ell)}$ {\scriptsize(grown forward)}\\[1pt]$d(\mat{U}_{1:k},\mat{W}^{(\ell)})\le\epsilon{+}\eta$};
  \node[amp] (Wamp) at (5.7,1.4)  {signal amplification \;[Prop.~\ref{lem:proj_nonstat}]\\[1pt]$s_k\big((\mat{M}{+}\mathcal{E})\mat{W}^{(\ell)}\big)\ge\sqrt{1{-}(\epsilon{+}\eta)^2}\,(\delta{+}\sigma^2)-\totalbound$};
  \node[io]  (Wout) at (11.4,1.4) {$\mat{W}^{(\ell+1)}$\\[1pt]$d(\mat{U}_{1:k},\mat{W}^{(\ell+1)})\le\epsilon$};
  \node[io]  (Nin)  at (0,-1.4)   {$\mat{N}^{(\ell)}$ {\scriptsize(produced)}\\[1pt]$\|\mat{U}_{1:k}^{\top}\mat{N}^{(\ell)}\|\le\epsilon$};
  \node[amp] (Namp) at (5.7,-1.4) {orthogonal amplification \;[Prop.~\ref{lem:orth_nonstat}]\\[1pt]$\big\|(\mat{M}{+}\mathcal{E})\mat{N}^{(\ell)}\big\|\le\dfrac{\sigma^2+\totalbound}{\sqrt{1{-}(\epsilon{+}\eta)^2}}$};
  \node[io]  (Nout) at (11.4,-1.4){$\mat{N}^{(\ell+1)}$ {\scriptsize(target)}\\[1pt]$\|\mat{U}_{1:k}^{\top}\mat{N}^{(\ell+1)}\|\le\epsilon{+}\eta$};
  \node[rt] (rt) at (5.7,0) {$\dfrac{s_k(\text{signal})}{\|\text{complement}\|}\ge\beta>1$ \;[Lem.~\ref{lem:thm2ratio}]};
  \draw[ar] (Win) -- (Wamp);  \draw[ar] (Wamp) -- (Wout);
  \draw[ar] (Nout) -- (Namp);  \draw[ar] (Namp) -- (Nin);
  \draw[cpl] (Wamp) -- (rt);  \draw[cpl] (Namp) -- (rt);
  \node[nb] (nb) at (14.9,0) {between blocks $\ell,\ell{+}1$\\[1pt]{\scriptsize target rotates by $\eta$}\\{\scriptsize so $\epsilon\!\leftrightarrow\!\epsilon{+}\eta$}};
  \draw[ar, dashed] (Wout.east) -- (nb.north west);
  \draw[ar, dashed] (nb.south west) -- (Nout.east);
  \node[uh] (uh) at (5.7,-3.4)
    {$\mat{\widehat{U}}_{1:k}$: its projection on $\mat{W}$ grows ($\times s_k$) while its projection on $\mat{N}$ stays capped ($\times\sigma^2$), so it rotates onto $\mat{W}$.
    Hence $\mat{\widehat{U}}_{1:k}(0)$ random, and after $L$ blocks $\mat{\widehat{U}}_{1:k}(L)\approx\mat{W}^{(L+1)}\approx\mat{U}_{1:k}(L)$ \;[Thm.~\ref{thm:drifting-power-method}].};
  \begin{scope}[on background layer]
    \node[draw, dashed, rounded corners, inner sep=9pt, fit=(Win)(Wout)(Nin)(Nout)(rt)] {};
  \end{scope}
\end{tikzpicture}%
}
\caption{\textbf{One block of the construction in Figure~\ref{fig:frame-chain}.} Under the regime~$(\star)$, block~$\ell$ enters a signal frame $\mat{W}^{(\ell)}$ and a complement anchor $\mat{N}^{(\ell+1)}$, both within $\epsilon{+}\eta$ of the leading and trailing subspaces of $\mat{M}(\ell)$. One multiplication by $\mat{M}(\ell)+\mathcal{E}(\ell)$ keeps the signal large (Proposition~\ref{lem:proj_nonstat}) and the complement capped at the noise level (Proposition~\ref{lem:orth_nonstat}), so their ratio grows by $\beta>1$ (Lemma~\ref{lem:thm2ratio}); it tightens each frame from $\epsilon{+}\eta$ to $\epsilon$, growing the signal \emph{forward} into $\mat{W}^{(\ell+1)}$ (left to right) and pulling the complement \emph{backward} into $\mat{N}^{(\ell)}$ (right to left). The tolerance reopens from $\epsilon$ to $\epsilon{+}\eta$ because the target subspace itself rotates between blocks: the per-block drift $B\Gamma$ separates $\mat{U}_{1:k}(\ell)$ and $\mat{U}_{1:k}(\ell{+}1)$ by $\eta\approx B\Gamma/\delta$ via Davis--Kahan. The signal is seeded freely at $\mat{W}^{(1)}$ and the complement anchored at $\mat{N}^{(L+1)}=\mat{U}_{k+1:p}(L)$. The random iterate $\mat{\widehat{U}}_{1:k}$ keeps a growing projection on $\mat{W}$ and a capped one on $\mat{N}$, so over the $L$ blocks it locks onto $\mat{W}^{(L+1)}$, which is $\epsilon$-close to the true subspace $\mat{U}_{1:k}(L)$ (Theorem~\ref{thm:drifting-power-method}).}
\label{fig:amplification}
\end{figure}

\paragraph{Why two frames.} Drift removed the one fixed pair of projectors: the true eigenspaces $\mat{U}_{1:k}(\ell)$ and $\mat{U}_{k+1:p}(\ell)$ turn with every block, so a per-vector bound stated against them says nothing about the next.
We rebuild a fixed reference by hand as two co-moving frames, the signal $\mat{W}^{(\ell)}$ and the complement $\mat{N}^{(\ell)}$, chained so that the image of one block falls in the next frame and the per-vector floor and ceiling (Lemma~\ref{lem:floor_ceiling}) compose into a single product.
\begin{itemize}
\item \emph{The complement frame $\mat{N}^{(\ell)}$} is anchored at the terminal block, $\mat{N}^{(L+1)}=\mat{U}_{k+1:p}(L)$, the only complement subspace known in advance.
It is built backward, so that one multiplication sends each range into the next, $\mathrm{ran}\big((\mat{M}(\ell){+}\mathcal{E}(\ell))\mat{N}^{(\ell)}\big)\subseteq\mathrm{ran}\,\mat{N}^{(\ell+1)}$ (condition~\ref{c2}); the ceiling of Proposition~\ref{lem:orth_nonstat} then composes block by block into the single product $\|\mathcal{M}^{(L)}\mat{N}^{(1)}\|$ of~\eqref{eq:sk1}.
\item \emph{The signal frame $\mat{W}^{(\ell)}$} is seeded freely at the first block, $\mat{W}^{(1)}$ taken orthogonal to $\mat{N}^{(1)}$, and grown forward as the orthonormalized image $\mat{W}^{(\ell+1)}=\mathrm{GS}\big((\mat{M}(\ell){+}\mathcal{E}(\ell))\mat{W}^{(\ell)}\big)$.
Each multiplication amplifies the signal over the complement, so a frame at distance $\epsilon+\eta$ from $\mat{U}_{1:k}(\ell)$ has image within $\epsilon$ of it.
So the terminal $\mat{W}^{(L+1)}$ lies within $\epsilon$ of $\mat{U}_{1:k}(L)$, and the floor of Proposition~\ref{lem:proj_nonstat} composes into the matching product $s_k(\mathcal{M}^{(L)}\mat{W}^{(1)})$ of~\eqref{eq:sk2}.
\end{itemize}
The forms are opposite because the bounds are opposite.
A ceiling needs only a range inclusion: where the image falls inside $\mathrm{ran}\,\mat{N}^{(\ell+1)}$ is never used, so $\mat{N}^{(\ell)}$ need only exist.
A floor on $s_k$ is a lower bound, attainable only by tracking the exact image forward.
The two products differ by $\beta$ per block (Lemma~\ref{lem:thm2ratio}), hence by $\beta^L$ over the $L$ blocks.
Decomposing the random iterate onto $\mat{W}^{(1)}$ and $\mat{N}^{(1)}$ at the first block and squeezing it between them yields the $\beta^{-L}$ contraction to distance $\epsilon$.

Figure~\ref{fig:amplification} traces one block: a frame enters at the loose tolerance $\epsilon+\eta$, opened by the target rotating $\eta$ between blocks, and one multiplication tightens it back to $\epsilon$.
Both frames follow the same cycle, time-reversed.
Drift opens each by the one rotation $\eta$, and one multiplication closes it.
But $\mat{W}$ grows forward to the terminal $\mat{U}_{1:k}(L)$ while $\mat{N}$ is anchored backward at $\mat{U}_{k+1:p}(L)$, so read forward the signal frame closes $\epsilon+\eta\to\epsilon$ exactly as the complement frame opens $\epsilon\to\epsilon+\eta$.

The recurring factor $\sqrt{1-(\epsilon+\eta)^2}$ is the cosine of that tolerance, the mass a distance-$(\epsilon+\eta)$ frame keeps on the true axis.
It discounts the signal floor (Proposition~\ref{lem:proj_nonstat}) and inflates the complement ceiling (Proposition~\ref{lem:orth_nonstat}) by that one factor, so their ratio $\beta$ (Lemma~\ref{lem:thm2ratio}) has its square, $1-(\epsilon+\eta)^2$.
The residual floor $\epsilon=4\totalbound/\delta$ is where the reopening and the tightening balance.
Figure~\ref{fig:frame-chain} shows the two frames and the order in which they are built.

\begin{figure}[t]
\centering
\resizebox{\columnwidth}{!}{%
\begin{tikzpicture}[
  >=Stealth, font=\footnotesize,
  sigfr/.style={->, very thick, blue!70!black},
  cmpfr/.style={->, very thick, orange!85!black, densely dashed},
  sigax/.style={blue!45, semithick},
  cmpax/.style={orange!60, semithick, densely dashed},
  fwd/.style={->, semithick, blue!55},
  bwd/.style={->, semithick, orange!65, densely dashed},
  lbl/.style={align=center, inner sep=1.5pt},
  wlbl/.style={align=center, inner sep=1.5pt, text=blue!70!black},
  nlbl/.style={align=center, inner sep=1.5pt, text=orange!85!black},
]
  \def\R{0.85}
  \def\Rax{1.08}
  \begin{scope}[shift={(0,-0.3)}, rotate=10]
     \draw[sigax] (0,0) -- (90:\Rax);
     \draw[cmpax] (0,0) -- (0:\Rax);
     \draw[sigfr] (0,0) -- (84:\R);
     \draw[cmpfr] (0,0) -- (-6:\R);
     \draw[thick] (84:0.22) -- ++(-6:0.22) -- ++(264:0.22);
  \end{scope}
  \foreach \x/\a in {2.9/22, 8.1/46, 11.0/58} {
     \begin{scope}[shift={(\x,-0.3)}, rotate=\a]
        \draw[sigax] (0,0) -- (90:\Rax);
        \draw[cmpax] (0,0) -- (0:\Rax);
        \draw[sigfr] (0,0) -- (84:\R);
        \draw[cmpfr] (0,0) -- (6:\R);
     \end{scope}
  }
  \node at (5.5,-0.15) {$\cdots$};
  \node[anchor=south, font=\scriptsize] at (5.6,2.48)
     {signal floor:\ \ $s_k\!\big(\mathcal{M}^{(L)}\mat{W}^{(1)}\big)\ge\prod_{\ell}\big[\sqrt{1{-}(\epsilon{+}\eta)^2}\,s_k(\mat{M}(\ell))-\totalbound\big]$};
  \node[anchor=south, font=\scriptsize] at (5.6,1.98)
     {forward transition:\ \ $\mat{W}^{(\ell+1)}=\mathrm{GS}\big((\mat{M}(\ell){+}\mathcal{E}(\ell))\,\mat{W}^{(\ell)}\big)$ \,[Prop.~\ref{lem:proj_nonstat}]};
  \node[wlbl] at (0,1.5)    {$\mat{W}^{(1)}$};
  \node[wlbl] at (2.9,1.5)  {$\mat{W}^{(2)}$};
  \node[lbl]  at (5.5,1.5)  {$\cdots$};
  \node[wlbl] at (8.1,1.5)  {$\mat{W}^{(L)}$};
  \node[wlbl] at (11.3,1.5) {$\mat{W}^{(L+1)}{\approx}\mat{U}_{1:k}(L)$};
  \draw[fwd] (0.4,1.5) -- (2.45,1.5) node[midway, above, font=\tiny, inner sep=1.5pt] {$\times(\mat{M}{+}\mathcal{E}),\,\mathrm{GS}$};
  \draw[fwd] (3.35,1.5) -- (5.0,1.5) node[midway, above, font=\tiny, inner sep=1.5pt] {$\times(\mat{M}{+}\mathcal{E})$};
  \draw[fwd] (6.0,1.5) -- (7.6,1.5) node[midway, above, font=\tiny, inner sep=1.5pt] {$\times(\mat{M}{+}\mathcal{E})$};
  \draw[fwd] (8.6,1.5) -- (10.05,1.5) node[midway, above, font=\tiny, inner sep=1.5pt] {$\times(\mat{M}{+}\mathcal{E})$};
  \node[nlbl] at (0,-1.5)    {$\mat{N}^{(1)}$};
  \node[nlbl] at (2.9,-1.5)  {$\mat{N}^{(2)}$};
  \node[lbl]  at (5.5,-1.5)  {$\cdots$};
  \node[nlbl] at (8.1,-1.5)  {$\mat{N}^{(L)}$};
  \node[nlbl] at (11.3,-1.55) {$\mat{N}^{(L+1)}{=}\mat{U}_{k+1:p}(L)$};
  \draw[bwd] (2.45,-1.5) -- (0.4,-1.5) node[midway, below, font=\tiny, inner sep=1.5pt] {construct};
  \draw[bwd] (5.0,-1.5) -- (3.35,-1.5) node[midway, below, font=\tiny, inner sep=1.5pt] {construct};
  \draw[bwd] (7.6,-1.5) -- (6.0,-1.5) node[midway, below, font=\tiny, inner sep=1.5pt] {construct};
  \draw[bwd] (10.05,-1.5) -- (8.6,-1.5) node[midway, below, font=\tiny, inner sep=1.5pt] {construct};
  \node[anchor=north, font=\scriptsize] at (5.6,-1.98)
     {backward construction:\ \ $\mat{N}^{(\ell)}$ chosen so $\mathrm{ran}\big((\mat{M}(\ell){+}\mathcal{E}(\ell))\,\mat{N}^{(\ell)}\big)\subseteq\mathrm{ran}\,\mat{N}^{(\ell+1)}$ \,[Prop.~\ref{lem:orth_nonstat}]};
  \node[anchor=north, font=\scriptsize] at (5.6,-2.48)
     {complement ceiling:\ \ $\|\mathcal{M}^{(L)}\mat{N}^{(1)}\|\le\prod_{\ell}\big(s_{k+1}(\mat{M}(\ell))+\totalbound\big)/\sqrt{1{-}(\epsilon{+}\eta)^2}$};
  \draw[->, thick] (-0.95,-1.5) -- (-0.95,1.5);
  \node[lbl, anchor=east] at (-1.05,0) {{\scriptsize$\mat{W}^{(1)}\!\perp\!\mat{N}^{(1)}$}\\[-1pt]{\scriptsize(W.1)}};
\end{tikzpicture}%
}
\caption{\textbf{The two co-moving frames and the build order.} From the one complement frame known in advance, the terminal trailing eigenspace $\mat{N}^{(L+1)}=\mat{U}_{k+1:p}(L)$, the complement frames are built \emph{backward} (bottom chain), $\mat{W}^{(1)}$ is taken orthogonal to $\mat{N}^{(1)}$ at the left turn (W.1), and the signal frames grow \emph{forward} (top chain) to $\mat{W}^{(L+1)}\approx\mat{U}_{1:k}(L)$. The two frames sandwich the random iterate between the signal floor on $\mat{W}$ and the complement ceiling on $\mat{N}$, which is what turns the per-block bounds into the $\beta^{-L}$ contraction. Blue marks the signal side (thin axis $\mat{U}_{1:k}$, bold frame $\mat{W}$) and orange the complement side (thin axis $\mat{U}_{k+1:p}$, bold frame $\mat{N}$); in each block the bold frames lie within $\epsilon{+}\eta$ and $\epsilon$ of the axes, exactly orthogonal only at the first block (right angle).}
\label{fig:frame-chain}
\end{figure}

The lemmas and propositions below are stated for a single positive definite matrix $\mat{M}$ and a single bounded perturbation $\mathcal{E}$, under explicit conditions on the gap, the perturbation size, and the drift.
The theorem then chains them across blocks, and each convergence proof instantiates these conditions for its own per-block signal matrix.

\subsection{Standing assumptions}
We assume throughout this appendix that
\begin{equation}
\label{eq:amp-standing}
\epsilon=\frac{4\totalbound}{\delta},\qquad
\totalbound\le\frac{\delta}{16},\qquad
\eta\le\frac{\epsilon}{5},\qquad
\delta\ge\frac{12}{17}\,\sigma^2 ,
\end{equation}
so that $\epsilon\le\tfrac14$ and $\epsilon+\eta\le\tfrac{6}{5}\epsilon\le\tfrac{3}{10}$.

Concretely, three roles are distinguished:

\noindent(i) $\epsilon$ bounds the post-multiply projection distance: $d(\mat{U}_{1:k}, (\mat{M}+\mathcal{E})\mat{W}) \le \epsilon$ when a signal frame enters at $d(\mat{U}_{1:k}, \mat{W}) \le \epsilon+\eta$ (Proposition~\ref{lem:proj_nonstat}).

\noindent(ii) The parameter $\eta$ bounds the per-block leading-subspace drift: $\lVert\mat{U}_{k+1:p}(\ell)^\top \mat{U}_{1:k}(\ell-1)\rVert \le \eta$ between consecutive blocks (Theorem~\ref{thm:drifting-power-method}'s hypothesis~\eqref{eq:amp-drift}).

\noindent(iii) The same $\epsilon$ is the residual error in Theorem~\ref{thm:drifting-power-method}'s $L$-block bound: $d(\mat{U}_{1:k}(L), \widehat{\mat{U}}_{1:k}(L)) \le \epsilon + O(\beta^{-L}\tau_0\sqrt{pk})$.

\subsection{Statements}

The leading and trailing blocks obey one geometric identity used throughout.
\begin{lemma}[Subspace geometry]
\label{lem:subspace_geom}
Let $\mat{U}_{1:k}$ and $\mat{U}_{k+1:p}$ be the leading and trailing blocks of an orthogonal matrix in $\mathbb{R}^{p\times p}$.
\begin{enumerate}[label=\textup{(\roman*)},ref=\textup{(\roman*)}]
\item\label{sg:pyth} For every $\vect{x}\in\mathbb{R}^{p}$, $\;\|\mat{U}_{1:k}^{\top}\vect{x}\|^{2}+\|\mat{U}_{k+1:p}^{\top}\vect{x}\|^{2}=\|\vect{x}\|^{2}$.
\item\label{sg:tilt} If $\|\mat{U}_{1:k}^{\top}\vect{x}\|\le\rho\,\|\vect{x}\|$ with $\rho\in[0,1)$, then $\|\vect{x}\|\le\|\mat{U}_{k+1:p}^{\top}\vect{x}\|\big/\sqrt{1-\rho^{2}}$.
\item\label{sg:dist} For every $\mat{V}\in\stiefel{k}{p}$, $\;d(\mat{U}_{1:k},\mat{V})=\sqrt{1-s_k(\mat{U}_{1:k}^{\top}\mat{V})^{2}}$.
\end{enumerate}
\end{lemma}

One per-vector estimate underlies the two amplification propositions below.
Multiplying any unit direction by $\mat{M}+\mathcal{E}$ scales its signal part by at least $s_k(\mat{M})$ and its complement part by at most $s_{k+1}(\mat{M})$, each up to the perturbation $\totalbound$.
\begin{lemma}[Per-vector amplification]
\label{lem:floor_ceiling}
Under the standing assumptions~\eqref{eq:amp-standing}, for every unit vector $\vect{u}\in\mathbb{R}^{p}$,
\begin{equation*}
\|\mat{U}_{1:k}^{\top}(\mat{M}+\mathcal{E})\vect{u}\|\ge s_k(\mat{M})\,\|\mat{U}_{1:k}^{\top}\vect{u}\|-\totalbound\,,\qquad
\|\mat{U}_{k+1:p}^{\top}(\mat{M}+\mathcal{E})\vect{u}\|\le s_{k+1}(\mat{M})\,\|\mat{U}_{k+1:p}^{\top}\vect{u}\|+\totalbound\,.
\end{equation*}
\end{lemma}

\begin{corollary}[Ratio after one multiplication]
\label{cor:ratio}
Under the standing assumptions~\eqref{eq:amp-standing}, for every unit vector $\vect{u}\in\mathbb{R}^{p}$ with $\|\mat{U}_{1:k}^{\top}\vect{u}\|\ge\sqrt{1-(\epsilon+\eta)^2}$, the signal floor $s_k(\mat{M})\,\|\mat{U}_{1:k}^{\top}\vect{u}\|-\totalbound$ is positive, and dividing the two bounds of Lemma~\ref{lem:floor_ceiling} gives
\begin{equation*}
\frac{\|\mat{U}_{k+1:p}^{\top}(\mat{M}+\mathcal{E})\vect{u}\|}{\|\mat{U}_{1:k}^{\top}(\mat{M}+\mathcal{E})\vect{u}\|}\le\frac{s_{k+1}(\mat{M})\,\|\mat{U}_{k+1:p}^{\top}\vect{u}\|+\totalbound}{s_k(\mat{M})\,\|\mat{U}_{1:k}^{\top}\vect{u}\|-\totalbound}\,.
\end{equation*}
\end{corollary}

Corollary~\ref{cor:ratio} tracks one scalar, the signal-to-complement ratio $\sqrt{1-d^2}/d$ of a direction at subspace distance $d$, and shows that one multiplication by $\mat{M}+\mathcal{E}$ amplifies it by about $s_k(\mat{M})/s_{k+1}(\mat{M})=(\delta+\sigma^2)/\sigma^2>1$.
Two effects oppose this gain.
The perturbation leaks $\totalbound$ into the numerator and denominator, and between blocks the target $\mat{U}_{1:k}(\ell)$ rotates by $\eta$, reopening the distance of each frame from the tight tolerance $\epsilon$ to the loose $\epsilon+\eta$.
The next lemma certifies that one multiplication still nets a gain every block, splitting it into a net gain above $1$, a ceiling that keeps the complement frame near its axis, and a floor that pulls the signal frame back to distance $\epsilon$.

\begin{lemma}[Ratio inequalities]
\label{lem:thm2ratio}
Under the standing assumptions~\eqref{eq:amp-standing}, \ref{ri:beta} the net gain $\beta$ exceeds $1$, giving the $\beta^{-L}$ contraction of Theorem~\ref{thm:drifting-power-method}; \ref{ri:ratio2} the complement ceiling holds $\|\mat{U}_{1:k}^{\top}\mat{N}\|\le\epsilon$ (Proposition~\ref{lem:orth_nonstat}); and \ref{ri:ratio3} the signal floor gives $d(\mat{U}_{1:k},(\mat{M}+\mathcal{E})\mat{W})\le\epsilon$ (Proposition~\ref{lem:proj_nonstat}):
\begin{enumerate}[label=\textup{(\roman*)},ref=\textup{(\roman*)}]
    \item\label{ri:beta} $\beta := (1-(\eta+\epsilon)^2)\frac{\delta + \sigma^2 - {\totalbound}/{\sqrt{1-(\eta+\epsilon)^2}}}{\sigma^2 + \totalbound} \geq 1.4464 > \frac{1}{0.7} >1$, so the per-step contraction satisfies $\beta^{-1}<0.7\,.$
    \item\label{ri:ratio2} $\frac{\epsilon}{\sqrt{1-\epsilon^2}} \frac{s_k(\mat{M})   - \totalbound/\epsilon }{ s_{k+1}(\mat{M})  + \totalbound/\sqrt{1-\epsilon^2}}\geq \frac{\epsilon}{\sqrt{1-\epsilon^2}} \frac{0.75 s_k(\mat{M}) + 0.25 s_{k+1}(\mat{M}) }{0.25  s_{k}(\mat{M})+  0.75 s_{k+1}(\mat{M})} \stackrel{(\spadesuit)}{>} \frac{\epsilon + \eta}{\sqrt{1-(\epsilon + \eta)^2}}. $
    \item\label{ri:ratio3} $\frac{s_k
  (\mat{M}) -\totalbound/\sqrt{1-(\epsilon+\eta)^2}}{s_{k+1}(\mat{M})+\totalbound/(\epsilon+\eta)}\frac{\sqrt{1-(\epsilon+\eta)^2}}{\epsilon+\eta}>\frac{\sqrt{1-\epsilon^2}}{\epsilon}\,.$
\end{enumerate}
\end{lemma}

With the ratio margins in hand, we turn to the two frame bounds.
The first control is a ceiling on the complement.
The next proposition shows that any subspace whose overlap with the signal is at most $\epsilon+\eta$ is mapped by $\mat{M}+\mathcal{E}$ back into a near-complement, with norm at most $(\sigma^2+\totalbound)/\cos\theta$.
This is the noise level $\sigma^2+\totalbound$ inflated by the tilt factor $1/\cos\theta$, where $\theta$ denotes the largest principal angle at subspace distance $\epsilon+\eta$, so that $\sin\theta=\epsilon+\eta$ and $\cos\theta=\sqrt{1-(\epsilon+\eta)^2}$.
A near-complement direction has little signal energy.
Hence the multiply $\mat{M}+\mathcal{E}$ cannot lift it above the noise level $\sigma^2+\totalbound$, which in turn caps how fast the trailing subspace can grow.

\begin{proposition}[Orthogonal amplification]
\label{lem:orth_nonstat}
Under the standing assumptions~\eqref{eq:amp-standing}, let $\mathcal{Y}$ be the set of $\mat{Y} \in$ $\stiefel{p-k}{p}$ such that $\left\|\mat{U}_{1:k}^{\top}\mat{Y} \right\| \le \epsilon + \eta$.
For every given $\mat{Y}\in \mathcal{Y}$, there exists a $\mat{N}\in\stiefel{p-k}{p}$ such that
\begin{align}
&\mathrm{ran}\left((\mat{M}+\mathcal{E})\mat{N}\right)  \subseteq \mathrm{ran} \left(\mat{Y} \right),~ \left\|\mat{U}_{1:k}^{\top} \mat{N} \right\| \le \epsilon ,\label{eq:conAZ}\\& \left\| (\mat{M}+\mathcal{E})\mat{N} \right\| \le \frac{\sigma^{2}+\totalbound}{\sqrt{1-(\epsilon+\eta )^2}}. \label{eq:conAZ2}
\end{align}
\end{proposition}

The second control is a floor on the signal.
On a subspace close to the truth the effect is complementary: one multiply by $\mat{M}+\mathcal{E}$ keeps its $k$-th singular value at least $\cos\theta\,(\delta+\sigma^2)-\totalbound$ and pulls it from distance $\epsilon+\eta$ down to $\epsilon$.
Here the signal directions are amplified by roughly $s_k(\mat{M})$ while the leaked perturbation is only $\totalbound$, so the signal both stays large and rotates toward $\mat{U}_{1:k}$.
\begin{proposition}[Signal amplification]
Under the standing assumptions~\eqref{eq:amp-standing}, let $\mat{W}\in\stiefel{k}{p}$.
When $d\left(\mat{U}_{1:k},\mat{W} \right) \le \epsilon +\eta$,
\begin{equation*}
s_k\left( (\mat{M}+\mathcal{E})\mat{W} \right) \ge \sqrt{1-(\epsilon+\eta)^2} (\delta+\sigma^{2}) -\totalbound\,,\:\:\mbox{and}\:\:d\left(\mat{U}_{1:k}, (\mat{M}+\mathcal{E})\mat{W} \right) \le \epsilon .
\end{equation*}
\label{lem:proj_nonstat}
\end{proposition}

Together Propositions~\ref{lem:orth_nonstat} and~\ref{lem:proj_nonstat} act as a rising floor and a falling ceiling.
At each step the tracked signal subspace grows while the complementary subspace stays capped, so their ratio, the signal floor over the complement ceiling, improves by the factor $\beta>1$.
Chaining this across the $L$ blocks shrinks the iterate's complement share like $\beta^{-L}$, squeezing it onto a subspace that is $\epsilon$-close to the truth, and yields the main convergence theorem of this appendix.
The noisy power method of Theorem~\ref{thm:robust-power-method} and Oja's method of Theorem~\ref{thm:ojas-algorithm} are its instantiations.

\begin{theorem}[Noisy power method under perturbation and drift]
\label{thm:drifting-power-method}
Let $\{\mat{M}(\ell)\}_{\ell=1}^{L}$ and $\{\mathcal{E}(\ell)\}_{\ell=1}^{L}$ satisfy the standing assumptions~\eqref{eq:amp-standing} blockwise with common $(\delta,\sigma^2,\totalbound,\epsilon,\eta)$, and write $\mat{U}_{1:k}(\ell)$ for the leading $k$ eigenvectors of $\mat{M}(\ell)$.
Assume the consecutive leading subspaces drift by at most $\eta$:
\begin{equation}
\label{eq:amp-drift}
\lVert \mat{U}_{k+1:p}(\ell)^\top \mat{U}_{1:k}(\ell-1)\rVert \le \eta,\qquad 2\le\ell\le L .
\end{equation}
Let $\mathcal{M}^{(L)}=\prod_{\ell=1}^{L}(\mat{M}(\ell)+\mathcal{E}(\ell))$, let $\mat{\widehat{U}}_{1:k}(0)\in\stiefel{k}{p}$ be a uniformly random orthonormal start, and let $\mat{\widehat{U}}_{1:k}(L)=b(\mathcal{M}^{(L)}\mat{\widehat{U}}_{1:k}(0))$.
With $\beta$ as in Lemma~\ref{lem:thm2ratio}, for any $\tau_0\ge 1$, if
\[
L>\frac{\log(\tau_0\sqrt{pk})}{\log\beta}\,,
\]
then on the event $\{\lVert\mathcal{E}(\ell)\rVert\le\totalbound\ \text{for all }\ell\}$, with probability at least $1-O(\tau_0^{-1})-e^{-\Omega(p)}$ over the initialization,
\[
d\big(\mat{U}_{1:k}(L),\mat{\widehat{U}}_{1:k}(L)\big)\le \epsilon+O(\beta^{-L}\tau_0\sqrt{pk}),
\]
where $\tau_0$ parameterizes the random-initialization tail bound of~\citep{npm}: larger $\tau_0$ gives a higher-confidence statement with a looser bound on the contraction prefactor.
\end{theorem}

\subsection{Proofs}

\begin{proof}[Proof of Lemma~\ref{lem:subspace_geom}]
Part (i) splits $\|\vect{x}\|^{2}=\|\mat{U}^{\top}\vect{x}\|^{2}$ across the two blocks.
For (ii), part (i) gives $\|\mat{U}_{k+1:p}^{\top}\vect{x}\|^{2}=\|\vect{x}\|^{2}-\|\mat{U}_{1:k}^{\top}\vect{x}\|^{2}\ge(1-\rho^{2})\|\vect{x}\|^{2}$.
For (iii), $d(\mat{U}_{1:k},\mat{V})=\|\mat{U}_{k+1:p}^{\top}\mat{V}\|$ is the largest principal-angle sine, and applying (i) to $\mat{V}\vect{f}$ for unit $\vect{f}$ gives $\|\mat{U}_{k+1:p}^{\top}\mat{V}\|^{2}=\max_{\vect{f}}\big(1-\|\mat{U}_{1:k}^{\top}\mat{V}\vect{f}\|^{2}\big)=1-s_k(\mat{U}_{1:k}^{\top}\mat{V})^{2}$.
\end{proof}

\begin{proof}[Proof of Lemma~\ref{lem:floor_ceiling}]
Write $\mat{M}=\mat{U}\mat{D}\mat{U}^{\top}$, and let $\mat{D}_{1:k},\mat{D}_{k+1:p}$ be the leading $k\times k$ and trailing $(p-k)\times(p-k)$ diagonal blocks of $\mat{D}$, so that
\begin{equation*}
\mat{U}_{1:k}^{\top}\mat{M}=\mat{D}_{1:k}\mat{U}_{1:k}^{\top}\,,\qquad \mat{U}_{k+1:p}^{\top}\mat{M}=\mat{D}_{k+1:p}\mat{U}_{k+1:p}^{\top}\,.
\end{equation*}
Since $s_k(\mat{M})$ is the smallest entry of $\mat{D}_{1:k}$ and $s_{k+1}(\mat{M})$ the largest of $\mat{D}_{k+1:p}$,
\begin{equation*}
\|\mat{U}_{1:k}^{\top}\mat{M}\vect{u}\|\ge s_k(\mat{M})\,\|\mat{U}_{1:k}^{\top}\vect{u}\|\,,\qquad
\|\mat{U}_{k+1:p}^{\top}\mat{M}\vect{u}\|\le s_{k+1}(\mat{M})\,\|\mat{U}_{k+1:p}^{\top}\vect{u}\|\,.
\end{equation*}
The two claims follow from the triangle inequality and $\|\mat{U}_{1:k}^{\top}\mathcal{E}\vect{u}\|,\,\|\mat{U}_{k+1:p}^{\top}\mathcal{E}\vect{u}\|\le\|\mathcal{E}\|\le\totalbound$.
\end{proof}

\begin{proof}[Proof of Lemma~\ref{lem:thm2ratio}]
All three inequalities follow from the standing assumptions; we prove them in order, with the third reusing the second's final step.
First note $\epsilon + \eta \leq 6\epsilon/5 \leq 3/10$.
For the first item:
\begin{align*}
\beta &{\geq} \frac{91}{100}  \frac{\delta + \sigma^2 - \sqrt{100/91}\,{\totalbound}}{\sigma^2 + \totalbound} {\geq}  \frac{91}{100}  \frac{\delta + \sigma^2 - \sqrt{100/91}\,{\delta}/16}{\sigma^2 + \delta/16} \\
&{=} \frac{91}{100} \frac{ \left(1 - \sqrt{\frac{100}{91}}\frac{1}{16} \right) + \frac{\sigma^2}{\delta}}{\frac{1}{16} + \frac{\sigma^2}{\delta}} {\geq} 1.4464\,.
\end{align*}
where the inequalities use $\epsilon + \eta\leq 3/10$, $\totalbound\leq\delta/16$, and $\delta\geq 12\sigma^2/17$, respectively.

For the second item, the first inequality is immediate from $\epsilon\leq 1/4$.
For $(\spadesuit)$, substitute $s_k(\mat{M})=\delta+\sigma^2$ and $s_{k+1}(\mat{M})=\sigma^2$ and divide by $\delta$, so its middle term equals $\frac{\epsilon}{\sqrt{1-\epsilon^2}}\frac{\sigma^2/\delta+3/4}{\sigma^2/\delta+1/4}$.
This fraction decreases in $\sigma^2/\delta$, and $\delta\geq\tfrac{12}{17}\sigma^2$ gives $\sigma^2/\delta\leq 17/12$, so it is at least $\frac{17/12+3/4}{17/12+1/4}=\frac{13}{10}$.
Hence $(\spadesuit)$ holds once
\begin{equation*}
\frac{13}{10}\frac{\epsilon}{\sqrt{1-\epsilon^2}}\geq \frac{\epsilon+\eta}{\sqrt{1-(\epsilon+\eta)^2}}
\quad\Longleftrightarrow\quad 1.69 \Big(\frac{\epsilon}{\epsilon + \eta} \Big)^2 - 1 \geq 0.69\,\epsilon^2 .
\end{equation*}
Since $\eta\leq\epsilon/5$ gives $\frac{\epsilon}{\epsilon+\eta}\geq\frac{5}{6}$, and $\epsilon\leq 1/4$,
\begin{equation*}
1.69 \Big(\frac{\epsilon}{\epsilon + \eta} \Big)^2 - 1 \geq 1.69\cdot\tfrac{25}{36}-1 > 0.17 \geq \tfrac{0.69}{16} \geq 0.69\,\epsilon^2 ,
\end{equation*}
which proves the second inequality.

For the third item, it suffices to show:
\begin{equation*}
\frac{\epsilon+\eta}{\sqrt{1-(\epsilon+\eta)^2}} \frac{\sigma^2 \:+\: {\delta \epsilon}/{4(\epsilon +\eta)}}{\sigma^2 + \delta \:-\: {\delta \epsilon}/{4 \sqrt{1 - (\epsilon + \eta)^2}}} \leq \frac{\epsilon}{\sqrt{1-\epsilon^2}}\,.
\end{equation*}
We bound the second factor on the left:
\begin{align*}
\frac{\sigma^2 \:+\: {\delta \epsilon}/{4(\epsilon +\eta)}}{\sigma^2 + \delta \:-\: {\delta \epsilon}/{4 \sqrt{1 - (\epsilon + \eta)^2}}} &= \frac{\sigma^2/\delta \:+\: { \epsilon}\big/{4(\epsilon +\eta)}}{\sigma^2/\delta + (1 \:-\: {\epsilon}\big/{4 \sqrt{1 - (\epsilon + \eta)^2}})}\\
&\stackrel{(\blacksquare)}{\leq} \frac{17/12 \:+\: { \epsilon}\big/{4(\epsilon +\eta)}}{17/12 + (1 \:-\: {\epsilon}\big/{4 \sqrt{1 - (\epsilon + \eta)^2}})}\\
&\stackrel{(\blacklozenge)}{\leq} \frac{17/12 \:+\: { \epsilon}\big/{4(\epsilon +\eta)}}{29/12 \:-\: {\epsilon}\big/{4(\epsilon +\eta)}}\\
&\leq \frac{17/3 \:+\: 1/{(1+\eta/\epsilon)}}{29/3 \:-\: 1/{(1+\eta/\epsilon)}}\leq \frac{10}{13}\,,
\end{align*}
where the $(\blacksquare)$ comes from:
\begin{align*}
\epsilon+\eta \leq \frac{1}{\sqrt{2}} \leq \sqrt{1-(\epsilon+\eta)^2}  
\Longrightarrow\:\:& \frac{\epsilon}{4}\left(\frac{1}{\epsilon+\eta} + \frac{1}{\sqrt{1-(\epsilon+\eta)^2}} \right) \leq \frac{\epsilon}{4}\frac{2}{\epsilon+\eta}\leq\frac{1}{2}\\
\Longrightarrow\:\:&\underline{1 -\frac{\epsilon}{4\sqrt{1-(\epsilon+\eta)^2}} \geq \frac{\epsilon}{4(\epsilon+\eta)} }\,,
\end{align*}
and the $(\blacklozenge)$ is immediate from $\epsilon+\eta \leq \frac{1}{\sqrt{2}} \leq \sqrt{1-(\epsilon+\eta)^2}$.
It remains to verify:
\begin{equation*}
\frac{10}{13} \frac{\epsilon+\eta}{\sqrt{1-(\epsilon+\eta)^2}} \leq \frac{\epsilon}{\sqrt{1-\epsilon^2}}\,,
\end{equation*}
which we already established (using $\eta\leq\epsilon/5$) in proving $(\spadesuit)$.
\end{proof}

\begin{proof}[Proof of Proposition~\ref{lem:orth_nonstat}]
The proof has three parts.
We first construct $\mat{N}$ with $\mathrm{ran}((\mat{M}+\mathcal{E})\mat{N})\subseteq\mathrm{ran}(\mat{Y})$; we then bound its signal overlap, $\|\mat{U}_{1:k}^{\top}\mat{N}\|\le\epsilon$, the second part of~\eqref{eq:conAZ}; and finally we bound its norm~\eqref{eq:conAZ2}.
For the construction, let $\mat{N}\in\stiefel{p-k}{p}$ be an orthonormal basis of any $(p-k)$-dimensional subspace of the preimage $(\mat{M}+\mathcal{E})^{-1}\mathrm{ran}(\mat{Y})=\{\vect{x}\in\mathbb{R}^{p}:(\mat{M}+\mathcal{E})\vect{x}\in\mathrm{ran}(\mat{Y})\}$; then $\mathrm{ran}((\mat{M}+\mathcal{E})\mat{N})\subseteq\mathrm{ran}(\mat{Y})$ at once.
Such a subspace exists because the preimage is large enough.
Restricted to the preimage, $\mat{M}+\mathcal{E}$ maps onto $\mathrm{ran}(\mat{M}+\mathcal{E})\cap\mathrm{ran}(\mat{Y})$ with kernel $\ker(\mat{M}+\mathcal{E})$, so with $r=\mathrm{rk}(\mat{M}+\mathcal{E})$ the rank--nullity theorem gives
\begin{equation*}
\dim\big[(\mat{M}+\mathcal{E})^{-1}\mathrm{ran}(\mat{Y})\big] = (p-r) + \dim\big[\mathrm{ran}(\mat{M}+\mathcal{E})\cap\mathrm{ran}(\mat{Y})\big] \ge (p-r)+\max(0,\,r-k) \ge p-k,
\end{equation*}
the intersection of the $r$- and $(p-k)$-dimensional ranges in $\mathbb{R}^p$ having dimension at least $r+(p-k)-p=r-k$.

Next we bound the signal overlap, the second part of~\eqref{eq:conAZ}, by contradiction: a unit direction in $\mathrm{ran}(\mat{N})$ with signal overlap above $\epsilon$ would be mapped outside $\mathrm{ran}(\mat{Y})$, contradicting $\mathrm{ran}((\mat{M}+\mathcal{E})\mat{N})\subseteq\mathrm{ran}(\mat{Y})$.
Concretely, we show:
\begin{equation*}
\mbox{If}\quad \vect{f} \in \mathrm{ran}(\mat{N}),\:\:\| \vect{f}\| =1 ,\:\: \mbox{and}\:\: \left\| \mat{U}_{1:k}^{\top}\vect{f}\right\| > \epsilon\,,\quad\mbox{then}\quad(\mat{M}+\mathcal{E})\vect{f} \notin
\mathrm{ran} \left(\mat{Y} \right)\,.
\end{equation*}
To show this, suppose $\left\| \mat{U}_{1:k}^{\top}\vect{f} \right\| > \epsilon$, so $\|\mat{U}_{k+1:p}^{\top}\vect{f}\|=\sqrt{1-\|\mat{U}_{1:k}^{\top}\vect{f}\|^2}<\sqrt{1-\epsilon^2}$.
\begin{itemize}
    \item $\displaystyle \|\mat{U}_{1:k}^\top (\mat{M}+\mathcal{E})\vect{f} \| \stackrel{(i)}{\ge} s_k(\mat{M})\,\|\mat{U}_{1:k}^{\top}\vect{f}\| - \totalbound > s_k(\mat{M}) \epsilon  - \totalbound$
    \item $\displaystyle \|\mat{U}_{k+1:p}^\top (\mat{M}+\mathcal{E})\vect{f} \|  \stackrel{(ii)}{\le} s_{k+1}(\mat{M})\,\|\mat{U}_{k+1:p}^{\top}\vect{f}\| + \totalbound \le s_{k+1}(\mat{M}) \sqrt{1-\epsilon^2} + \totalbound$
\end{itemize}
where $(i)$ and $(ii)$ follow from Lemma~\ref{lem:floor_ceiling}.
Dividing them, the signal-to-complement ratio of $(\mat{M}+\mathcal{E})\vect{f}$ obeys
\begin{equation*}
\frac{\|\mat{U}_{1:k}^\top (\mat{M}+\mathcal{E})\vect{f}\|}{\|\mat{U}_{k+1:p}^\top (\mat{M}+\mathcal{E})\vect{f}\|}
> \frac{s_k(\mat{M})\,\epsilon-\totalbound}{s_{k+1}(\mat{M})\sqrt{1-\epsilon^2}+\totalbound}
> \frac{\epsilon+\eta}{\sqrt{1-(\epsilon+\eta)^2}},
\end{equation*}
where the second inequality is Lemma~\ref{lem:thm2ratio}\,\ref{ri:ratio2} (whose left-hand side equals the middle term above).
In contrast, every unit-norm $\vect{v}\in\mathrm{ran}(\mat{Y})$ obeys $\|\mat{U}_{1:k}^\top\vect{v}\|/\|\mat{U}_{k+1:p}^\top\vect{v}\|\le(\epsilon+\eta)/\sqrt{1-(\epsilon+\eta)^2}$, because $\|\mat{U}_{1:k}^\top\mat{Y}\|\le\epsilon+\eta$.
As $(\mat{M}+\mathcal{E})\vect{f}$ has a strictly larger ratio than any vector in $\mathrm{ran}(\mat{Y})$, it cannot lie in $\mathrm{ran}(\mat{Y})$.
We can derive \eqref{eq:conAZ2} from \eqref{eq:conAZ} as follows: 
\begin{align*}
\left\| (\mat{M}+\mathcal{E})\mat{N} \right\| =& \sup_{\vect{y} \in \mathbb{R}^{p-k}:\|\vect{y} \| =1}\left\|(\mat{M}+\mathcal{E}) \mat{N}\vect{y} \right\| \\
\stackrel{(iii)}{\le}&\sup_{\vect{y} \in \mathbb{R}^{p-k}:\|\vect{y}\| =1} \frac{\left\|\mat{U}_{k+1:p}^\top (\mat{M}+\mathcal{E})\mat{N}\vect{y}\right\|}{\sqrt{1-(\epsilon+\eta)^2}}\le \frac{\left\|\mat{U}_{k+1:p}^\top (\mat{M}+\mathcal{E})\right\|}{\sqrt{1-(\epsilon+\eta)^2}}\\
\stackrel{(iv)}{\le} &\frac{s_{k+1}(\mat{M})+\totalbound}{\sqrt{1-(\epsilon+\eta )^2}}.
\end{align*}
Step $(iii)$ applies Lemma~\ref{lem:subspace_geom}\,\ref{sg:tilt} to $\vect{x}=(\mat{M}+\mathcal{E})\mat{N}\vect{y}$: since $\vect{x}\in\mathrm{ran}(\mat{Y})$ and $\|\mat{U}_{1:k}^{\top}\mat{Y}\|\le\epsilon+\eta$, we have $\|\mat{U}_{1:k}^{\top}\vect{x}\|\le(\epsilon+\eta)\|\vect{x}\|$, so $\|\vect{x}\|\le\|\mat{U}_{k+1:p}^{\top}\vect{x}\|/\sqrt{1-(\epsilon+\eta)^2}$.
Step $(iv)$ uses $\left\|\mat{U}_{k+1:p}^\top (\mat{M}+\mathcal{E})\right\| \le s_{k+1}(\mat{M})+\totalbound$, the supremum over unit directions of the ceiling bound in Lemma~\ref{lem:floor_ceiling}.
\end{proof}

\begin{proof}[Proof of Proposition~\ref{lem:proj_nonstat}]
We first record two per-vector bounds.
For every unit $\vect{f}\in\mathbb{R}^{k}$, the unit vector $\mat{W}\vect{f}$ has $\|\mat{U}_{1:k}^{\top}\mat{W}\vect{f}\|\ge s_k(\mat{U}_{1:k}^{\top}\mat{W})=\sqrt{1-d(\mat{U}_{1:k},\mat{W})^2}\ge\sqrt{1-(\epsilon+\eta)^2}$ and $\|\mat{U}_{k+1:p}^{\top}\mat{W}\vect{f}\|\le d(\mat{U}_{1:k},\mat{W})\le\epsilon+\eta$ (Lemma~\ref{lem:subspace_geom}\,\ref{sg:dist}), so Lemma~\ref{lem:floor_ceiling} with $\vect{u}=\mat{W}\vect{f}$ gives
\begin{equation}
\label{eq:sigamp-fc}
\|\mat{U}_{1:k}^{\top}(\mat{M}+\mathcal{E})\mat{W}\vect{f}\|\ge s_k(\mat{M})\sqrt{1-(\epsilon+\eta)^2}-\totalbound\,,\qquad
\|\mat{U}_{k+1:p}^{\top}(\mat{M}+\mathcal{E})\mat{W}\vect{f}\|\le s_{k+1}(\mat{M})(\epsilon+\eta)+\totalbound\,.
\end{equation}

First, since $\|(\mat{M}+\mathcal{E})\mat{W}\vect{f}\|\ge\|\mat{U}_{1:k}^{\top}(\mat{M}+\mathcal{E})\mat{W}\vect{f}\|$, taking the minimum over unit $\vect{f}$ of the floor in~\eqref{eq:sigamp-fc} bounds the $k$-th singular value:
\begin{equation*}
s_k\left( (\mat{M}+\mathcal{E})\mat{W} \right) = \min_{\|\vect{f}\|=1}\|(\mat{M}+\mathcal{E})\mat{W}\vect{f}\| \ge s_k(\mat{M})\sqrt{1-(\epsilon+\eta)^2}-\totalbound \ge \sqrt{1-(\epsilon+\eta)^2}(\delta+\sigma^2)-\totalbound\,.
\end{equation*}

Next we show $d\left(\mat{U}_{1:k},(\mat{M}+\mathcal{E})\mat{W} \right) \le \epsilon$.
Since $(\mat{M}+\mathcal{E})\mat{W}$ and its orthonormal basis $b((\mat{M}+\mathcal{E})\mat{W})$ span the same subspace, Lemma~\ref{lem:subspace_geom}\,\ref{sg:dist} gives $d(\mat{U}_{1:k},(\mat{M}+\mathcal{E})\mat{W})=\sqrt{1-s_k(\mat{U}_{1:k}^{\top}\,b((\mat{M}+\mathcal{E})\mat{W}))^2}$, so $d\le\epsilon$ is equivalent to $s_k(\mat{U}_{1:k}^{\top}\,b((\mat{M}+\mathcal{E})\mat{W}))^{2}\ge 1-\epsilon^{2}$.
Parametrizing the unit vectors of $\mathrm{ran}((\mat{M}+\mathcal{E})\mat{W})$ as $(\mat{M}+\mathcal{E})\mat{W}\vect{f}/\|(\mat{M}+\mathcal{E})\mat{W}\vect{f}\|$ and splitting the norm by Lemma~\ref{lem:subspace_geom}\,\ref{sg:pyth},
\begin{equation*}
s_k\left(\mat{U}_{1:k}^{\top}\,b((\mat{M}+\mathcal{E})\mat{W}) \right)^2 = \min_{\vect{f}\in \mathbb{S}^{k-1}} \frac{q(\vect{f})^2}{1+q(\vect{f})^2}\,,\qquad q(\vect{f})=\frac{\|\mat{U}_{1:k}^{\top} (\mat{M}+\mathcal{E})\mat{W}\vect{f}\|}{\|\mat{U}_{k+1:p}^{\top} (\mat{M}+\mathcal{E})\mat{W}\vect{f}\|}\,.
\end{equation*}
Since $x^2/(1+x^2)$ increases in $x\ge0$, the minimum is at the smallest $q(\vect{f})$, and dividing the two bounds in~\eqref{eq:sigamp-fc} gives
\begin{equation*}
q(\vect{f})\ge\frac{s_k(\mat{M})\sqrt{1-(\epsilon+\eta)^2}-\totalbound}{s_{k+1}(\mat{M})(\epsilon+\eta)+\totalbound}=\frac{s_k(\mat{M}) -\totalbound/\sqrt{1-(\epsilon+\eta)^2}}{s_{k+1}(\mat{M})+\totalbound/(\epsilon+\eta)}\frac{\sqrt{1-(\epsilon+\eta)^2}}{\epsilon+\eta} >\frac{\sqrt{1-\epsilon^2}}{\epsilon}\,,
\end{equation*}
the last step being Lemma~\ref{lem:thm2ratio}\,\ref{ri:ratio3}.
As $q(\vect{f})>\sqrt{1-\epsilon^2}/\epsilon$ gives $q(\vect{f})^2/(1+q(\vect{f})^2)\ge 1-\epsilon^2$, we conclude $s_k(\mat{U}_{1:k}^{\top}\,b((\mat{M}+\mathcal{E})\mat{W}))^2\ge1-\epsilon^2$ and $d(\mat{U}_{1:k},(\mat{M}+\mathcal{E})\mat{W})\le\epsilon$.
\end{proof}

\begin{proof}[Proof of Theorem~\ref{thm:drifting-power-method}]
We split the proof into three steps.
Step~1 builds a backward sequence of complement frames $\mat{N}^{(\ell)}$ that trap the trailing $(p-k)$-dimensional subspace, using the orthogonal-amplification Proposition~\ref{lem:orth_nonstat}.
Step~2 builds a forward Gram-Schmidt sequence of signal frames $\mat{W}^{(\ell)}$ that track the leading $k$-dimensional subspace, using the signal-amplification Proposition~\ref{lem:proj_nonstat}.
Step~3 squeezes the random iterate $\mat{\widehat{U}}_{1:k}(L)$ onto the signal frame $\mat{W}^{(L+1)}$, which spans the same space as $\mathcal{M}^{(L)}\mat{W}^{(1)}$, and bounds its distance to the true subspace $\mat{U}_{1:k}(L)$ by a triangle inequality and the $\beta^{-L}$ contraction.

The two sequences run in opposite directions because the recovery target is at the last block: the output must align with $\mat{U}_{1:k}(L)$ and avoid $\mat{U}_{k+1:p}(L)$.
The signal frames are therefore grown forward from an arbitrary $\mat{W}^{(1)}$, each pushed through the same map as the iterate, so they track the leading subspace and lower-bound the signal the iterate accumulates.
The complement frames are pulled backward from $\mat{N}^{(L+1)}=\mat{U}_{k+1:p}(L)$, each $\mat{N}^{(\ell)}$ chosen to map into the next, so that $\mathcal{M}^{(L)}\mat{N}^{(1)}$ stays trapped in the final trailing subspace.
The asymmetry is temporal: $\mat{U}_{1:k}(\ell)$ is tracked at every step, so the forward iteration's local analysis suffices; $\mat{U}_{k+1:p}(L)$ is pinned only at the terminal block, so the backward chain anchors at $\ell=L+1$ as the only available seed.

\textbf{Step 1: constructing the complement frames $\mat{N}^{(\ell)}$}

We construct the sequence $\{\mat{N}^{(\ell)} \}_{1 \le \ell \le L+1}, \mat{N}^{(\ell)} \in \stiefel{p-k}{p}$ so that the following is satisfied: 
\begin{enumerate}[label=\textbf{N.\arabic*}]
\item \label{c1} $\displaystyle \mat{N}^{(L+1)} = \mat{U}_{k+1:p}(L)\,.$
\item \label{c2} $\displaystyle \mbox{ran}((\mat{M}(\ell)+\mathcal{\mat{\mathcal{E}}}(\ell))\mat{N}^{(\ell)}) \subseteq
  \mbox{ran}(\mat{N}^{(\ell+1)})\,,\quad\forall \ell\in[L]\,.$
\item \label{c3} $\displaystyle \|\mat{U}_{1:k}(\ell)^\top \mat{N}^{(\ell)}\| \le \epsilon \:\:\mbox{and}\:\: \| (\mat{M}(\ell)+\mat{\mathcal{E}}(\ell)) \mat{N}^{(\ell)} \| \le \frac{s_{k+1}(\mat{M}(\ell))+\totalbound}{\sqrt{1-(\epsilon+\eta)^2}}\,,\quad\forall \ell\in[L]\,.$
\end{enumerate}
We build $\{\mat{N}^{(\ell)}\}_{\ell=1}^{L+1}$ satisfying \ref{c1}-\ref{c3} by backward induction with Proposition~\ref{lem:orth_nonstat}.
Applying the proposition at block $\ell$ requires its target to be a near-complement of the signal, $\|\mat{U}_{1:k}(\ell)^\top \mat{N}^{(\ell+1)}\| \le \epsilon+\eta$; the induction carries this admissibility from each block to the previous one.

Base case: at $\ell = L+1$, set $\mat{N}^{(L+1)}=\mat{U}_{k+1:p}(L)$.
It is admissible for the first backward step because $\|\mat{U}_{1:k}(L)^\top \mat{N}^{(L+1)}\|=\|\mat{U}_{1:k}(L)^\top \mat{U}_{k+1:p}(L)\|=0$, the leading and trailing blocks of $\mat{M}(L)$ being orthogonal.
Conditions \ref{c2} and \ref{c3} are required only for $\ell\in[L]$, so they impose nothing at $\ell=L+1$.

Inductive step: suppose $\mat{N}^{(\ell+1)}$ is admissible, $\|\mat{U}_{1:k}(\ell)^\top \mat{N}^{(\ell+1)}\|\le\epsilon+\eta$.
Then $\mat{N}^{(\ell+1)}$ lies in the admissible set of Proposition~\ref{lem:orth_nonstat} for $\mat{M}=\mat{M}(\ell)$ and $\mathcal{E}=\mathcal{E}(\ell)$, so taking $\mat{Y}=\mat{N}^{(\ell+1)}$ yields $\mat{N}^{(\ell)}$ satisfying \ref{c2} and \ref{c3}.
For $\ell\ge2$ we propagate the admissibility to block $\ell-1$: since
\begin{align*}
\|\mat{U}_{1:k}(\ell-1)^{\top}\mat{N}^{(\ell)}\| &\stackrel{(i)}{=} \Vert \mat{U}_{1:k}(\ell-1)\mat{U}^{\top}_{1:k}(\ell-1)-(\mat{I}_{p \times p}-\mat{N}^{(\ell)}(\mat{N}^{(\ell)})^{\top}) \Vert \\ 
&\stackrel{(ii)}{\leq} \Vert \mat{U}_{1:k}(\ell-1)\mat{U}^{\top}_{1:k}(\ell-1) - \mat{U}_{1:k}(\ell)\mat{U}^{\top}_{1:k}(\ell)\Vert+\Vert \mat{U}_{1:k}(\ell)\mat{U}^{\top}_{1:k}(\ell)-(\mat{I}_{p \times p}-\mat{N}^{(\ell)}(\mat{N}^{(\ell)})^{\top}) \Vert \\
&\stackrel{(iii)}{\le} \eta+\epsilon
\end{align*}
where $(i)$ expresses the overlap as the operator-norm gap between the projectors onto $\mathrm{ran}(\mat{U}_{1:k}(\ell-1))$ and $\mathrm{ran}(\mat{N}^{(\ell)})^{\perp}$ via Lemma~\ref{lem:proj_nonstat.6.1_golub}, and $(ii)$ is the triangle inequality that inserts the projector onto $\mathrm{ran}(\mat{U}_{1:k}(\ell))$.
Step $(iii)$ then bounds the first term by $\eta$ through the drift hypothesis $\|(\mat{U}_{k+1:p}(\ell))^\top \mat{U}_{1:k}(\ell-1)\| \le \eta$, and the second by $\epsilon$ through Lemma~\ref{lem:proj_nonstat.6.1_golub} again with $\|\mat{U}_{1:k}(\ell)^\top\mat{N}^{(\ell)}\|\le\epsilon$ from~\ref{c3}.
Hence $\|\mat{U}_{1:k}(\ell-1)^\top\mat{N}^{(\ell)}\|\le\epsilon+\eta$, so $\mat{N}^{(\ell)}$ is admissible and the induction continues, producing $\{\mat{N}^{(\ell)}\}_{1\le \ell \le L+1}$ with properties \ref{c1}-\ref{c3}.

From the properties \ref{c1}-\ref{c3} of $ \{\mat{N}^{(\ell)} \}_{1\le \ell \le L+1}$, we have:
\begin{equation}
\label{eq:sk1}
\mbox{ran}(\mathcal{M}^{(L)} \mat{N}^{(1)}) \subseteq \mbox{ran}(\mat{U}_{k+1:p}(L))\:\:\mbox{and}\:\:
\|\mathcal{M}^{(L)} \mat{N}^{(1)}\| \le \prod_{\ell=1}^L\left( \frac{s_{k+1}(\mat{M}(\ell))+\totalbound}{\sqrt{1-(\epsilon+\eta)^2}} \right).
\end{equation}
The range inclusion chains \ref{c2} block by block and terminates at $\mbox{ran}(\mat{U}_{k+1:p}(L))$ by \ref{c1}.
For the norm bound, \ref{c2} lets each block factor as $(\mat{M}(\ell)+\mathcal{E}(\ell))\mat{N}^{(\ell)}=\mat{N}^{(\ell+1)}\mat{C}^{(\ell)}$ with $\|\mat{C}^{(\ell)}\|=\|(\mat{M}(\ell)+\mathcal{E}(\ell))\mat{N}^{(\ell)}\|$, as $\mat{N}^{(\ell+1)}$ is orthonormal.
Chaining these gives $\mathcal{M}^{(L)}\mat{N}^{(1)}=\mat{N}^{(L+1)}\mat{C}^{(L)}\cdots\mat{C}^{(1)}$, so $\|\mathcal{M}^{(L)}\mat{N}^{(1)}\|\le\prod_{\ell=1}^{L}\|\mat{C}^{(\ell)}\|$ with each factor bounded by~\ref{c3}.

\textbf{Step 2: constructing the signal frames $\mat{W}^{(\ell)}$}

Next, we define the sequence  $\{\mat{W}^{(\ell)}\}_{\ell=1}^{L+1}$, $\mat{W}^{(\ell)}\in \stiefel{k}{p}$ as follows: 
\begin{enumerate}[label=\textbf{W.\arabic*}]
\item $\displaystyle \mat{W}^{(1)}\in \stiefel{k}{p}$ chosen so that $(\mat{W}^{(1)})^\top \mat{N}^{(1)} = \mat{0}$, i.e., $\mathrm{ran}(\mat{W}^{(1)})$ and $\mathrm{ran}(\mat{N}^{(1)})$ are orthogonal.
\item $\displaystyle \mat{W}^{(\ell +1)} = \text{Gram-Schmidt}((\mat{M}(\ell)+\mat{\mathcal{E}}(\ell))\mat{W}^{(\ell)})\,,\quad\forall \ell\in[L]\,.$
\end{enumerate}

From \ref{c3} at $\ell=1$ and \textbf{W.1}, we have $d(\mat{U}_{1:k}(1),\mat{W}^{(1)}) \le \epsilon +
\eta$. Proposition~\ref{lem:proj_nonstat} then applies at each block: for every $\ell\in[L]$,
\begin{equation*}
s_k\left( (\mat{M}(\ell)+\mathcal{E}(\ell))\mat{W}^{(\ell)} \right) \ge \sqrt{1-(\epsilon+\eta)^2}\, s_{k}(\mat{M}(\ell)) -\totalbound\,,\qquad d(\mat{U}_{1:k}(\ell),\mat{W}^{(\ell+1)}) \le \epsilon\,,
\end{equation*}
with the hypothesis $d(\mat{U}_{1:k}(\ell),\mat{W}^{(\ell)}) \le \epsilon + \eta$ holding at $\ell=1$ by \textbf{W.1} and~\ref{c3}, and at $2\le\ell\le L$ because the drift $\Vert \mat{U}_{k+1:p}(\ell)^{\top}\mat{U}_{1:k}(\ell-1) \Vert \le \eta$ and the previous block's bound $d(\mat{U}_{1:k}(\ell-1),\mat{W}^{(\ell)}) \le \epsilon$ give $d(\mat{U}_{1:k}(\ell),\mat{W}^{(\ell)}) \le \eta+\epsilon$.
Taking the second bound at $\ell=L$ and multiplying the first over $\ell=1,\dots,L$,
\begin{equation}
\label{eq:sk2}
d(\mat{W}^{(L+1)}, \mat{U}_{1:k}(L))\le \epsilon \:\:\mbox{and}\:\:
s_k(\mathcal{M}^{(L)} \mat{W}^{(1)}) \ge \prod_{\ell=1}^L\left( \sqrt{1-(\epsilon+\eta)^2}\, s_{k}(\mat{M}(\ell)) -\totalbound \right)\,.
\end{equation}
This establishes the two frame sequences $\{\mat{N}^{(\ell)},\mat{W}^{(\ell)}\}_{\ell=1}^{L+1}$, summarized by the block products~\eqref{eq:sk1} and~\eqref{eq:sk2}: the complement frames cap the trailing growth of $\mathcal{M}^{(L)}$, while the signal frames track $\mat{U}_{1:k}$.

\textbf{Step 3: distance between the true and recovered subspaces}

Now, we upper bound the distance between the output of the power method $\mat{\widehat{U}}_{1:k}(L)$ and the first $k$ singular vectors of the true underlying subspace $\mat{U}_{1:k}(L)$, $d (\mat{U}_{1:k}(L) ,\mat{\widehat{U}}_{1:k}(L) )$.
From the triangle inequality we have,
\begin{equation}
d(\mat{U}_{1:k}(L),\mat{\widehat{U}}_{1:k}(L) ) \le d(\mat{U}_{1:k}(L) ,\mat{W}^{(L+1)} ) + d (\mathcal{M}^{(L)}\mat{W}^{(1)} ,\mat{\widehat{U}}_{1:k}(L) ).\label{eq:dD}
\end{equation}
(Here $\mat{W}^{(L+1)}$ and $\mathcal{M}^{(L)}\mat{W}^{(1)}$ span the same column space.) From~\eqref{eq:sk2}, we have
\begin{equation}
d(\mat{U}_{1:k}(L) ,\mathcal{M}^{(L)} \mat{W}^{(1)} ) \le \epsilon\,.
\label{eq:5}
\end{equation}
To bound the second term in~\eqref{eq:dD}, recall that the power iteration makes $\mat{\widehat{U}}_{1:k}(L)=b(\mathcal{M}^{(L)}\mat{\widehat{U}}_{1:k}(0))$, and that $\mat{W}^{(1)},\mat{N}^{(1)}$ have orthogonal ranges (\textbf{W.1}) of dimensions $k$ and $p-k$, so $\mat{W}^{(1)}(\mat{W}^{(1)})^\top+\mat{N}^{(1)}(\mat{N}^{(1)})^\top=\mat{I}$.
Writing $\mat{P}=(\mathcal{M}^{(L)}\mat{W}^{(1)})_{\perp}$ for the unit-norm projection onto the complement of $\mathrm{ran}(\mathcal{M}^{(L)}\mat{W}^{(1)})$ (i.e., $\mat{P}^\top\mathcal{M}^{(L)}\mat{W}^{(1)}=\mat{0}$),
\begin{align*}
d(\mathcal{M}^{(L)}\mat{W}^{(1)} ,\mat{\widehat{U}}_{1:k}(L) )
\stackrel{(i)}{=}&\ \big\|\mat{P}^\top \mathcal{M}^{(L)}\mat{\widehat{U}}_{1:k}(0)\,\big((\mathcal{M}^{(L)}\mat{\widehat{U}}_{1:k}(0))^\top \mathcal{M}^{(L)}\mat{\widehat{U}}_{1:k}(0)\big)^{-1/2}\big\|\\
\stackrel{(ii)}{\le}&\ \frac{\big\|\mat{P}^\top \mathcal{M}^{(L)}\mat{\widehat{U}}_{1:k}(0)\big\|}{s_k(\mathcal{M}^{(L)}\mat{\widehat{U}}_{1:k}(0))}
\stackrel{(iii)}{=} \frac{\big\|\mat{P}^\top \mathcal{M}^{(L)}\mat{N}^{(1)}(\mat{N}^{(1)})^\top\mat{\widehat{U}}_{1:k}(0)\big\|}{s_k(\mathcal{M}^{(L)}\mat{\widehat{U}}_{1:k}(0))} \\
\stackrel{(iv)}{\le}& \frac{\|\mathcal{M}^{(L)}\mat{N}^{(1)}\|\,\|(\mat{N}^{(1)})^\top\mat{\widehat{U}}_{1:k}(0)\|}{s_k(\mathcal{M}^{(L)}\mat{\widehat{U}}_{1:k}(0))}.
\end{align*}
Step $(i)$ substitutes $\mat{\widehat{U}}_{1:k}(L)=b(\mathcal{M}^{(L)}\mat{\widehat{U}}_{1:k}(0))$ into the distance.
Step $(ii)$ is Cauchy--Schwarz with $\|((\mathcal{M}^{(L)}\mat{\widehat{U}}_{1:k}(0))^\top\mathcal{M}^{(L)}\mat{\widehat{U}}_{1:k}(0))^{-1/2}\|=1/s_k(\mathcal{M}^{(L)}\mat{\widehat{U}}_{1:k}(0))$.
Step $(iii)$ inserts the decomposition of $\mat{\widehat{U}}_{1:k}(0)$ above and drops the signal term, since $\mat{P}^\top\mathcal{M}^{(L)}\mat{W}^{(1)}=\mat{0}$; step $(iv)$ is Cauchy--Schwarz with $\|\mat{P}\|=1$.

Denote the noise-component product by $n=\|\mathcal{M}^{(L)}\mat{N}^{(1)}\|\,\|(\mat{N}^{(1)})^\top\mat{\widehat{U}}_{1:k}(0)\|$ and the signal-component product by $g=s_k(\mathcal{M}^{(L)}\mat{W}^{(1)})\,s_k((\mat{W}^{(1)})^\top\mat{\widehat{U}}_{1:k}(0))$.
Define the initialization-imbalance ratio $R:=\|(\mat{N}^{(1)})^\top\mat{\widehat{U}}_{1:k}(0)\|/s_k((\mat{W}^{(1)})^\top\mat{\widehat{U}}_{1:k}(0))$.
By~\eqref{eq:sk1} and~\eqref{eq:sk2}, $n/g=\big(\|\mathcal{M}^{(L)}\mat{N}^{(1)}\|/s_k(\mathcal{M}^{(L)}\mat{W}^{(1)})\big)R\le\beta^{-L}R$.
When $\beta^{-L}R<1$ we have $n/g<1$ and hence $g>n$; the decomposition $\mat{\widehat{U}}_{1:k}(0)=(\mat{W}^{(1)}(\mat{W}^{(1)})^\top+\mat{N}^{(1)}(\mat{N}^{(1)})^\top)\mat{\widehat{U}}_{1:k}(0)$ then gives $s_k(\mathcal{M}^{(L)}\mat{\widehat{U}}_{1:k}(0))\ge g-n$, so
\begin{equation}
\label{eq:tildehat}
d(\mathcal{M}^{(L)}\mat{W}^{(1)} ,\mat{\widehat{U}}_{1:k}(L) )\le \frac{n}{g-n}=\frac{n/g}{1-n/g}\le \frac{\beta^{-L}R}{1-\beta^{-L}R}\,.
\end{equation}
As a subspace distance, $d(\mathcal{M}^{(L)}\mat{W}^{(1)} ,\mat{\widehat{U}}_{1:k}(L) )\le 1$ always; and when $\beta^{-L}R\le\tfrac12$ the right-hand side of~\eqref{eq:tildehat} is at most $2\beta^{-L}R$.
Both regimes give $d(\mathcal{M}^{(L)}\mat{W}^{(1)} ,\mat{\widehat{U}}_{1:k}(L) )\le\min\{1,2\beta^{-L}R\}$, so with~\eqref{eq:5} and the triangle inequality~\eqref{eq:dD},
\begin{equation*}
    d(\mat{U}_{1:k}(L) ,\mat{\widehat{U}}_{1:k}(L) ) \le \epsilon + \min\left\{1, 2 \beta^{-L}R \right\}.
\end{equation*}
Finally, since $\mat{\widehat{U}}_{1:k}(0)$ is a uniformly random $p\times k$ orthonormal matrix, the random-initialization analysis of~\citep{npm} gives $R\le\tau_0\sqrt{pk}$ with probability $1-O(\tau_0^{-1})-e^{-\Omega(p)}$ for any $\tau_0\ge 1$ ($\tau_0$ is the free scale parameter, distinct from the trajectory length $\tau = \lceil 3s\delta/\Gamma\rceil$ used in Appendix~\ref{appendix:lower-bound}).
Since $\min\{1,2\beta^{-L}R\}\le 2\beta^{-L}R$, this gives
\begin{equation*}
d(\mat{U}_{1:k}(L),\mat{\widehat{U}}_{1:k}(L)) \le \epsilon+O(\beta^{-L}\tau_0\sqrt{pk})\,,
\end{equation*}
on the event $\{\lVert\mathcal{E}(\ell)\rVert\le\totalbound\ \text{for all }\ell\}$, with the stated probability.
\end{proof}

\section{Proof of Theorem~\ref{thm:robust-power-method}: noisy power method convergence under drift}
\label{appendix:robust-power-method}
We prove Theorem~\ref{thm:robust-power-method} under the condition in~\eqref{eq:npmlemmabound}, which holds with probability at least $1-1/T$.

\subsection{Deriving the optimal block size}
\label{derive_npm_optimal_B}
Consider the upper bound (for probability greater than $1-1/T$) on $\max_{\ell}\lVert\mathcal{E}(\ell)\rVert$ from Appendix~\ref{appendix:power-error-bound}:
\[
\max_{\ell}\lVert\mathcal{E}(\ell)\rVert
  \leq C_{\mathrm{NPM}}\sqrt{\frac{\log(2pT^2)}{B}}
    + \frac{B\Gamma}{2}\,,
\quad\text{where}\quad
C_{\mathrm{NPM}} = C_{\mathrm{B}}\sqrt{\mathcal{V}} \le 3.2\sqrt{\mathcal{V}}\quad\text{(Remark~\ref{rmk:absolute-constants})}.
\]
Differentiating in $B$ and solving for the critical point:
\[
B_{\mathrm{opt}} = \frac{C_{\mathrm{NPM}}^{2/3}\log(2pT^2)^{1/3}} {\Gamma^{2/3}}\,.
\]
The uniform upper bound for the error matrix becomes:
\[
\max_{\ell}\lVert\mathcal{E}(\ell)\rVert
  \leq \tfrac{3}{2}\,
    C_{\mathrm{NPM}}^{2/3}\log(2pT^2)^{1/3}\Gamma^{1/3}.
\]

\subsection{Instantiating the parameters}
Set $B = B_{\mathrm{opt}}$ and consider the regime:
\begin{enumerate}[label=(\Alph*)]
    \item $36 B_{\mathrm{opt}}\Gamma  =36\,{C}_{\mathrm{NPM}}^{2/3} \log(2pT^2)^{1/3} \Gamma^{1/3} \leq \delta$, from $\Gamma =  O\left(\frac{\delta^3}{{C}_{\mathrm{NPM}}^{2}\log(2pT^2)}\right)\,.$
    \item $\delta \geq \frac{12}{17}\sigma^2\,.$
\end{enumerate}

We instantiate the parameters of Appendix~\ref{appendix:amplification} as:
\begin{enumerate}[label=(\alph*)]
    \item $\displaystyle\totalbound := \frac{3}{2} B_{\mathrm{opt}}\Gamma = \frac{3}{2} {C}_{\mathrm{NPM}}^{2/3} \log(2pT^2)^{1/3} \Gamma^{1/3}$
    
    $\,(\geq \max_{\ell}{\lVert \mathcal{E}(\ell) \rVert} \text{ on probability greater than }1-1/T, \text{ derived in Subsection~\ref{derive_npm_optimal_B}})\,.$
    \item $\displaystyle\epsilon := \frac{4\totalbound}{\delta} = \frac{6 B_{\mathrm{opt}}\Gamma}{\delta} \leq \frac{1}{6}$.
    \item $\displaystyle \eta := \frac{B_{\mathrm{opt}}\Gamma}{\delta 
    -B_{\mathrm{opt}}\Gamma}\,.$
\end{enumerate}
From items (A) and (c) above, we have:
\[
\frac{\eta}{\epsilon} = \frac{B_{\mathrm{opt}}\Gamma/(\delta - B_{\mathrm{opt}}\Gamma)}{4\totalbound/\delta} = \frac{B_{\mathrm{opt}}\Gamma/(\delta - B_{\mathrm{opt}}\Gamma)}{6B_{\mathrm{opt}}\Gamma/\delta} = \frac{1}{6} \frac{\delta}{\delta - B_{\mathrm{opt}}\Gamma}\leq \frac{6}{35} \leq \frac{1}{5}\,.
\]
The definition $\epsilon=4\totalbound/\delta\le\tfrac16$ gives $\totalbound\le\delta/16$ and $\epsilon\le\tfrac14$.
With the bound $\eta/\epsilon\le 1/5$ just shown and regime~(B), the per-block matrices $\mat{M}(\ell)$ and errors $\mathcal{E}(\ell)$ satisfy the standing assumptions~\eqref{eq:amp-standing} for each block $\ell$ (i.e., blockwise).
The consecutive leading subspaces obey the drift bound~\eqref{eq:amp-drift}, $\lVert\mat{U}_{k+1:p}(\ell)^\top\mat{U}_{1:k}(\ell-1)\rVert\le\eta$, by the Davis--Kahan $\sin\Theta$ theorem (Lemma~\ref{lem:dk_sin}) applied to the per-block drift $B\Gamma$ and the spectral gap $\delta$, with $\eta=B_{\mathrm{opt}}\Gamma/(\delta-B_{\mathrm{opt}}\Gamma)$.

Theorem~\ref{thm:drifting-power-method} therefore applies.
On the event $\{\lVert\mathcal{E}(\ell)\rVert\le\totalbound\ \text{for all }\ell\}$, the $L$-step noisy power method satisfies
\begin{equation*}
d(\mat{U}_{1:k}(L),\widehat{\mat{U}}_{1:k}(L)) \le \epsilon+O(\beta^{-L}\tau_0\sqrt{pk})
\end{equation*}
with probability at least $1-O(\tau_0^{-1})-e^{-\Omega(p)}$ over the initialization.
Setting $\tau_0=T$ collapses the initialization failure into the $1/T$ scale, and by Lemma~\ref{lem:power-error}, the event $\{\lVert\mathcal{E}(\ell)\rVert\le\totalbound\}$ holds with probability at least $1-1/T$; hence the overall bound holds with probability greater than $1-2/T-e^{-\Omega(p)}$.

To verify that the exponential tail dominates in the regime $T \ge T_c$: the prefactor $\tau_0\sqrt{pk}=T\sqrt{pk}$ is absorbed into $\beta^{-L}$ whenever $L\ge\log(T\sqrt{pk})/\log\beta$, a $\tilde O(\log(pT))$ threshold on the block count $L=T/B_{\mathrm{opt}}$.
Since $B_{\mathrm{opt}}$ depends on $T$ only through a $\log^{1/3}$ factor, $L$ grows almost linearly in $T$, whereas the threshold grows only logarithmically.
Thus $L$ clears the threshold once $T$ exceeds $T_c$ by the polylogarithmic factor implied by $T=\Omega(T_c)$.
At $T=T_c$ itself $B_{\mathrm{opt}}=\tilde\Theta(T_c)$ (the critical time is one optimal block up to a $\log^{1/3}$ factor), so $L=T/B_{\mathrm{opt}}=\tilde\Theta(1)$ and the margin comes from the $\Omega(\cdot)$ rather than from $T_c$ alone.
Lemma~\ref{lem:thm2ratio}(i) gives $\beta\ge 1.4464 > 1/0.7$, so $\beta^{-L} \le 0.7^L = 0.7^{T/B}$.

Finally, substituting $\epsilon=4\totalbound/\delta$,
\begin{equation*}
\epsilon = \frac{4 \totalbound}{\delta} \propto \frac{{C}_{\mathrm{NPM}}^{2/3} \log(2pT^2)^{1/3} \Gamma^{1/3}}{\delta}  = \tilde{\Theta}\Big( \Gamma^{1/3} \cdot \frac{(\mu\delta + \sigma^2)^{1/3} (p\sigma^2 + k\mu\delta)^{1/3}}{\delta}\Big) \,,
\end{equation*}
gives the statement of Theorem~\ref{thm:robust-power-method}.

\paragraph{Tail-to-expectation conversion.}
\label{appendix:npm-tail-to-expectation}
The high-probability conclusion above, together with $d\le 1$ on the $\tilde{O}(1/T)$-scale complementary event, yields $\mathbb{E}[d]\le\epsilon+0.7^{T/B}+\tilde{O}(1/T)$.
In the regime $T\ge T_c$, the geometric tail $0.7^{T/B}$ falls below $\epsilon$, and $\tilde{O}(1/T)\le\epsilon$ holds iff $T^{3}\Gamma\ge\delta^{3}/(p\sigma^{2}(\sigma^{2}+\delta))$.
At $T=T_c$ this reads $(p\sigma^{2}(\sigma^{2}+\delta))^{2}/\Gamma\ge\delta^{3}$ via~\eqref{eq:critical-time}, automatic under the $\Gamma$-regime of Theorem~\ref{thm:robust-power-method} (noise-energy-dominated, $p\sigma^{2}\gtrsim k\delta$): $\Gamma=\tilde{O}(\delta^{3}/(p\sigma^{2}(\sigma^{2}+\delta)))$.
Hence $\mathbb{E}[d]\le\mathcal{R}^{\mathrm{NPM}}+\tilde{O}(1/T)$ with $\mathcal{R}^{\mathrm{NPM}}=\tilde{O}\bigl(\Gamma^{1/3}(p\sigma^{2}(\sigma^{2}+\delta)/\delta^{2})^{1/3}\bigr)$.
\endproof  

\section{Proof of Lemma~\ref{lem:oja-error}: Oja error bound}
\label{appendix:oja-error-bound}

\subsection{Truncation event and almost-sure boundedness}
\label{appendix:tail-bounds}

The matrix product concentration of~\citep[Corollary~6.1]{huang2020matrix} demands an almost-sure upper bound on the spectral norm of the centered observation, which the sub-Gaussian structure alone does not supply.
We obtain this by conditioning on a high-probability truncation event that bounds the coordinate-wise magnitudes of $\vect{z}_t$.

The almost-sure corollary is the looser of the two forms available.
The master statement of the same inequality~\citep{huang2020matrix} is governed by an accumulated variance $\sum_i\sigma_i^2$, parallel to the matrix variance $\mathcal{V}$ of the noisy power method, and the sharp finite-sample analysis of Oja's algorithm attains the matrix Bernstein rate through this variance route~\citep{jain2016streaming, Allen2017, HuangNW21}.
That line of work is inseparable from the stationary i.i.d.\ model: the Schatten-norm recursion of~\citet{HuangNW21}, for instance, conditions on a good event $\{\lVert\mat{W}_t\rVert\le\gamma\}$ that drives the iterate to the fixed top-$k$ eigenspace of a single covariance at a fixed spectral gap, a contraction the drifting target removes (Appendix~\ref{appendix:amplification}).
Lacking such a fixed point, we take the almost-sure corollary; the ambient $p$ inside $\mathcal{M}_\infty$ is the source of the $p^{1/3}$ slack in Theorem~\ref{thm:ojas-algorithm}.

\paragraph{Truncation event.}
Recall under model~\eqref{eqn:spike} that $\vect{x}_t = \Sigma_t^{1/2}\vect{z}_t$ where $\vect{z}_t \in \mathbb{R}^p$ has independent, mean-zero, unit-variance entries satisfying $\lVert z_{t,i}\rVert_{\psi_2} \leq \kappa$.
Define
\begin{equation}
\label{eq:truncevent}
\mathfrak{E}
  := \bigl\{\,
    |z_{t,i}| \leq \rho
    \;\;\forall\, t \in [T],\; i \in [p]
  \,\bigr\},
\qquad
\rho := \kappa\sqrt{2\log(2pT^2)}.
\end{equation}

\begin{lemma}[Truncation probability]
\label{lem:trunc-prob}
$\mathbb{P}[\mathfrak{E}] \geq 1 - 1/T$.
\end{lemma}
\begin{proof}
By the sub-Gaussian tail bound (Definition~2.6.4 of~\citet{Vershynin2018}),
\[
\mathbb{P}(|z_{t,i}| \geq \rho)
  \leq 2\exp\!\bigl(-\rho^2/(2\kappa^2)\bigr)
  = 2\exp\!\bigl(-\log(2pT^2)\bigr) = \frac{1}{pT^2}
\]
A union bound over $p$ coordinates and $T$ time steps gives $\mathbb{P}[\mathfrak{E}^c]\leq pT/(pT^2) = 1/T$.
\end{proof}

\paragraph{Conditional bias.}
Since the entries $z_{t,i}$ are independent across both $t$ and $i$, conditioning on $\mathfrak{E}$ retains a product structure:
\[
\mathcal{L}(\vect{z}_t \mid \mathfrak{E})
  = \bigotimes_{i=1}^{p}
    \mathcal{L}(z_{t,i} \mid |z_{t,i}| \leq \rho).
\]
Each truncated entry has second moment $\mathbb{E}[z_{t,i}^2 \mid |z_{t,i}| \leq \rho] = 1 - \nu$, where
\begin{equation}
\label{eq:nu-def}
\begin{aligned}
\nu
  := 1 - \mathbb{E}[z_{t,i}^2 \mid |z_{t,i}| \leq \rho]
  &= \frac{\mathbb{P}(|z_{t,i}| \leq \rho) - \mathbb{E}[z_{t,i}^2 \mathbf{1}_{|z_{t,i}| \leq \rho}]}
    {\mathbb{P}(|z_{t,i}| \leq \rho)} \\
  &= \frac{\mathbb{E}[z_{t,i}^2 \mathbf{1}_{|z_{t,i}| > \rho}] - \mathbb{P}(|z_{t,i}| > \rho)}
    {\mathbb{P}(|z_{t,i}| \leq \rho)}
  \leq \frac{\mathbb{E}[z_{t,i}^2 \mathbf{1}_{|z_{t,i}| > \rho}]}
    {\mathbb{P}(|z_{t,i}| \leq \rho)}.
\end{aligned}
\end{equation}
The second equality splits the unit second moment $\mathbb{E}[z_{t,i}^2]=1$ across $\{|z_{t,i}|\le\rho\}$ and $\{|z_{t,i}|>\rho\}$ and uses $\mathbb{P}(|z_{t,i}|\le\rho)=1-\mathbb{P}(|z_{t,i}|>\rho)$; the inequality drops the non-negative $\mathbb{P}(|z_{t,i}|>\rho)$ term.
We bound the numerator using the layer-cake identity $\mathbb{E}[Y]=\int_0^\infty\mathbb{P}(Y>y)\,dy$ for $Y = z_{t,i}^2 \mathbf{1}_{|z_{t,i}|>\rho}$.
The event $\{Y>y\}=\{|z_{t,i}|>\max(\rho,\sqrt{y})\}$ changes form at $y=\rho^2$, so we split the integration range there.
The substitution $s=\sqrt{y}$ on the upper piece then gives
\[
\mathbb{E}[Y] = \rho^2 \mathbb{P}(|z_{t,i}|>\rho) + 2\int_\rho^\infty s\,\mathbb{P}(|z_{t,i}|>s)\,ds.
\]
Bounding both terms with the sub-Gaussian tail $\mathbb{P}(|z_{t,i}|>s)\leq 2\exp(-s^2/(2\kappa^2))$ and the threshold $\exp(-\rho^2/(2\kappa^2))=1/(2pT^2)$ (so $\rho^2=2\kappa^2\log(2pT^2)$),
\[
\mathbb{E}\bigl[z_{t,i}^2 \cdot
  \mathbf{1}_{|z_{t,i}| > \rho}\bigr]
\leq \frac{2\kappa^2 \log(2pT^2)}{pT^2}
  + \frac{2\kappa^2}{pT^2}
\leq \frac{4\kappa^2 \log(2pT^2)}{pT^2}\,,
\]
where the last inequality uses $\log(2pT^2) \geq 1$ for $pT^2 \geq e/2$.
Since $\mathbb{P}(|z_{t,i}| \leq \rho) \geq 1 - 1/(pT^2) \geq 1/2$, we obtain
\begin{equation}
\label{eq:nu-bound}
\nu \leq \frac{8\kappa^2 \log(2pT^2)}{pT^2}.
\end{equation}
This gives $\nu = \tilde{O}(1/(pT^2))$.
The conditional covariance is therefore
\begin{equation}
\label{eq:cond-cov}
\mathbb{E}[\xxT{\vect{x}}{t} \mid \mathfrak{E}]
  = (1-\nu)\,\Sigma_t + \Delta_t,
\qquad
\lVert\Delta_t\rVert
  = \tilde{O}(1/(pT^2)),
\end{equation}
where $\Delta_t = \Sigma_t^{1/2} R_t \Sigma_t^{1/2}$ and $R_t$ is the off-diagonal part of $\mathbb{E}[\xxT{\vect{z}}{t}\mid\mathfrak{E}]$, with entries $R_t[i,j] = \mathbb{E}[z_{t,i}\mid\mathfrak{E}]\,\mathbb{E}[z_{t,j}\mid\mathfrak{E}]$ for $i\ne j$ and zero on the diagonal.

\begin{remark}[Modified parameters under $\mathfrak{E}$]
\label{rmk:modified-params}
Throughout the Oja analysis we treat $(1-\nu)\Sigma_t+\Delta_t$ as $\Sigma_t$.
Since $\nu,\lVert\Delta_t\rVert=\tilde{O}(1/(pT^2))$, Weyl's inequality (Lemma~\ref{weyl}) gives $|s_i((1-\nu)\Sigma_t+\Delta_t)-s_i(\Sigma_t)|\le\nu\lVert\Sigma_t\rVert+\lVert\Delta_t\rVert=\tilde{O}(1/(pT^2))$ for every $i$.
Each spectral parameter therefore shifts only by additive $\tilde{O}(1/(pT^2))$ terms: $\delta\to\delta-\tilde{O}(1/(pT^2))$, and similarly for $\sigma^2$, while $\mu$ and $\Gamma$ inherit the same order via the ratio/difference.
These shifts are absorbed into the absolute constants of the final bounds.
\end{remark}

\paragraph{Almost-sure spectral norm $\mathcal{M}_\infty$.}
Under $\mathfrak{E}$ the centered observation admits a deterministic spectral-norm bound.
For the raw observation, each $|z_{t,i}|\le\rho$ gives $\lVert\vect{z}_t\rVert^2\le p\rho^2$, so $\lVert\vect{x}_t\rVert^2 = \vect{z}_t^\top\Sigma_t\vect{z}_t \le \lVert\Sigma_t\rVert\,\lVert\vect{z}_t\rVert^2 \le p\rho^2(\mu\delta+\sigma^2)$.
For the conditional expectation, Remark~\ref{rmk:modified-params} gives $\lVert\mathbb{E}[\xxT{\vect{x}}{t}\mid\mathfrak{E}]\rVert\le\lVert\Sigma_t\rVert \le \mu\delta+\sigma^2$.
Combining the two by the triangle inequality,
\begin{equation}
\label{eq:M-ambient}
\bigl\lVert\xxT{\vect{x}}{t} - \mathbb{E}[\xxT{\vect{x}}{t}\mid\mathfrak{E}]\bigr\rVert \;\le\; p\rho^2(\mu\delta+\sigma^2) + (\mu\delta+\sigma^2) \;=:\; \mathcal{M}_\infty.
\end{equation}
The factor $p$ from coordinate-wise truncation enters only this parameter and hence only the Oja analysis; the noisy power method's matrix Bernstein step instead uses the $\psi_1$ parameter $\mathcal{M}$ derived without truncation in Appendix~\ref{appendix:npm-params}.

\subsection{Block decomposition}

The Oja virtual block size is $B_\zeta = \lceil 1/\zeta \rceil$; for $\zeta \le 1$ the ceiling introduces a multiplicative slack of at most a factor of $2$ in downstream bounds, absorbed into the $\tilde{\Theta}$ notation.
For brevity we use the equality $\zeta B_\zeta = 1$ throughout the derivation.
The product decomposition yields
\begin{equation*}
\prod_{\mathclap{t=(\ell-1)B_\zeta+1}}^{\ell B_\zeta}
  (\mat{I} + \zeta\,\xxT{\vect{x}}{t})
= \underbrace{\prod_{\mathclap{t=(\ell-1)B_\zeta+1}}^{\ell B_\zeta}
  (\mat{I} + \zeta\,
  \mathbb{E}[\xxT{\vect{x}}{\ell B_\zeta}\mid\mathfrak{E}])}_{
  \mat{M}^{\mathrm{Oja}}(\ell)}
+ \mathcal{E}(\ell),
\end{equation*}
Here $\mathbb{E}[\xxT{\vect{x}}{\ell B_\zeta}\mid\mathfrak{E}] = (1-\nu)\Sigma_{\ell B_\zeta}+\Delta_{\ell B_\zeta}$ agrees with the population covariance $\Sigma_{\ell B_\zeta}$ used in the body up to a $\tilde{O}(1/(pT^2))$-order perturbation, absorbed via Remark~\ref{rmk:modified-params}.
We decompose the error as $\mathcal{E}(\ell) = \mathcal{E}_1(\ell) + \mathcal{E}_2(\ell)$ with
\begin{align*}
\mathcal{E}_1(\ell)
  &= \prod_{\mathclap{t=(\ell-1)B_\zeta+1}}^{\ell B_\zeta}
    \bigl(\mat{I} + \tfrac{1}{B_\zeta}\,\xxT{\vect{x}}{t}\bigr)
  - \prod_{\mathclap{t=(\ell-1)B_\zeta+1}}^{\ell B_\zeta}
    \bigl(\mat{I} + \tfrac{1}{B_\zeta}\,
    \mathbb{E}[\xxT{\vect{x}}{t}\mid\mathfrak{E}]\bigr),\\
\mathcal{E}_2(\ell)
  &= \prod_{\mathclap{t=(\ell-1)B_\zeta+1}}^{\ell B_\zeta}
    \bigl(\mat{I} + \zeta\,
    \mathbb{E}[\xxT{\vect{x}}{t}\mid\mathfrak{E}]\bigr)
  - \prod_{\mathclap{t=(\ell-1)B_\zeta+1}}^{\ell B_\zeta}
    \bigl(\mat{I} + \zeta\,
    \mathbb{E}[\xxT{\vect{x}}{\ell B_\zeta}\mid\mathfrak{E}]\bigr).
\end{align*}

\subsection{Bounding the stochastic error}

We apply the following matrix product concentration inequality.

\begin{lemma}[Perturbations of the
identity~{\citep[Corollary~6.1]{huang2020matrix}}, instantiated to the present setting]
\label{lem:concent_prod}
Let $\mat{Z}_1, \ldots, \mat{Z}_{B_\zeta} \in \mathbb{R}^{p\times p}$ be independent random matrices satisfying $\lVert\mathbb{E}\,\mat{Z}_t\rVert
  \leq \mu\delta + \sigma^2$
and $\lVert\mat{Z}_t - \mathbb{E}\,\mat{Z}_t\rVert
  \leq \mathcal{M}_\infty$
for all $t \in [B_\zeta]$.
Then for $\Pi \geq p\,e^{-B_\zeta/(2\mathcal{M}_\infty^2)}$, the product $\mat{Z} = (\mat{I} + \mat{Z}_{B_\zeta}/B_\zeta)\cdots(\mat{I} + \mat{Z}_1/B_\zeta)$ satisfies, with probability at least $1 - \Pi$,
\[
\lVert\mat{Z} - \mathbb{E}\,\mat{Z}\rVert
  \leq e^{\mu\delta+\sigma^2}
    \sqrt{\frac{2e^2\,\mathcal{M}_\infty^2}{B_\zeta}
    \log\frac{p}{\Pi}}\,.
\]
\end{lemma}

On $\mathfrak{E}$, the matrices $\mat{Z}_t = \xxT{\vect{x}}{t}$ satisfy $\lVert\mathbb{E}[\xxT{\vect{x}}{t}\mid\mathfrak{E}]\rVert
  \leq (1-\nu)\lVert\Sigma_t\rVert+\lVert\Delta_t\rVert
  \leq \mu\delta + \sigma^2$
(via Remark~\ref{rmk:modified-params}) and $\lVert\xxT{\vect{x}}{t} - \mathbb{E}[\xxT{\vect{x}}{t}\mid\mathfrak{E}]\rVert
  \leq \mathcal{M}_\infty$
with $\mathcal{M}_\infty$ as in~\eqref{eq:M-ambient}.
Set $\Pi = 1/T^2$: the lemma's hypothesis $\Pi\ge p\,e^{-B_\zeta/(2\mathcal{M}_\infty^2)}$ then reads $B_\zeta\ge 2\mathcal{M}_\infty^2\log(pT^2)$, matching the precondition stated in Lemma~\ref{lem:oja-error}.
Taking a union bound over $L \leq T$ blocks:
\begin{equation}
\label{eq:oja-E1}
\max_{1 \leq \ell \leq L}
  \lVert\mathcal{E}_1(\ell)\rVert
  \leq e^{\mu\delta+\sigma^2}\,
    C_{\mathrm{Oja}}
    \sqrt{\frac{\log(pT^2)}{B_\zeta}}\,,
\end{equation}
with probability at least $1 - 1/T$ conditionally on~$\mathfrak{E}$, where $C_{\mathrm{Oja}} := \sqrt{2}\,e\,\mathcal{M}_\infty$.

\subsection{Bounding the drift error}

Define $\mat{Y}^{(\ell)}_t := \mathbb{E}[\xxT{\vect{x}}{t}\mid\mathfrak{E}] - \mathbb{E}[\xxT{\vect{x}}{\ell B_\zeta}\mid\mathfrak{E}]$, which satisfies $\lVert\mat{Y}^{(\ell)}_t\rVert
  \leq (1-\nu)(\ell B_\zeta - t)\Gamma+\lVert\Delta_t-\Delta_{\ell B_\zeta}\rVert
  \leq (\ell B_\zeta - t)\Gamma$,
where the second step absorbs $\lVert\Delta_t-\Delta_{\ell B_\zeta}\rVert=\tilde{O}(1/(pT^2))$ via Remark~\ref{rmk:modified-params}.
We expand the product difference into its $n$-fold perturbation terms, in which each chosen factor contributes $\zeta\,\mat{Y}^{(\ell)}_t$ and the remaining factors contribute the unperturbed expectation.
The first-order total is then $\zeta\sum_t\lVert\mat{Y}^{(\ell)}_t\rVert\le\zeta\Gamma\,\tfrac{B_\zeta(B_\zeta-1)}{2}\le\tfrac{B_\zeta\Gamma}{2}$, which yields
\begin{align*}
\lVert\mathcal{E}_2(\ell)\rVert
  &\leq \bigl(1 + \zeta(\mu\delta+\sigma^2)\bigr)^{B_\zeta}
    \sum_{n=1}^{B_\zeta}
    \left(\frac{B_\zeta\Gamma}
      {2(1+\zeta(\mu\delta+\sigma^2))}\right)^{\!n}\\
  &\leq e^{\mu\delta+\sigma^2}\,
    \frac{B_\zeta\Gamma}{2(1 - B_\zeta\Gamma/2)}\,,
\end{align*}
where the second line bounds $(1+\zeta(\mu\delta+\sigma^2))^{B_\zeta} \leq e^{\zeta B_\zeta(\mu\delta+\sigma^2)} = e^{\mu\delta+\sigma^2}$ (using $\zeta B_\zeta = 1$) and extends the finite geometric sum to its infinite-series closed form.
The closed form requires $B_\zeta\Gamma < 2$.
At the optimal block size $B_\zeta = \tilde{\Theta}(\Gamma^{-2/3}(p(\mu\delta+\sigma^2))^{2/3})$ (Appendix~\ref{appendix:oja-algorithm-convergence}), we have $B_\zeta\Gamma = \tilde{O}(\Gamma^{1/3}(p(\mu\delta+\sigma^2))^{2/3})$, which tends to $0$ as $\Gamma\to0$.
Hence $B_\zeta\Gamma < 2$ holds for every $\Gamma$ below an explicit threshold, and in particular under the $\Gamma$-regime of Theorem~\ref{thm:ojas-algorithm}.
Writing $\epsilon_{B_\zeta,\Gamma} := (B_\zeta\Gamma)^2/(4 - 2B_\zeta\Gamma)$, which is negligible when $B_\zeta\Gamma \ll 1$, we obtain
\begin{equation}
\label{eq:oja-E2}
\lVert\mathcal{E}_2(\ell)\rVert
  \leq e^{\mu\delta+\sigma^2}
    \left(\frac{B_\zeta\Gamma}{2}
    + \epsilon_{B_\zeta,\Gamma}\right).
\end{equation}

\subsection{Combining the error bounds}

Combining~\eqref{eq:oja-E1} and~\eqref{eq:oja-E2} with the conditioning probability $\mathbb{P}[\mathfrak{E}^c] \leq 1/T$, we obtain: with probability at least $1 - 2/T$,
\begin{equation}
\label{eq:oja-combined}
\max_{1 \leq \ell \leq L}
  \lVert\mathcal{E}(\ell)\rVert
  \leq e^{\mu\delta+\sigma^2}
    \left[C_{\mathrm{Oja}}
      \sqrt{\frac{\log(pT^2)}{B_\zeta}}
    + \frac{B_\zeta\Gamma}{2}
    + \epsilon_{B_\zeta,\Gamma}\right].
\end{equation}
Equivalently, for the scaled error $\mathcal{E}'(\ell) := e^{-(\mu\delta+\sigma^2)}\mathcal{E}(\ell)$:
\begin{equation}
\label{eq:oja-scaled}
\max_{1 \leq \ell \leq L}
  \lVert\mathcal{E}'(\ell)\rVert
  \leq C_{\mathrm{Oja}}
    \sqrt{\frac{\log(pT^2)}{B_\zeta}}
  + \frac{B_\zeta\Gamma}{2}
  + \epsilon_{B_\zeta,\Gamma}\,,
\qquad B_\zeta = 1/\zeta\,.
\end{equation}
The explicit constants in~\eqref{eq:oja-error-main} and~\eqref{eq:oja-scaled} are stated under the idealization $\zeta B_\zeta = 1$.
The ceiling in $B_\zeta = \lceil 1/\zeta\rceil$ inflates the drift term by at most a factor of $2$ for $\zeta\le 1$, absorbed downstream into the $\tilde{\Theta}$ notation of Theorem~\ref{thm:ojas-algorithm}.

\section{Proof of Theorem~\ref{thm:ojas-algorithm}: Oja's algorithm convergence}
\label{appendix:oja-algorithm-convergence}
Throughout this section we operate in the regime $\delta = O(1)$, $\sigma^2 = O(1)$, and $\mu = \Theta(1)$.
Under this regime, $e^{\delta} - 1 = \Theta(\delta)$ (since $\delta \le e^{\delta} - 1 \le \delta\, e^{\delta}$) and $e^{\sigma^2 + \delta} = \Theta(1)$; constants of this form are absorbed into the $\tilde{\Theta}$ notation in the bounds below.

The proof follows the same iterative framework as the noisy power method (Appendix~\ref{appendix:robust-power-method}).
The key difference is that Oja's multiplicative updates exponentiate the spectrum: the signal matrix
\[
\mat{M}^{\mathrm{Oja}}(\ell)
  = \left(\mat{I} + \frac{\Sigma_{\ell B_\zeta}}{B_\zeta}\right)^{\!B_\zeta}
\]
has eigenvalues $(1 + s_i(\Sigma_{\ell B_\zeta})/B_\zeta)^{B_\zeta}$ rather than $s_i(\Sigma_{\ell B_\zeta})$.
Consequently, we must reparametrize the spectral gap and noise level before applying Lemma~\ref{lem:thm2ratio}.

\subsection{Exponentiated spectral parameters}

Under $\mathfrak{E}$, replacing $\Sigma_{\ell B_\zeta}$ with its conditional version $(1-\nu)\Sigma_{\ell B_\zeta}$ and using Remark~\ref{rmk:modified-params} to absorb the bias, the $k$-th and $(k+1)$-th eigenvalues of $\mat{M}^{\mathrm{Oja}}(\ell)$ are
\[
s_k(\mat{M}^{\mathrm{Oja}}) = \left(1 + \frac{\delta+\sigma^2}{B_\zeta}\right)^{\!B_\zeta},
\qquad
s_{k+1}(\mat{M}^{\mathrm{Oja}}) = \left(1 + \frac{\sigma^2}{B_\zeta}\right)^{\!B_\zeta}.
\]
We define the exponentiated spectral gap and noise level:
\begin{align}
\delta_{\mathrm{Oja}}
  &:= s_k(\mat{M}^{\mathrm{Oja}})
     - s_{k+1}(\mat{M}^{\mathrm{Oja}})
  = \left(1+\frac{\delta+\sigma^2}{B_\zeta}\right)^{\!B_\zeta}
    - \left(1+\frac{\sigma^2}{B_\zeta}\right)^{\!B_\zeta},
  \label{eq:delta-oja}\\
\sigma^2_{\mathrm{Oja}}
  &:= s_{k+1}(\mat{M}^{\mathrm{Oja}})
  = \left(1+\frac{\sigma^2}{B_\zeta}\right)^{\!B_\zeta}.
  \label{eq:sigma-oja}
\end{align}
Each parameter increases with $B_\zeta$ toward its limit, $\sigma^2_{\mathrm{Oja}} \to e^{\sigma^2}$ and $\delta_{\mathrm{Oja}} \to e^{\sigma^2}(e^{\delta}-1)$ as $B_\zeta\to\infty$.
For every finite $B_\zeta$ the mean value theorem gives $\delta_{\mathrm{Oja}} = \delta\,(1+\xi/B_\zeta)^{B_\zeta-1}$ for some $\xi\in[\sigma^2,\delta+\sigma^2]$, hence $\delta \le \delta_{\mathrm{Oja}} \le \delta\,e^{\delta+\sigma^2}$ and $\delta_{\mathrm{Oja}} = \Theta(\delta)$; likewise $\sigma^2_{\mathrm{Oja}} = (1+\sigma^2/B_\zeta)^{B_\zeta} \le e^{\sigma^2} = \Theta(1)$.

The subspace drift parameter $\eta_{\mathrm{Oja}}$ is identical to the NPM case:
\begin{equation}
\label{eq:eta-oja}
\eta_{\mathrm{Oja}}
  := \frac{B_{\zeta}\Gamma}
    {\delta - B_{\zeta}\Gamma}\,.
\end{equation}
This is because the column spaces of $\Sigma_{\ell B_\zeta}$ and $\mat{M}^{\mathrm{Oja}}(\ell)$ are identical, since the monotone transformation $s \mapsto (1 + s/B_\zeta)^{B_\zeta}$ preserves eigenvector ordering.

\subsection{Deriving the optimal learning rate}

From the error bound~\eqref{eq:oja-scaled} (neglecting $\epsilon_{B_\zeta,\Gamma}$, negligible since $B_\zeta\Gamma\to0$ as $\Gamma\to0$):
\[
\max_\ell \lVert\mathcal{E}'(\ell)\rVert
  \leq C_{\mathrm{Oja}}
    \sqrt{\frac{\log(pT^2)}{B_\zeta}}
  + \frac{B_\zeta\Gamma}{2}\,.
\]
Minimizing over $B_\zeta = 1/\zeta$ by differentiation yields
\begin{equation}
\label{eq:B-oja-opt}
B_{\zeta_{\mathrm{opt}}}
  = \frac{C_{\mathrm{Oja}}^{2/3}\,
    \log(pT^2)^{1/3}}{\Gamma^{2/3}}\,,
\end{equation}
and the resulting uniform bound:
\begin{equation}
\label{eq:oja-opt-bound}
\max_\ell \lVert\mathcal{E}'(\ell)\rVert
  \leq \tfrac{3}{2}\,
    C_{\mathrm{Oja}}^{2/3}\,
    \log(pT^2)^{1/3}\,\Gamma^{1/3}\,.
\end{equation}
The optimal block size clears Lemma~\ref{lem:oja-error}'s precondition $B_\zeta\ge 2\mathcal{M}_\infty^2\log(pT^2)$ throughout the operating regime.
Substituting $C_{\mathrm{Oja}}=\sqrt{2}\,e\,\mathcal{M}_\infty$ into~\eqref{eq:B-oja-opt} gives $B_{\zeta_{\mathrm{opt}}}=\tilde{\Theta}(\mathcal{M}_\infty^{2/3}\Gamma^{-2/3})$.
This exceeds $2\mathcal{M}_\infty^2\log(pT^2)$ once $\Gamma=\tilde{O}(\mathcal{M}_\infty^{-2})$, which is implied by the $\Gamma$-regime~\eqref{eq:oja-regime-A} verified below: substituting $\mathcal{M}_\infty=\tilde{\Theta}(p(\mu\delta+\sigma^2))$ and using $\delta=O(1)$ matches the right-hand side of~\eqref{eq:oja-regime-A}.

\subsection{Regime verification}

To apply Theorem~\ref{thm:drifting-power-method} with the exponentiated parameters, we verify the standing assumptions~\eqref{eq:amp-standing} in the Oja setting; the drift bound~\eqref{eq:amp-drift} carries over unchanged, since $\eta_{\mathrm{Oja}}$ from~\eqref{eq:eta-oja} equals the NPM drift parameter and $\mat{M}^{\mathrm{Oja}}(\ell)$ shares the leading subspaces of $\Sigma_{\ell B_\zeta}$.

\paragraph{Condition (A): non-trivial error.}
We need $36\,e^{\mu\delta+\sigma^2}\,B_{\zeta_{\mathrm{opt}}}\Gamma
  \leq \delta_{\mathrm{Oja}}$
(equivalently $\epsilon_{\mathrm{Oja}}\le 1/6$, mirroring NPM's regime~(A) in Appendix~\ref{appendix:robust-power-method}), which gives the regime
\begin{equation}
\label{eq:oja-regime-A}
\Gamma
  = \tilde{O}\!\left(
    \frac{\delta^3}
      {(p(\mu\delta + \sigma^2))^2\,\log(pT^2)}\right).
\end{equation}
Here we used $\delta_{\mathrm{Oja}} = \Theta(\delta)$ and $\totalbound_{\mathrm{Oja}} = e^{\mu\delta+\sigma^2}\,
    \frac{3}{2}B_{\zeta_{\mathrm{opt}}}\Gamma$;
under $\mu = \Theta(1)$ and $\delta,\sigma^2 = O(1)$ the factor $e^{\mu\delta+\sigma^2} = \Theta(1)$ is absorbed into $\tilde{O}$.

\paragraph{Condition (B): spectral gap.}
We need $\delta_{\mathrm{Oja}}
  \geq \frac{12}{17}\,\sigma^2_{\mathrm{Oja}}$.
The ratio $\delta_{\mathrm{Oja}}/\sigma^2_{\mathrm{Oja}} = (1+\delta/(B_\zeta+\sigma^2))^{B_\zeta} - 1$ is monotone in $B_\zeta$ and approaches $e^{\delta}-1$ from below as $B_\zeta\to\infty$; in the limit, the condition reduces to $e^{\delta} - 1 \geq 12/17$, i.e., $\delta \geq \log(29/17) \approx 0.53$.
Since the ratio rises continuously to its strict supremum $e^{\delta}-1 > 12/17$ under $\delta > \log(29/17)$, there is a finite block size $B_\zeta^\star(\delta,\sigma^2)$, independent of $\Gamma$, beyond which the finite-$B_\zeta$ inequality $\delta_{\mathrm{Oja}} \geq \tfrac{12}{17}\sigma^2_{\mathrm{Oja}}$ holds.
The operating value $B_{\zeta_{\mathrm{opt}}}=\tilde{\Theta}(\mathcal{M}_\infty^{2/3}\Gamma^{-2/3})$ diverges as $\Gamma\to 0$ by~\eqref{eq:B-oja-opt} and so exceeds $B_\zeta^\star$ throughout the $\Gamma$-regime~\eqref{eq:oja-regime-A}.
Unlike the NPM condition $\delta \geq \tfrac{12}{17}\sigma^2$ (which scales with noise), this is an absolute lower bound on the spectral gap, arising because the exponentiation $x \mapsto e^x$ maps the additive gap $\delta$ to the multiplicative gap $e^{\delta}-1$.
The condition is mild: it requires the signal strength to exceed a fixed threshold, independent of the noise level.

\paragraph{Condition (C): small perturbation ratio.}
Define $\epsilon_{\mathrm{Oja}} := 4\,\totalbound_{\mathrm{Oja}} / \delta_{\mathrm{Oja}}$.
Condition~(A) gives $\epsilon_{\mathrm{Oja}}\le 1/6\le 1/4$ immediately.
The check $\eta_{\mathrm{Oja}}\le\epsilon_{\mathrm{Oja}}/5$ uses the ratio $\eta_{\mathrm{Oja}}/\epsilon_{\mathrm{Oja}}=\delta_{\mathrm{Oja}}/(6\,e^{\mu\delta+\sigma^2}(\delta-B_\zeta\Gamma))$.
The mean-value-theorem bound $\delta_{\mathrm{Oja}}\le\delta\,e^{\delta+\sigma^2}$ from Appendix~\ref{appendix:oja-algorithm-convergence}'s exponentiated-parameter discussion upper-bounds this ratio by $(\delta/(6(\delta-B_\zeta\Gamma)))\cdot e^{(1-\mu)\delta}$, and $\mu\ge 1$ collapses the exponential factor to at most $1$.
The residual $\delta/(6(\delta-B_\zeta\Gamma))$ is the NPM ratio bounded by $1/5$ in Appendix~\ref{appendix:robust-power-method}.

\subsection{Final bound}

With conditions (A)--(C) and the drift bound satisfied, Theorem~\ref{thm:drifting-power-method} applies with the exponentiated parameters.
The estimation error is of order $\epsilon_{\mathrm{Oja}}$:
\begin{equation}
\label{eq:oja-final}
\epsilon_{\mathrm{Oja}}
  \sim \frac{e^{\mu\delta+\sigma^2}\,
    C_{\mathrm{Oja}}^{2/3}\,
    \log(pT^2)^{1/3}\,\Gamma^{1/3}}
    {\delta_{\mathrm{Oja}}}
  = \tilde{\Theta}\Big(
    \frac{\Gamma^{1/3}\,(p(\mu\delta + \sigma^2))^{2/3}}{\delta}\Big)\,,
\end{equation}
where we used $\delta_{\mathrm{Oja}} = \Theta(\delta)$, substituted $C_{\mathrm{Oja}} = \sqrt{2}\,e\,\mathcal{M}_\infty$ and $\mathcal{M}_\infty$ from~\eqref{eq:M-ambient}, and, under $\mu = \Theta(1)$ and $\delta,\sigma^2 = O(1)$, absorbed the constant factor $e^{\mu\delta+\sigma^2} = \Theta(1)$ into $\tilde{\Theta}$.
Adding Theorem~\ref{thm:drifting-power-method}'s initialization tail $O(\beta^{-L}\tau_0\sqrt{pk})$ with $L=T/B_\zeta=\zeta T$ virtual blocks and $\tau_0=T$ absorbs $T\sqrt{pk}$ into $\beta^{-L}$ above Theorem~\ref{thm:drifting-power-method}'s threshold $L > \log(\tau_0\sqrt{pk})/\log\beta$; Lemma~\ref{lem:thm2ratio}(i)'s $\beta\ge 1.4464 > 1/0.7$ then gives $\beta^{-L} \le 0.7^{\zeta T}$.
This bound holds with probability at least $1-O(\tau_0^{-1})-e^{-\Omega(p)}$ over the random initialization (Theorem~\ref{thm:drifting-power-method}); taking $\tau_0=T$ collapses this into $1/T$.
The event $\{\lVert\mathcal{E}(\ell)\rVert \le \totalbound_{\mathrm{Oja}}\text{ for all }\ell\}$ holds with probability at least $1-2/T$ by Lemma~\ref{lem:oja-error}.
Combining the steady-state error~\eqref{eq:oja-final} with this initialization tail gives, with probability greater than $1-3/T-e^{-\Omega(p)}$,
\[
d(\mathrm{ran}(\mat{A}_T),\, \mathrm{Oja}_\mathcal{X}) = \tilde{O}\!\left(
    \frac{(p(\mu\delta+\sigma^2))^{2/3}\,\Gamma^{1/3}}{\delta}
    + 0.7^{\zeta T}\right),
\]
the statement of Theorem~\ref{thm:ojas-algorithm}.

\section{Experiments}
\label{appendix:experimental-details}
\subsection{Synthetic experiments}
\paragraph{Random matrix generation.}
\label{appendix:exp-random}
We generate $(\mat{A}_t)_{t=1}^{T}\in\mathbb{R}^{p \times k}$ as the product of three matrices, $\mat{U}_t\in\stiefel{p}{p}, \mat{D}_t\in\mathbb{R}^{p \times k}$ (diagonal), and $\mat{V}_t\in\stiefel{k}{k}$.
At each iteration we update each matrix and form $\mat{A}_t=\mat{U}_t\,\mat{D}_t\,{\mat{V}_t}^{\top}$.
For $\mat{V}_t$, we draw a Gaussian random matrix, perform a QR decomposition, and take the resulting right factor.
For $\mat{D}_t$, we use a diagonal matrix whose diagonal entries are sampled uniformly from $\{-\sqrt{\delta},\,\sqrt{\delta}\}$.
We construct $\mat{U}_t$ below.
Together, these choices make $\mat{A}_t$ satisfy the adversarial budget ($\Gamma$) and spectral gap ($\delta$) conditions:
\begin{equation}
s_k(\mat{A}_{t}\mat{A}^{\top}_{t}) = \delta\cdot s_k\left[\mat{U}_{t}\ioblock\mat{U}^{\top}_{t}\right]  = \delta\,,
\end{equation}
and
\begin{equation}
\Vert \mat{A}_{t}\mat{A}^{\top}_{t}-\mat{A}_{t-1}\mat{A}^{\top}_{t-1}\Vert = \delta\left\Vert  \mat{U}_{t}\ioblock\mat{U}^{\top}_{t} -  \mat{U}_{t-1}\ioblock\mat{U}^{\top}_{t-1}\right\Vert \leq \Gamma\,.
\end{equation}
We initialize $\mat{U}_0$ to random orthogonal matrix and rotate $\mat{U}_{t-1}$ to generate $\mat{U}_{t}$.
Then the first condition is automatically satisfied.
For the second condition, we restrict the structure of the random rotation matrix $\mat{R}_t$.
Assume that $\mat{U}_{t} = \mat{U}_{t-1}\mat{R}_t$ ($\mat{R}_t\,\mat{R}_t^{\top}=\mat{I}_p$).
Then the second condition becomes:
\begin{equation}
\label{rmg1}
\left\Vert \mat{R}_t\ioblock \mat{R}^{\top}_t - \ioblock \right\Vert \leq \Gamma/\delta < 1.
\end{equation}
(We consider only $\Gamma < \delta$ in our experiments.) To satisfy the above condition, we proceed in two steps.
First, we choose $N \le k$ index pairs $(i_{2n-1}, i_{2n})$ drawn without replacement from $\{1, \ldots, p\}$, with each pair having exactly one index in the signal block $\{1, \ldots, k\}$ and the other in the complement $\{k+1, \ldots, p\}$.
Second, we select angles $\theta_1, \ldots, \theta_{N}$ from the range $\mathopen[-\sin^{-1}(\Gamma/\delta),\,\sin^{-1}(\Gamma/\delta)\mathclose]$.

After initializing $\mat{R}_t$ ($=\mat{I}_{p\times p}$), we write $\theta_n$-rotation matrix on the $2\times2 \ (i_{2n-1}, i_{2n})$-submatrix.
A direct $2\times 2$ block computation shows that the singular values of the matrix on the right-hand side of~\eqref{rmg1} are $|\sin{\theta_1}|,\,\ldots\,|\sin{\theta_{N}}|\,,0\,,\ldots\,,0$.
Since each $\theta_n$ lies in $\mathopen[-\sin^{-1}(\Gamma/\delta),\,\sin^{-1}(\Gamma/\delta)\mathclose]$, the operator norm is bounded by $\Gamma/\delta$, giving~\eqref{rmg1}.

\paragraph{Environments.}
\label{subsec:expenv}
For all synthetic experiments, we instantiate the rotation construction of the previous paragraph with the single index pair $N=1$, $(i_1, i_2)=(1, p)$ (one signal index, one complement index), and the maximal angle $\theta_1 = \sin^{-1}(\Gamma/\delta)$.
We implement algorithms with NumPy and Numba.
We used the dimensions $(p,k) = (100,5)$.
We run each algorithm for $T=144000$ steps, long enough to cover the slowest convergence time among the cases we tested.
Generally, we consider the case $(\delta, \sigma)$=$(1.00, 0.15)$.
When running each algorithm, we maintained B-list and $\zeta$-list respectively as follows:
\begin{align*}
\text{(Noisy Power Method)}\quad \texttt{B\_list} &= [\,2,3,8,10,20,30,40,60,300,400,600,800,1000,\\
&\qquad 1200,1500,1800,2000,3000,4000,6000,8000,9600\,]\\[8pt]
\text{(Oja's Algorithm)}\quad \texttt{zeta\_list} &= [\,\texttt{float}(1/B)\text{ for }B\text{ in }\texttt{B\_list}\,]
\end{align*}

Together, the diagonal sampling $\pm\sqrt{\delta}$ for $\mat{D}_t$ and the single-plane, maximal-angle rotation for $\mat{U}_t$ make the generator realize the worst-case instance that binds the upper bounds rather than a general draw from $\tempuncert$.
Specifically, the diagonal entries $\pm\sqrt{\delta}$ give an isotropic spike ($\mu = 1$), and a single plane rotated at the maximal angle $\sin^{-1}(\Gamma/\delta)$ gives a constant, budget-saturating drift.
Since the guarantees are uniform over $\tempuncert$ and set by the drift budget $\Gamma$, this stresses the $\Gamma$-dependence and the optimal $B,\zeta$ trade-off, though not the condition-number ($\mu$) dependence discussed in Appendix~\ref{appendix:mu-gap}.

\paragraph{Additional synthetic experiment.}
\label{appendix:exp-additional}

\begin{figure}[H]
\begin{subfigure}{0.495\columnwidth}
\centering
\includegraphics[width=0.999\columnwidth]{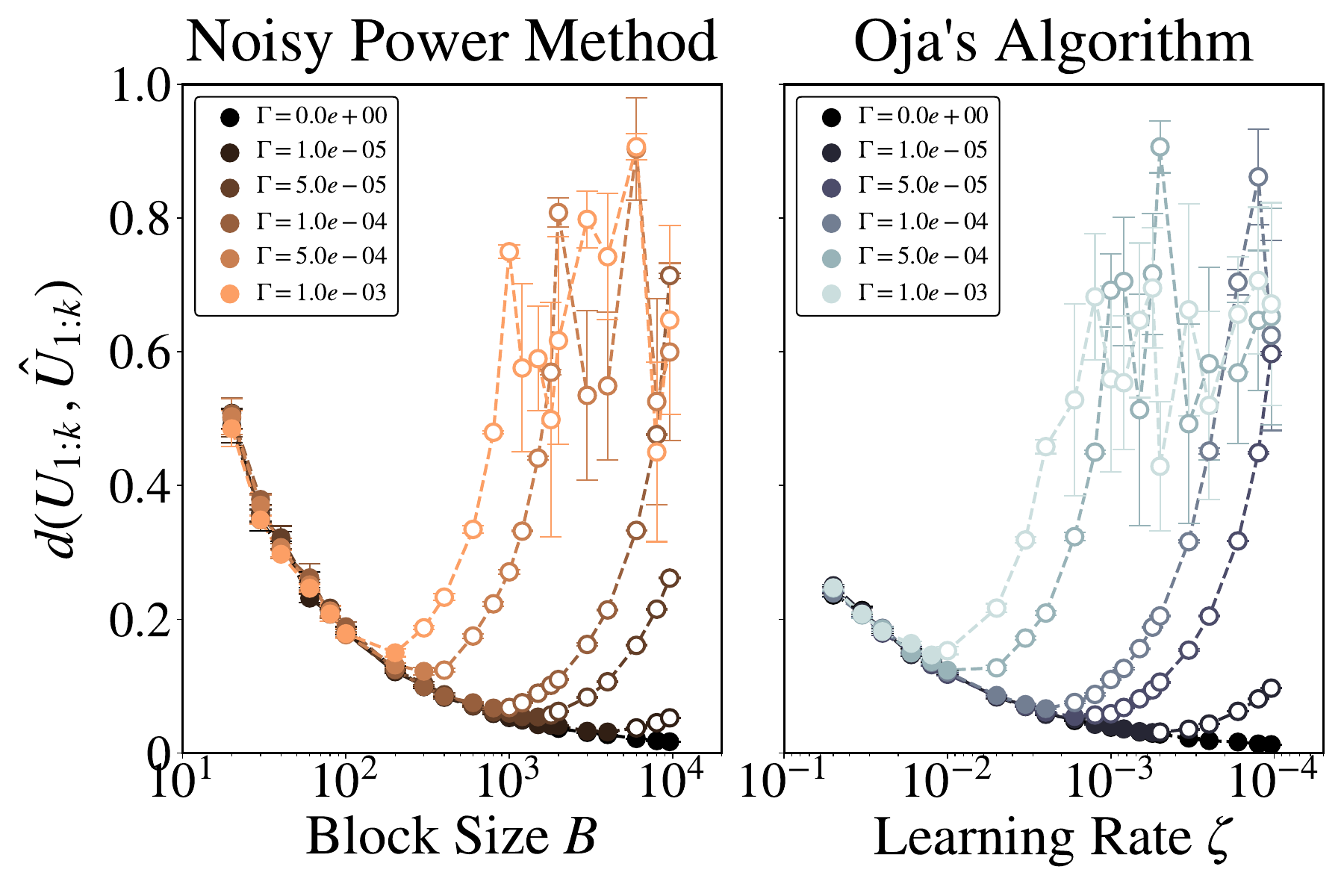}
\caption{Recovery error of the noisy power method and Oja's algorithm at $t=T$, against block size $B$ and learning rate $\zeta$, for the bigger perturbations $\Gamma \in [0.0,1.0e\!\shortminus\!5,5.0e\!\shortminus\!5,1.0e\!\shortminus\!4,5.0e\!\shortminus\!4,1.0e\!\shortminus\!3]$.}
\label{fig:large_perturb}
\end{subfigure}
\begin{subfigure}{0.495\columnwidth}
\centering
\includegraphics[width=0.999\columnwidth]{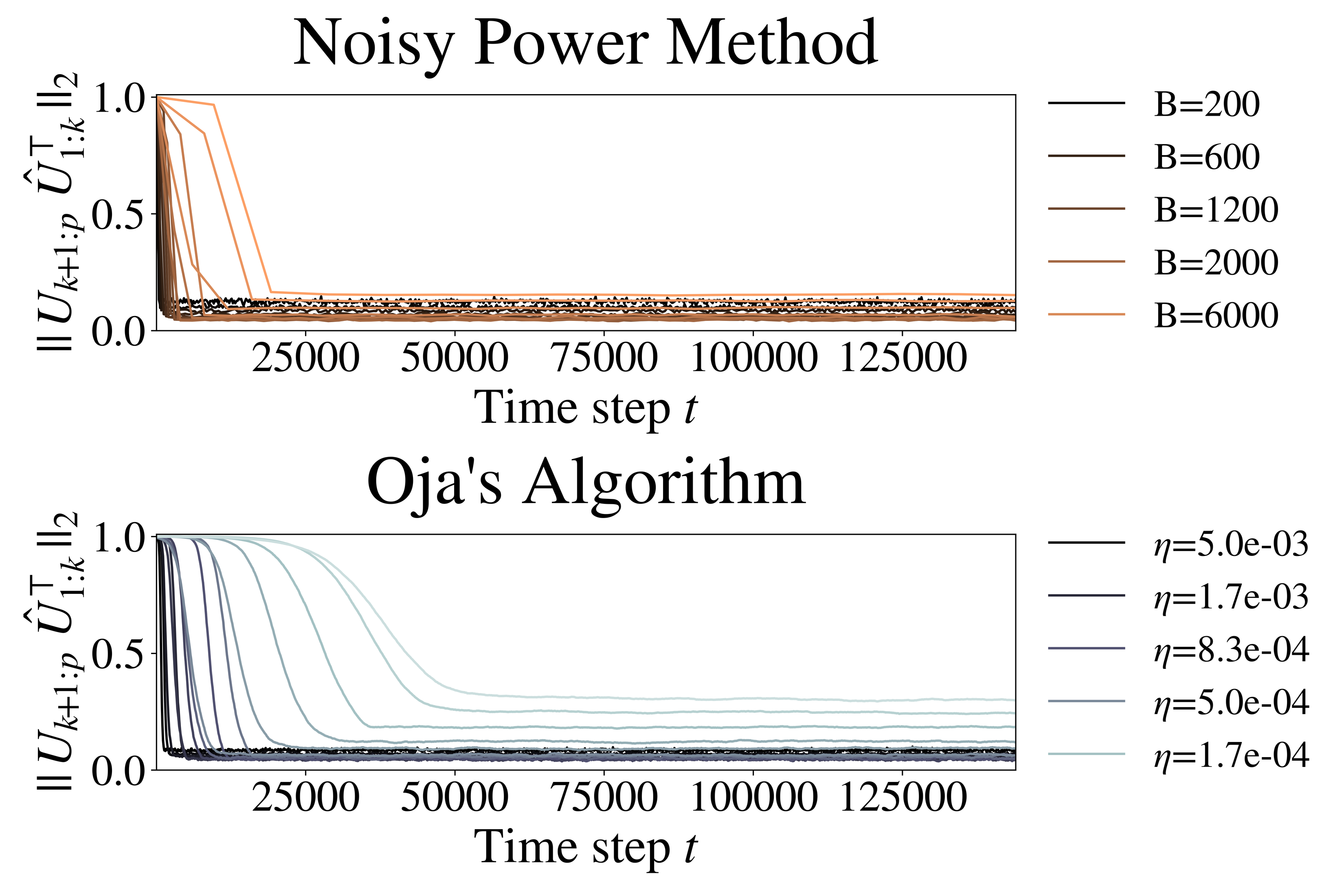}
\caption{Convergence of the noisy power method and Oja's algorithm over the stream, at the block sizes $B$ and learning rates $\zeta$ shown (the panel labels the latter $\eta$) and the $\Gamma$ values of Figure~\ref{fig:adversarial-gamma}.}
\label{fig:convergence-pm}
\end{subfigure}
\caption{Numerical results. We used the setting $(\sigma,\delta,p,k)\!=\!(0.15,1.0,100,5)$.}
\label{fig:additional}
\end{figure}

We provide the additional result for larger perturbations in Figure~\ref{fig:large_perturb}.
We repeated five experiments for each algorithm and learning parameter.
Apart from the magnitude of $\Gamma$ and the number of repetitions, every experimental setting matched that of Figure~\ref{fig:adversarial-gamma}.
Qualitatively, the result shows the same tendency as the result provided in the main paper.

In the second set of experiments in Figure~\ref{fig:convergence-pm}, we show that the noisy power method converges for different values of adversarial budget $\Gamma$ for different block sizes, $B$.
The first observation from these sets of experiments is that there is an optimal block size $B$ that attains the minimum error.
Such behavior is in line with Theorem~\ref{thm:robust-power-method}.
Another key observation from these figures is that a smaller block size implies faster convergence; this is also in line with the dependence on the number of blocks ($L  = \frac{T}{B}$) in Theorem~\ref{thm:robust-power-method}.

\paragraph{Details for each experiment.}
\label{appendix:exp-details}
We expand on the computational setup used to generate Figure~\ref{fig:synthetic_main} and Figure~\ref{fig:additional}.

For the Figure~\ref{fig:adversarial-gamma}, we plot the distance between the estimated subspace and true space at the termination time $T$ for two algorithms with the learning parameters from the Environments paragraph above.
We used four different $\Gamma$ : $[0.0,1.0e\!\shortminus\!5,3.0e\!\shortminus\!5,5.0e\!\shortminus\!5]$.
We run the experiment 10 times, and plot error bar on each marker.

For Figure~\ref{fig:large_perturb}, we used the same methodology as Figure~\ref{fig:adversarial-gamma}, but the larger perturbations $\Gamma$ : $[0.0,1.0e\!\shortminus\!5,5.0e\!\shortminus\!5,1.0e\!\shortminus\!4,5.0e\!\shortminus\!4,1.0e\!\shortminus\!3]$ were used.
We ran 5 experiments each.

For the Figure~\ref{fig:optimal_lp}, we find the empirically optimal value of the block size $B$ of the noisy power method and the learning rate $\zeta$ of Oja's algorithm for various $\Gamma$.
The procedure has three steps.
(i) Using the simulations from Figure~\ref{fig:adversarial-gamma}, which used the learning parameters from the Environments paragraph above, we identify a lower and upper bound on the optimal value of $B$ (and likewise $\zeta$).
(ii) We split the resulting interval into 50 points and run each experiment at a fixed value of $B$ from this interval for 30 runs.
(iii) We compute the average and standard deviation across those 30 runs.
We mark the 50 candidates for each $\Gamma$ with small dots, and plot as a big marker the candidate that incurs the least average convergence error.
The remaining smaller markers for each $\Gamma$ denote candidates whose $\texttt{avg}+\texttt{std}$ is lower than that of the optimal value.

In Figure~\ref{fig:convergence-pm}, we visualize the convergence of two algorithms for various learning parameters $B$ and $\zeta$, reusing the experiment result from Figure~\ref{fig:adversarial-gamma}.

\subsection{S\&P500 stock dataset}
\label{appendix:real-data}

\paragraph{Experimental details.}
\label{appendix:real-detail}
Each daily return is the relative change in adjusted closing price between consecutive trading days.
We ran five experiments for each algorithm and $k=1,2,\ldots,5$, over various regimes of the learning parameters ($B:1\text{--}1600$, $\zeta:10^{-3.5}\text{--}10^{2.5}$).
The data stream is fixed, so runs differ only in the algorithm's random initialization $\hat{\mat{U}}(0)$; each curve of Figure~\ref{fig:sp500_result} shows the mean over the five runs, with error bars at one standard error of the mean, the across-run sample standard deviation divided by $\sqrt{5}$.
For the noisy power method, we discarded the first $T\bmod B$ samples so the remaining data form whole blocks.

\paragraph{Notation.}
\label{appendix:real-nonstationary}
Throughout this subsection we split the stream $\{\vect{x}_t\}_{t\in[T]}$ into consecutive non-overlapping windows of even length $w$, indexed by $\ell$, and write $\hat\Sigma_\ell = \frac{1}{w}\sum_{t}\xxT{\vect{x}}{t}$ for the uncentered second moment of the $\ell$-th window and $\bar\Sigma_\ell$ for the time-average of $\Sigma_t$ over it, so that $\hat\Sigma_\ell$ is unbiased for $\bar\Sigma_\ell$.
We leave the second moment uncentered, because daily returns are close to mean-zero: each stock's sample mean is a few percent of its daily standard deviation, so the rank-one term that centering would remove is negligible against $\hat\Sigma_\ell$.\footnote{Centering moves the terminal window's top-$k$ subspace by less than $0.03$ in projection distance for every $k\le5$.}
Centering on a mean estimated from the same $w$ observations would instead bias the estimate to $(1-1/w)\bar\Sigma_\ell$, the factor Bessel's correction removes.
Every expectation below is over the observations with the covariance path $\{\Sigma_t\}_{t\in[T]}$ held fixed, so $\Delta_\ell$ and the constants of Definition~\ref{defn:tempuncert} are deterministic and only the sampling error is random.
Write $\mat{S}_t := \xxT{\mat{A}}{t}$ for the rank-$k$ signal, so that $\Sigma_t = \mat{S}_t + \sigma^2\mat{I}$ and the noise floor cancels from any difference of covariances, and let $\bar{\mat{S}}_\ell := \bar\Sigma_\ell - \sigma^2\mat{I}$ be the signal part of the $\ell$-th window average.

\paragraph{An additional modeling assumption.}
Definition~\ref{defn:tempuncert} is assumed throughout this subsection rather than tested: the constants below are estimated under it, and whether equity returns admit a rank-$k$ signal with isotropic noise is a separate question.
We assume a stronger version of it.
Definition~\ref{defn:tempuncert} bounds the size of each increment of $\mat{S}_t$ and leaves its direction free, so the signal subspace is free to turn inside a window; we assume instead that the $\mat{S}_t$ of one window share a single $k$-dimensional subspace and differ only in their eigenvalues.
It is needed because the rank-$k$ truncation below is computed on a window rather than at a single $t$.
The assumption is made per window, so the readings at different $w$ are made under different versions of it.

\subsubsection{Estimating $\varbudget$ and $\specgap$ from the stream.}
\label{appendix:real-budget}
In the notation above, Definition~\ref{defn:tempuncert} reads $\oper{\mat{S}_t - \mat{S}_{t-1}} \le \varbudget$ at every $t$.
No single $t$ is observable, so any estimate is forced onto a window.
The budget then controls the change $\Delta_\ell := \bar{\mat{S}}_{\ell+1} - \bar{\mat{S}}_{\ell}$ between consecutive window averages through
\begin{equation}
\label{eq:window-budget}
\Delta_\ell = \frac{1}{w}\sum_{j=1}^{w}\bigl(\mat{S}_{\ell w + j} - \mat{S}_{(\ell-1)w+j}\bigr),
\qquad\text{hence}\qquad
\oper{\Delta_\ell} \;\le\; w\varbudget ,
\end{equation}
since each bracket telescopes into exactly $w$ consecutive increments.
The bound is tight for increments aligned in one direction, and such a sequence lies in $\tempuncert$, so $\varbudget \ge \oper{\Delta_\ell}/w$ is the tightest per-step drift budget consistent with an observed $\Delta_\ell$.

The data supply $\hat\Delta_\ell := \hat\Sigma_{\ell+1} - \hat\Sigma_{\ell}$, from which the noise floor cancels as it does from $\Delta_\ell$, so that $\hat\Delta_\ell = \Delta_\ell + \mat{Z}_\ell$ with a sampling error $\mat{Z}_\ell$ of mean zero.
Being unbiased for $\Delta_\ell$ is not enough, because $\mat{Z}_\ell$ is of the order of the covariance whose change is being measured.
The sub-Gaussian covariance bound~\citep[Theorem~4.7.1]{Vershynin2018} gives $\oper{\hat\Sigma_\ell - \bar\Sigma_\ell} \lesssim \oper{\bar\Sigma_\ell}\bigl(\sqrt{p/w} + p/w\bigr)$, and at $p=137$ and the window lengths used here the bracket is of order one.
To size $\mat{Z}_\ell$ we read $\hat\Delta_\ell$ in the squared Frobenius norm rather than the operator norm, which does not decompose: the square is quadratic, so $\mathbb{E}\mat{Z}_\ell = \mat{0}$ removes the cross term and leaves
\begin{equation}
\label{eq:plugin-square}
\mathbb{E}\frob{\hat\Delta_\ell}^2 \;=\; \frob{\Delta_\ell}^2 + \mathbb{E}\frob{\mat{Z}_\ell}^2 .
\end{equation}

The size of that term follows from independence.
One window's error, $\hat\Sigma_\ell - \bar\Sigma_\ell = w^{-1}\sum_{t}(\xxT{\vect{x}}{t} - \Sigma_t)$, is a sum of independent mean-zero terms, so no cross term stands in its expected squared Frobenius norm and each summand contributes $\mathbb{E}\lVert\vect{x}_t\rVert^4 - \frob{\Sigma_t}^2$.
The two windows are independent of each other as well, so their errors add the same way and $\mathbb{E}\frob{\mat{Z}_\ell}^2 = w^{-2}\sum_t\bigl(\mathbb{E}\lVert\vect{x}_t\rVert^4 - \frob{\Sigma_t}^2\bigr)$ with $t$ running over both.
With $\mathbb{E}\lVert\vect{x}_t\rVert^4 \le C_4'\kappa^4\,\mathrm{tr}(\Sigma_t)^2$, the trace of the fourth-moment bound in Appendix~\ref{appendix:npm-params},
\begin{equation}
\label{eq:plugin-bias}
\mathbb{E}\frob{\mat{Z}_\ell}^2 \;\le\; \frac{2C_4'\kappa^4}{w}\,\max_t\,\mathrm{tr}(\Sigma_t)^2 .
\end{equation}

\paragraph{An estimator free of the noise floor.}
The estimator that avoids it is an inner product of two independent random matrices, and it measures $\frob{\Delta_\ell}$ rather than the operator norm of Definition~\ref{defn:tempuncert}.
Since $\oper{\Delta_\ell} \le \frob{\Delta_\ell}$, that reading over-states the operator-norm drift Definition~\ref{defn:tempuncert} constrains.

It pairs two estimates rather than squaring one, so that the two sampling errors are formed from disjoint observations and their product has mean zero.
Split the observations of window $\ell$ by the parity of their offset, half $h\in\{1,2\}$ holding $\{\vect{x}_{(\ell-1)w+j} : j \equiv h \!\!\pmod 2\}$, so that each half has $w/2$ of them.
Each half gives its own signal estimate: its second moment, truncated to the leading $k$ eigenpairs.
Write $\hat{\mat{S}}^{(h)}_\ell$ for the estimate from half $h$, and $\hat\Delta^{(h)}_\ell := \hat{\mat{S}}^{(h)}_{\ell+1} - \hat{\mat{S}}^{(h)}_{\ell}$ for its window difference.
Our estimator is
\begin{equation}
\label{eq:crossfit-def}
\widehat{Q}_\ell \;:=\; \bigl\langle \hat\Delta^{(1)}_\ell,\; \hat\Delta^{(2)}_\ell \bigr\rangle_{\mathrm{F}},
\qquad
\langle \mat{A},\mat{B}\rangle_{\mathrm{F}} \;=\; \sum_{i,j}\mat{A}_{i,j}\mat{B}_{i,j} \;=\; \tr(\mat{A}^\top\mat{B}) .
\end{equation}
The halves are independent under the model and the Frobenius inner product is bilinear, so the expectation factors across its two slots:
\begin{equation}
\label{eq:crossfit}
\mathbb{E}\,\widehat{Q}_\ell
  \;=\; \bigl\langle \mathbb{E}\hat\Delta^{(1)}_\ell,\; \mathbb{E}\hat\Delta^{(2)}_\ell \bigr\rangle_{\mathrm{F}} .
\end{equation}

Neither expectation in~\eqref{eq:crossfit} equals $\Delta_\ell$, however.
Half $h$ averages $\mat{S}_t$ over the offsets it holds rather than over the whole window.
Write $\bar{\mat{S}}^{(h)}_\ell$ for that half-window average and $\Delta^{(h)}_\ell := \bar{\mat{S}}^{(h)}_{\ell+1} - \bar{\mat{S}}^{(h)}_{\ell}$ for its difference across the pair, so that $\mathbb{E}\hat\Delta^{(h)}_\ell = \Delta^{(h)}_\ell$.
The truncation is where the second assumption is spent: with every $\mat{S}_t$ of a window in one $k$-dimensional subspace, $\bar{\mat{S}}^{(h)}_\ell$ has rank $k$, so keeping $k$ eigenpairs of $\bar{\mat{S}}^{(h)}_\ell + \sigma^2\mat{I}$ recovers it rather than cutting part of it away.
Half and whole differ by an alternating sum,
\begin{equation}
\label{eq:parity-bias}
\bar{\mat{S}}^{(h)}_\ell - \bar{\mat{S}}_\ell \;=\; \frac{1}{w}\sum_{j=1}^{w} (-1)^{j+h}\,\mat{S}_{(\ell-1)w+j} ,
\end{equation}
whose consecutive terms pair into $w/2$ single increments of $\mat{S}_t$, each of operator norm at most $\varbudget$.
So $\oper{\bar{\mat{S}}^{(h)}_\ell - \bar{\mat{S}}_\ell} \le \varbudget/2$ in each of the two windows, and $\oper{\Delta^{(h)}_\ell - \Delta_\ell} \le \varbudget$ across the pair.

$\widehat{Q}_\ell$ is biased too, but its bias is drift where that of $\frob{\hat\Delta_\ell}^2$ is sampling error, and drift is the smaller of the two: one is set by how far $\Sigma_t$ moves inside a window, the other by the size of $\Sigma_t$ itself.
Equation~\eqref{eq:parity-bias} changes sign with $h$, so the two halves miss $\Delta_\ell$ by opposite matrices $\pm\mat{G}$, with $\mat{G} := \Delta^{(1)}_\ell - \Delta_\ell$, and~\eqref{eq:crossfit} reads
\begin{equation}
\label{eq:crossfit-bias}
\mathbb{E}\,\widehat{Q}_\ell \;=\; \bigl\langle \Delta_\ell + \mat{G},\; \Delta_\ell - \mat{G} \bigr\rangle_{\mathrm{F}} \;=\; \frob{\Delta_\ell}^2 - \frob{\mat{G}}^2 .
\end{equation}
The two cross terms in $\mat{G}$ cancel against each other and its square is subtracted, so~\eqref{eq:crossfit-def} under-states $\frob{\Delta_\ell}^2$ rather than inflating it.
Taking $h=1$ in the bound just derived gives $\frob{\mat{G}}^2 \le p\,\varbudget^2$, so against~\eqref{eq:plugin-bias} the bias of $\widehat{Q}_\ell$ is the smaller of the two when $p\,w\,\varbudget^2 \le 2C_4'\kappa^4\max_t\mathrm{tr}(\Sigma_t)^2$.
Substituting $w\varbudget \lesssim \oper{\Sigma_t}$, the condition under which a window average estimates $\Sigma_t$ at all, this holds once
\begin{equation}
\label{eq:bias-crossover}
w \;\gtrsim\; \frac{p}{2C_4'\kappa^4}\Bigl(\frac{\oper{\Sigma_t}}{\mathrm{tr}(\Sigma_t)}\Bigr)^{2} ,
\end{equation}
the bracket being the squared reciprocal effective rank of $\Sigma_t$, at most one for any covariance.

\begin{figure}[t]
\centering
\includegraphics[width=.62\textwidth]{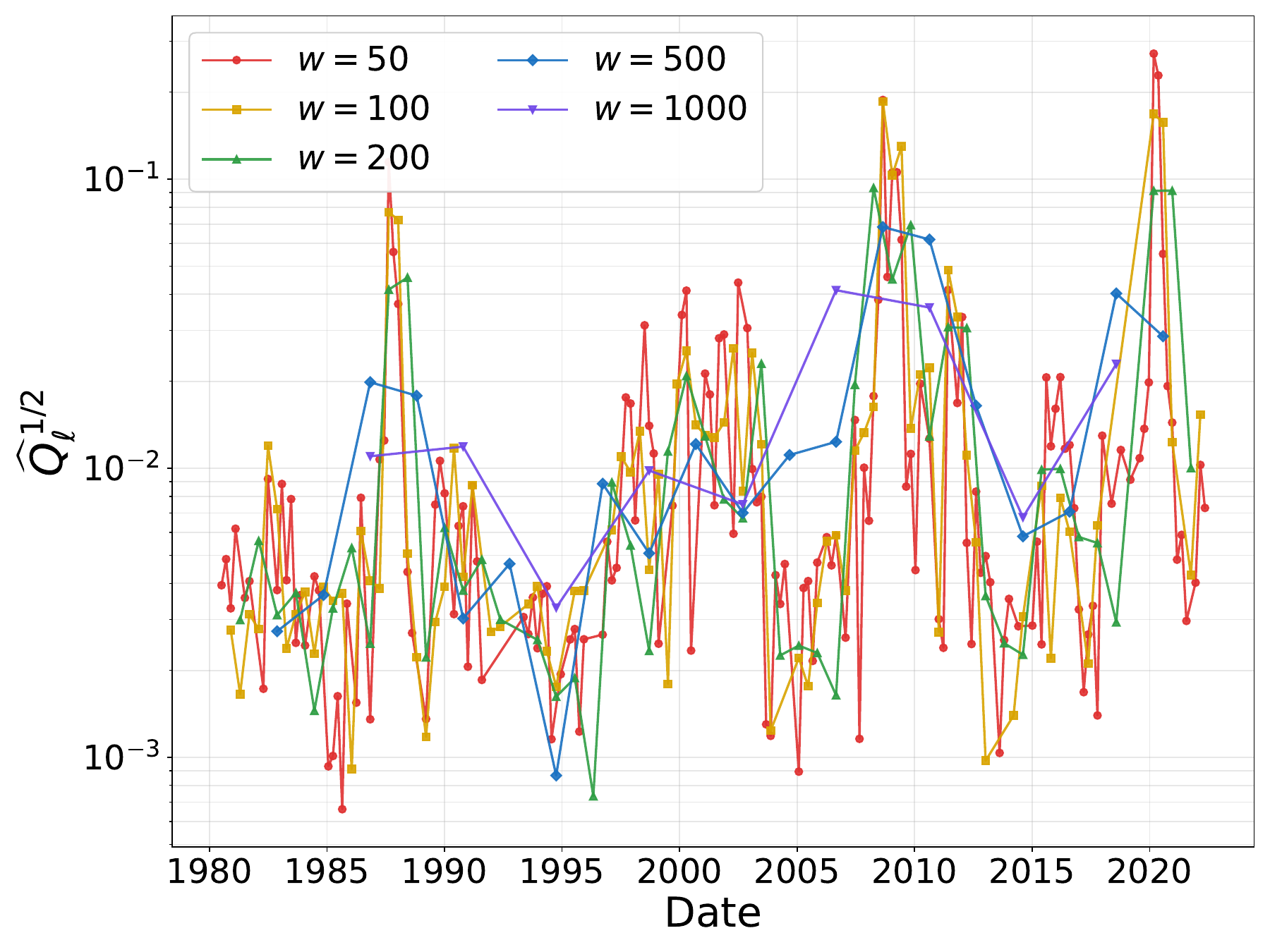}
\caption{Signal drift in the S\&P500 daily-return stream, read as $\widehat{Q}_\ell^{1/2}$ between consecutive length-$w$ windows, on a logarithmic scale, for $w$ from $50$ to $1000$. Window pairs whose estimate is not positive are omitted: a quarter of them at $w=50$, none at $w\ge500$. Shorter windows are not plotted, since more than a third of the pairs are unresolved already at $w=20$.}
\label{fig:sp_drift}
\end{figure}

Figure~\ref{fig:sp_drift} plots $\widehat{Q}_\ell^{1/2}$ at each $\ell$ across the four decades of the stream, before any averaging over $\ell$.
The drift is not spread evenly: at $w=100$ its two largest values, both above thirty times the median, are the windows ending August 2008 and March 2020.

\paragraph{Reading the constants from the stream.}
Three constants are read from the same windows.
The noise level $\hat\sigma^2$ is the mean of the trailing $p-k$ eigenvalues of a half's second moment.
The budget is $\hat{\varbudget}(w) = \widehat{Q}_\ell^{1/2}/w$: the estimator of $\frob{\Delta_\ell}$ above, divided by $w$ to invert~\eqref{eq:window-budget}.
It exceeds $\oper{\Delta_\ell}/w$, the tightest budget the observed $\Delta_\ell$ forces, and is that reading rather than an estimate of $\varbudget$ itself.
The estimated spectral gap is $\hat{\specgap} = s_k(\hat\Sigma_\ell) - \hat\sigma^2$, and Definition~\ref{defn:tempuncert} asks for its bound at every $t$, so we report the minimum over the length-$200$ windows the budget is read on.
That is a third to a half of the median over the same windows, and the comparison below holds under it.

\paragraph{Validating the budget estimator.}
Before $\hat{\varbudget}$ is read on the data we run it on synthetic streams, for two reasons that do not apply to the other two constants.
$\hat\sigma^2$ and $\hat{\specgap}$ are functions of one window's eigenvalues, so Weyl's inequality (Lemma~\ref{weyl}) bounds their deviation by the concentration bound above; $\widehat{Q}_\ell$ is not covered by it.
And what was derived for $\widehat{Q}_\ell$ is an expectation at a fixed $\ell$, where the stream offers one draw and no repeated sampling, so the reported figure averages across $\ell$ instead and its target is the stream average of $\frob{\Delta_\ell}^2$.

The streams come from the generator of Appendix~\ref{appendix:exp-random}, run at this dataset's dimension, length and spectral scale: $p = 137$, $T = 10677$, $k = 1$, $\specgap = 1.6\times10^{-2}$ and $\sigma^2 = 3.0\times10^{-4}$.
Its single-plane maximal-angle rotation holds $\oper{\mat{S}_t-\mat{S}_{t-1}}$ at $\varbudget$ for every $t$, so $\frob{\Delta_\ell}$ is one value throughout and the average across $\ell$ becomes an average over repetitions of a known target.
Over eight such streams at $\varbudget = 10^{-5}$ the estimate runs between $0.76$ and $1.15$ times the planted value at $w=200$, averaging $0.97$, and tightens to $0.89$--$0.94$ at $w=1000$; at $w=100$ it ranges from zero to $1.96$ and does not resolve.
Over eight more at $\varbudget = 0$ the averaged $\widehat{Q}_\ell$ is non-positive in three of the eight at every $w$ and, where positive, never above a tenth of the value measured on the data.

\begin{table}[h]
\centering
\caption{Drift budget and spectral gap estimated from the S\&P500 stream. Left: $\widehat{Q}_\ell^{1/2}$, the square root of $\widehat{Q}_\ell$ averaged over consecutive length-$w$ window pairs, and the implied per-step drift budget $\hat{\varbudget}(w) = \widehat{Q}_\ell^{1/2}/w$. Right: $\hat{\specgap}$ per $k$, minimized over the same length-$200$ windows, and its ratio to $\hat{\varbudget}(200)$. A hat marks a quantity estimated from the stream, as against the model constants $\varbudget$ and $\specgap$ of Definition~\ref{defn:tempuncert}.}
\label{tab:sp-budget}
\small
\begin{tabular}{ccc|ccc}
\toprule
$w$ & $\widehat{Q}_\ell^{1/2}$ & $\hat{\varbudget}(w)$ & $k$ & $\hat{\specgap}$ & $\hat{\varbudget}/\hat{\specgap}$\\
\midrule
$100$  & $0.036$ & $3.6\times10^{-4}$ & $1$ & $3.4\times10^{-3}$ & $0.04$\\
$200$  & $0.028$ & $1.4\times10^{-4}$ & $2$ & $1.0\times10^{-3}$ & $0.14$\\
$500$  & $0.025$ & $5.0\times10^{-5}$ & $3$ & $7.1\times10^{-4}$ & $0.20$\\
$1000$ & $0.021$ & $2.1\times10^{-5}$ & $4$ & $5.4\times10^{-4}$ & $0.26$\\
       &         &                    & $5$ & $4.7\times10^{-4}$ & $0.30$\\
\bottomrule
\end{tabular}
\end{table}

\paragraph{Results.}
Table~\ref{tab:sp-budget} reports the two readings taken on the S\&P500 stream: $\hat{\varbudget}(w)$ at four window lengths, and $\hat{\specgap}$ for each $k$ the experiments run, with the ratio between them.
$\hat{\varbudget}$ falls steadily as $w$ grows, and that belongs to the stream rather than to the reading: run on the aligned-increment streams above, the same estimator returns a $\hat{\varbudget}$ with no trend in $w$.
The inversion at~\eqref{eq:window-budget} is tight only for aligned increments, so no single per-step budget is identified here and $\hat{\varbudget}(w)$ is the budget forced at horizon $w$.
Behind each of them is $\oper{\Delta_\ell}/w$, a lower bound on $\varbudget$, so the comparison should use the largest reading the calibration supports, $\hat{\varbudget}(200)$ at the shortest window the estimator resolves.
There the standing hypothesis $\specgap > \varbudget$ holds for every $k$ the experiments run, by a margin that narrows as $k$ grows, since $\hat{\specgap}$ falls with $k$ while the budget does not.
That is the same effect Section~\ref{numerical} reports as a smaller gap raising the minimum achievable error.

\section{Discussion}

\subsection{On the \texorpdfstring{$\mu$}{mu} gap}
\label{appendix:mu-gap}

This subsection diagnoses the condition-number slack $\mu$ that the NPM upper bound (Theorem~\ref{thm:robust-power-method}) has but the minimax lower bound (Theorem~\ref{thm:lower_bound}) does not.
The four diagnostic paragraphs below (plus a supporting derivation closing the last one) ask, in order, (i) where $\mu$ enters; (ii) why drift forces a finite block and brings $\mu$ into the NPM rate, with $s_k$ rather than $s_1$ setting recovery difficulty; (iii) whether the lower bound can be sharpened to recover $\mu$; and (iv) whether the upper bound can drop $\mu$ via a relative-perturbation route.

\paragraph{Where the $\mu$-dependence enters.}
The minimax lower bound (Theorem~\ref{thm:lower_bound}) is free of the condition-number $\mu$ for a structural reason: every candidate in Step~1 of the reduction (Appendix~\ref{appendix:lower-bound}) uses the isotropic spike $\sqrt{\delta}\,\mat{U}^{(i)}_t$, so $\mu = 1$ on every candidate, whereas the upper bound is uniform over $\tempuncert$ with $\mu \ge 1$.
Oja's bound (Theorem~\ref{thm:ojas-algorithm}) assumes $\mu=\Theta(1)$, so its $\mu$-dependence collapses by assumption; the diagnostic below applies to NPM.

\paragraph{Drift forces a finite block.}
Subspace recovery difficulty is set by the smallest signal eigenvalue $s_k$, not the largest, so the upper bound's dependence on the $s_1/s_k$ ratio $\mu$ is suspect.
The slack persists because drift forces a finite block: with a fixed covariance the estimator could grow the block to $B=T\to\infty$ and the per-block sampling error $\sqrt{\mathcal{V}/B}$ (Lemma~\ref{lem:power-error}) vanishes, leaving the $\mu$ in $\mathcal{V}$ to set only the decay rate, not the limit.
Under drift, stale samples describe a rotated subspace, capping $B$ at a finite value before the moving-target bias dominates; the per-block error no longer vanishes and the $\mu$ inside it enters the rate.
The technical channel is the operator-norm and trace bounds $\norm{\Sigma_t} \le \mu\delta + \sigma^2$ and $\tr(\Sigma_t) \le k\mu\delta + p\sigma^2$ on $\Sigma_t$, which set $\mathcal{M}$ and $\mathcal{V}$ for the sub-exponential matrix Bernstein behind Lemma~\ref{lem:power-error}.
The resulting NPM slack ranges from $\mu^{1/3}$ in the noise-energy-dominated regime ($p\sigma^2\gtrsim k\mu\delta$) to $\mu^{2/3}$ in the signal-energy-dominated regime ($p\sigma^2\lesssim k\mu\delta$).

\paragraph{The lower bound cannot recover $\mu$.}
To diagnose whether the gap can be closed from the lower-bound side, we examine the simplest non-trivial tilt of the spike spectrum: replace $\sqrt{\delta}\,\mat{U}^{(i)}_t$ with $\mat{U}^{(i)}_t D^{1/2}$ where $D = \mathrm{diag}(\lambda, \delta, \dots, \delta)$ and $\lambda \in [\delta, \mu\delta]$.
Metric separation between any two candidates is unaffected, since the subspace pair $(\mat{U}^{(i)}, \mat{U}^{(j)})$ is fixed by the geometric construction.
We show below that the per-pair KL is monotonically non-decreasing in $\lambda$ over $[\delta, \mu\delta]$, so the tilt only tightens the Fano constraint and forces $s$ downward.

The Appendix~\ref{constructAt} geodesic moves all $k$ principal angles uniformly via the principal rotation $\mathcal{G}$, so every column of $\mat{U}^{(i)}_t$ rotates between consecutive timesteps and the tilt direction (whichever column receives $\lambda$) rotates with it.
The per-step operator-norm drift takes the form $\delta\theta_P + (\lambda-\delta)\theta_u$ instead of $\delta\theta_P$, where $\theta_P = \sin\psi^{(i)}_{\max}$ is the largest principal angle between the leading subspaces $[\mat{U}^{(i)}_{t-1}]$ and $[\mat{U}^{(i)}_t]$ and $\theta_u = \sin\psi^{(i)}_{\mathrm{tilt}}$ the principal angle of the rank-1 tilt direction $\mat{u}^{(i)}_t := \mat{U}^{(i)}_t \vect{e}_1$ between the same two timesteps (Definition~\ref{def:principal_angles}).
Holding this drift below $\Gamma$ forces a smaller per-step rotation, so the cap sharpens from $\Gamma/\delta$ to $\Gamma/\lambda$ and the lazy-allocation stretch lengthens from $3s\delta/\Gamma$ to $3s\lambda/\Gamma$; combined with the per-step KL monotonicity below, this pushes $s$ downward.
The only way to keep the tilt direction static is to restrict the packing to candidates whose terminal subspaces share $\vect{e}_1$ as a common leading column,
\[
\mat{U}^{(i)}_T = \begin{bmatrix} 1 & \vect{0}^\top \\ \vect{0} & \widetilde{\mat{U}}^{(i)}_T \end{bmatrix},
\qquad
\widetilde{\mat{U}}^{(i)}_T \in \stiefel{k-1}{p-1},
\]
packed via Gilbert--Varshamov on the trailing $\grass{k-1}{p-1}$ and with the geodesic motion confined to that sub-Grassmannian.
Setting $D = \mathrm{diag}(\lambda,\delta,\dots,\delta)$ assigns the tilt-eigenvalue $\lambda$ to the fixed $\vect{e}_1$, so $V_{11}=1$ for every pair, the $\lambda$-dependent terms in~\eqref{eq:kl-restricted} vanish, and the analysis reduces to Appendix~\ref{constructAt}'s on the $\grass{k-1}{p-1}$ packing, yielding the Appendix~\ref{constructAt} bound with $k$ replaced by $k-1$.
Neither version improves over the isotropic case.

For two candidates with covariances $\Sigma^{(\ell)} = \sigma^2 \mat{I} + \mat{U}^{(\ell)} D \mat{U}^{(\ell)\top}$ sharing a common spectrum $D = \mathrm{diag}(d_1, \dots, d_k)$, Lemma~\ref{lem:spiked-kl} gives
\begin{equation}
\label{eq:kl-anisotropic}
\kl{\Pprob_i}{\Pprob_j}
  = \frac{1}{2\sigma^2}\sum_{a=1}^{k} c(d_a) \bigl[d_a - (V D V^\top)_{aa}\bigr],
\qquad
c(d) := \frac{d}{\sigma^2+d},
\quad V := (\mat{U}^{(j)})^\top \mat{U}^{(i)}.
\end{equation}
Substitute $D = \mathrm{diag}(\lambda, \delta, \dots, \delta)$ into~\eqref{eq:kl-anisotropic}.
Using $(V D V^\top)_{aa} = \delta\,(V V^\top)_{aa} + (\lambda-\delta)\,V_{a1}^2$ in each bracket and rearranging the top row,
\begin{equation}
\label{eq:kl-restricted}
2\sigma^2\,\kl{\Pprob_i}{\Pprob_j}
 = c(\lambda)\bigl[\delta\bigl(1 - (V V^\top)_{11}\bigr) + (\lambda-\delta)\bigl(1 - V_{11}^2\bigr)\bigr]
 + c(\delta)\sum_{a\ge 2}\bigl[\delta\bigl(1 - (V V^\top)_{aa}\bigr) - V_{a1}^2(\lambda-\delta)\bigr].
\end{equation}
Since $V$ records the cosines of principal angles between two $k$-dimensional subspaces, its singular values lie in $[0, 1]$; consequently $V V^\top \preceq \mat{I}$ and $V^\top V \preceq \mat{I}$, and every diagonal entry of either matrix lies in $[0, 1]$.

Differentiating~\eqref{eq:kl-restricted} in $\lambda$,
\begin{equation*}
2\sigma^2\,\frac{d}{d\lambda}\kl{\Pprob_i}{\Pprob_j}
 = c'(\lambda)\underbrace{\bigl[\delta\bigl(1 - (V V^\top)_{11}\bigr) + (\lambda-\delta)\bigl(1 - V_{11}^2\bigr)\bigr]}_{\ge\,0}
 + c(\lambda)\bigl(1 - V_{11}^2\bigr) - c(\delta)\sum_{a\ge 2} V_{a1}^2.
\end{equation*}
The underbraced bracket is a sum of two non-negative summands on $[\delta, \mu\delta]$, and $c'(\lambda) > 0$, so the first term is non-negative.
For the remaining two, $\sum_{a\ge 2} V_{a1}^2 = (V^\top V)_{11} - V_{11}^2 \le 1 - V_{11}^2$, and $c(\lambda) \ge c(\delta)$ when $\lambda \ge \delta$:
\begin{equation*}
c(\lambda)\bigl(1 - V_{11}^2\bigr) - c(\delta)\textstyle\sum_{a\ge 2} V_{a1}^2
 \;\ge\; c(\lambda)\bigl(1 - V_{11}^2\bigr) - c(\delta)\bigl(1 - V_{11}^2\bigr)
 \;=\; \bigl[c(\lambda) - c(\delta)\bigr]\bigl(1 - V_{11}^2\bigr)
 \;\ge\; 0.
\end{equation*}
Hence each per-step $\kl{\Pprob_i}{\Pprob_j}$ is non-decreasing in $\lambda$; the trajectory KL inherits this monotonicity by additivity over the active stretch~\eqref{eq:kl_decomposition}, and so does the candidate-average KL that enters the Fano constraint.
Since the packing number $M$ is essentially scale-invariant in $s$ (Step~4 of Appendix~\ref{appendix:lower-bound}), a larger KL only forces $s$ downward, so the isotropic case $\lambda = \delta$ gives the tightest lower bound this tilt family can produce.

\paragraph{The upper-bound side: relative perturbation.}
As shown above, $\mu$ enters our upper bound through the matrix-concentration parameters $\mathcal V$ and $\mathcal M$ that bound $\|\mathcal{E}(\ell)\|$ via matrix Bernstein in the original basis.
An alternative is to derive a bound on the same $\|\mathcal{E}(\ell)\|$ by whitening with $\Sigma_{\ell B}^{-1/2}$ first and converting back via $\|\mathcal{E}(\ell)\| \le \|\Sigma_{\ell B}\|\cdot\|\Sigma_{\ell B}^{-1/2}\mathcal{E}(\ell)\Sigma_{\ell B}^{-1/2}\|$.
Both routes feed Appendix~\ref{appendix:amplification}'s chain through the same input $\|\mathcal{E}(\ell)\|$ in operator norm, the same per-step contraction $\beta$ of order $(\delta+\sigma^2)/\sigma^2$, and the same $L$-block amplification.
They differ only in the operator-norm bound on $\|\mathcal{E}(\ell)\|$ they produce:
\begin{align*}
\text{matrix Bernstein (our route)}:&\quad \|\mathcal{E}(\ell)\| \;\lesssim\; \sqrt{\mathcal{V}\log T/B} + B\Gamma/2,\\
\text{whitening + conversion}:&\quad \|\mathcal{E}(\ell)\| \;\le\; \|\Sigma_{\ell B}\|\cdot\bigl\|\Sigma_{\ell B}^{-1/2}\mathcal{E}(\ell)\Sigma_{\ell B}^{-1/2}\bigr\| \;\le\; (\mu\delta+\sigma^2)\bigl[\sqrt{p\log T/B} + B\Gamma/(2\sigma^2)\bigr],
\end{align*}
with $\mathcal{V}\propto(p\sigma^2+k\mu\delta)(\sigma^2+\mu\delta)$ on our side; the whitened concentration $\sqrt{p\log T/B} + B\Gamma/(2\sigma^2)$ on the right is derived at the end of this subsection.
Both bounds depend on $\mu$: the whitened concentration is itself $\mu$-free, but the conversion factor $\|\Sigma_{\ell B}\| \le \mu\delta+\sigma^2$ brings $\mu$ back.
Substituting each bound into Appendix~\ref{appendix:amplification}'s tolerance $\epsilon_{\mathrm{tol}}\propto\|\mathcal{E}(\ell)\|/\delta$ and optimizing the block size $B$ (balance stochastic and drift, chain over $L=T/B$ blocks) gives final NPM rates
\begin{align*}
\text{matrix Bernstein}:&\quad \epsilon \;\lesssim\; \bigl(\Gamma(p\sigma^2+k\mu\delta)(\sigma^2+\mu\delta)/\delta^3\bigr)^{1/3},\\
\text{whitening + conversion}:&\quad \epsilon \;\lesssim\; \bigl(p\Gamma(\mu\delta+\sigma^2)^3/(\sigma^2\delta^3)\bigr)^{1/3}.
\end{align*}
Cubing the ratio of the whitening bound to the matrix-Bernstein bound gives $p(\mu\delta+\sigma^2)^2/(\sigma^2(p\sigma^2+k\mu\delta))$.
Bounding the denominator by $p\sigma^2(\sigma^2+\mu\delta)$ using $k\le p$, the ratio is at least $(\sigma^2+\mu\delta)/\sigma^2 \ge 1$.
The matrix-Bernstein bound is therefore at least as tight as the whitening alternative, so the whitening perspective does not improve the final NPM rate under Appendix~\ref{appendix:amplification}'s existing framework.

The factor $\|\Sigma_{\ell B}\|\sim\mu\delta+\sigma^2$ enters the whitening route as an unavoidable multiplicative cost, larger than what matrix Bernstein extracts from the variance of $\xxT{\vect x}{t}-\Sigma_t$ directly.
Whether the $\mu$ in our final bound is an artifact of Appendix~\ref{appendix:amplification}'s operator-norm analysis (removable by a different analytical structure) or is genuine remains open.

\paragraph{Derivation of the whitened concentration on $\mathcal{E}(\ell)$.}
Write $\mathcal{E}(\ell)=\mathcal{E}_1(\ell)+\mathcal{E}_2(\ell)$ with
$\mathcal{E}_1(\ell)=\frac1B\sum_t(\xxT{\vect{x}}{t}-\Sigma_t)$ and
$\mathcal{E}_2(\ell)=\frac1B\sum_t(\Sigma_t-\Sigma_{\ell B})$, and whiten each by
$\Sigma_{\ell B}^{-1/2}$.

\emph{Stochastic part.} Set $\vect{y}_t:=\Sigma_{\ell B}^{-1/2}\vect{x}_t
=\Sigma_{\ell B}^{-1/2}\Sigma_t^{1/2}\vect{z}_t$, so
$\Sigma_{\ell B}^{-1/2}\mathcal{E}_1(\ell)\Sigma_{\ell B}^{-1/2}
=\frac1B\sum_t(\xxT{\vect{y}}{t}-\Sigma_y^{(t)})$,
$\;\Sigma_y^{(t)}:=\Sigma_{\ell B}^{-1/2}\Sigma_t\Sigma_{\ell B}^{-1/2}$.
Under
$B\Gamma\lesssim\sigma^2$,
\begin{equation*}
\|\vect{y}_t\|_{\psi_2}\le C\kappa,\qquad
\bigl\|\Sigma_y^{(t)}-\mat{I}\bigr\|
\le\|\Sigma_{\ell B}^{-1}\|\,\|\Sigma_t-\Sigma_{\ell B}\|
\le \tfrac{B\Gamma}{\sigma^2}=O(1),
\end{equation*}
hence $\|\Sigma_y^{(t)}\|=O(1)$ with effective rank $\Theta(p)$.
The sub-Gaussian
covariance bound~\citep[Theorem~4.7.1]{Vershynin2018} gives, with probability
$\ge1-1/T$,
\begin{equation*}
\bigl\|\Sigma_{\ell B}^{-1/2}\mathcal{E}_1(\ell)\Sigma_{\ell B}^{-1/2}\bigr\|
\;\lesssim\; \sqrt{\mathcal{V}_y\log(2pT^2)/B}+\mathcal{M}_y\log(2pT^2)/B,
\qquad \mathcal{V}_y\lesssim\kappa^4 p,\;\; \mathcal{M}_y\lesssim\kappa^2 p,
\end{equation*}
both $\mu$-free, against $\mathcal{V}\le C_{\mathcal{V}}\kappa^4(k\mu\delta+p\sigma^2)
(\mu\delta+\sigma^2)$ and $\mathcal{M}\le\bigl(C_{\mathcal{M}}\kappa^2(k\mu\delta+p\sigma^2)
+(\mu\delta+\sigma^2)\bigr)\log T$ in the unwhitened basis, where
$\xxT{\vect{x}}{t}-\Sigma_t$ has unbounded operator norm and the $\psi_1$ matrix
Bernstein passes the spike scales through $\mathcal{V}$
(Appendix~\ref{appendix:power-error-bound}).

\emph{Drift part.} Since $\lambda_{\min}(\Sigma_{\ell B})=\sigma^2$,
\begin{equation*}
\bigl\|\Sigma_{\ell B}^{-1/2}\mathcal{E}_2(\ell)\Sigma_{\ell B}^{-1/2}\bigr\|
\;\le\; \frac{1}{B\sigma^2}\sum_{t=(\ell-1)B+1}^{\ell B}\|\Sigma_t-\Sigma_{\ell B}\|
\;\le\; \frac{B\Gamma}{2\sigma^2}.
\end{equation*}

\emph{Combined.} For $B\gtrsim p\log(2pT^2)$ the $\mathcal{M}_y$ term is dominated by
the variance term, so
\begin{equation*}
\bigl\|\Sigma_{\ell B}^{-1/2}\mathcal{E}(\ell)\Sigma_{\ell B}^{-1/2}\bigr\|
\;\lesssim\; \sqrt{p\log(2pT^2)/B}+B\Gamma/(2\sigma^2),
\end{equation*}
both terms $\mu$-free.

\subsection{On the cumulative metric}
\label{appendix:cumulative-metric}
As discussed in Section~\ref{framework}, we report the terminal subspace error $d(\mathrm{ran}(\mat{A}_T),\phi_\mathcal{X})$, equivalently the per-step error along the trajectory.
When $\Gamma>0$ this does not vanish as $T\to\infty$: Theorem~\ref{thm:lower_bound} forces a positive floor, and Figure~\ref{fig:convergence-pm} shows both algorithms settling at it.
That non-vanishing floor is what makes the terminal error the informative quantity here, since it discriminates between algorithms and pins down the optimal block size $B$ and learning rate $\zeta$.
This reverses the classical picture.
In stationary streaming PCA, and in bandit or online settings, the instantaneous error (or per-round regret) decays to zero, so the terminal snapshot is uninformative and cumulative regret becomes the standard summary.

Cumulative versions of our metric follow from the same bounds and leave this picture unchanged.
Theorems~\ref{thm:robust-power-method} and~\ref{thm:ojas-algorithm} apply at every intermediate $t\in[T]$ by re-running Theorem~\ref{thm:drifting-power-method} with the block count set to the number of blocks elapsed by time $t$ ($t/B$ for NPM, $\zeta t$ for Oja) in place of the terminal block count.
The same-form bound holds at each $t$, and union-bounding over $t\in[T]$ pulls a $\log T$ factor into the rate (absorbed into $\tilde{O}$).
The time-average $\frac{1}{T}\sum_{t=1}^T\mathrm{metric}(\mat{A}_t,\phi_\mathcal{X})$ and the discounted sum $\sum_{t=1}^T\mathrm{coeff}^{T-t}\,\mathrm{metric}(\mat{A}_t,\phi_\mathcal{X})$ with $\mathrm{coeff}\in(0,1]$ therefore inherit the terminal error's rate up to that factor, the discounted form measuring recent-tracking accuracy, the natural target under drift.
The lower bound transfers by reconstructing the packing $\mathcal{A}_1,\dots,\mathcal{A}_{M'}$ against the cumulative objective.

\subsection{On the choice of subspace distance metric}
\label{appendix:distance-metric}

Our bounds are stated in the projection 2-distance $d(\mat M, \tilde{\mat M}) = \|\mat U_{1:k}\mat U_{1:k}^\top - \tilde{\mat U}_{1:k}\tilde{\mat U}_{1:k}^\top\| = \|\sin\Psi\|_\infty = \sin\psi_k$ (Definition~\ref{def:projection_dist}), the sine of the largest of the $k$ principal angles $\psi_1\le\dots\le\psi_k\in[0,\pi/2]$.
Operationally $d=0$ iff the subspaces coincide and $d=1$ iff some direction in one is orthogonal to the entire other; $d$ measures the worst-direction misalignment.
The Grassmannian-standard alternative is the chordal distance $d_F = \|\sin\Psi\|_F$, with squared form $d_F^2 = \sum_i\sin^2\psi_i$ summing per-direction misalignments.
The packing-volume bound of~\citet{dai2005} is stated for $d_F$, and Proposition~\ref{prop:grassvolume} adapts it to $d$.
The two metrics relate by $d \le d_F \le \sqrt k\cdot d$.

By norm equivalence our NPM and minimax bounds translate to $d_F$ statements,
\begin{align*}
\text{NPM upper bound}:&\quad d_F \;\lesssim\; \sqrt k\cdot\bigl(\Gamma(p\sigma^2+k\mu\delta)(\sigma^2+\mu\delta)/\delta^3\bigr)^{1/3},\\
\text{minimax lower bound}:&\quad d_F \;\gtrsim\; \bigl(\Gamma p\sigma^2(\sigma^2+\delta)/\delta^3\bigr)^{1/3},
\end{align*}
widening the matching tightness gap by at most $\sqrt k$.
Whether a direct $d_F$ analysis would close this $\sqrt k$ slack is open.
For stationary spike-covariance PCA the minimax rates in $d_F$ and $d$ differ by an order-$\sqrt k$ factor~\citep{Vu2013, Cai2015}, so direct $d_F$ analysis does not close the gap in the stationary case; the drifting case is structurally similar but we have not verified it.

Closing the slack would require re-deriving Appendix~\ref{appendix:amplification} in $d_F$ form.
The current signal-floor and complement-ceiling argument follows the worst principal angle through $s_k(\mat M\mat W)$ and $\|\mat M\mat N\|$, single spectral-norm quantities that produce $d$.
A $d_F$ version would have to follow $\sum_i\sin^2\psi_i$ across the iteration, with the directions coupled by the per-block multiplication, and chain the per-block bounds in this Frobenius form across $L$ blocks.
We do not pursue this; the norm-equivalence translation gives $\sqrt k\cdot d$ with no proof-structure changes.

\end{document}